%% file: main.tex
\documentclass[10pt,twocolumn,letterpaper]{article}

\usepackage{cvpr}              

\usepackage{graphicx}
\usepackage{amsmath}
\usepackage{amssymb}
\usepackage{booktabs}

\usepackage[utf8]{inputenc} 
\usepackage[T1]{fontenc}    
\usepackage[pagebackref,breaklinks,colorlinks]{hyperref}
\usepackage{url}            
\usepackage{booktabs}       
\usepackage{amsfonts}       
\usepackage{nicefrac}       
\usepackage{microtype}      
\usepackage{xcolor}         
\usepackage{colortbl}
\usepackage{graphicx}       
\usepackage{setspace}
\usepackage{enumitem}
\usepackage{multirow}
\usepackage{adjustbox}
\usepackage{caption}
\usepackage{dirtytalk}
\usepackage{mathtools}
\usepackage{xspace}
\usepackage{soul}
\usepackage{amssymb}
\usepackage{pifont}
\usepackage{algorithm}
\usepackage{algpseudocode}
\usepackage{amsmath}
\usepackage{pbox}
\DeclareMathOperator*{\argmax}{arg\,max}

\newcommand{\cmark}{\ding{51}}%
\newcommand{\xmark}{\ding{55}}%

\newif\ifwithappendix
\withappendixfalse

\usepackage{tikz}
\usetikzlibrary{calc}
\graphicspath{{fig/}}

\usepackage[disable]{todonotes}
\definecolor{darkgreen}{rgb}{0,0.45,0}
\definecolor{lightgray}{gray}{0.9}


\DeclarePairedDelimiterX{\infdivx}[2]{(}{)}{%
	#1\;\delimsize|\delimsize|\;#2%
}
\newcommand{\kld}[2]{\ensuremath{D_{\text{KL}}\infdivx{#1}{#2}}\xspace}

\usepackage[capitalize]{cleveref}
\crefname{section}{Sec.}{Secs.}
\Crefname{section}{Section}{Sections}
\Crefname{table}{Table}{Tables}
\crefname{table}{Tab.}{Tabs.}

\def\cvprPaperID{7867} 
\def\confName{CVPR}
\def\confYear{2023}

\begin{document}

\title{Back to the Source: \\ Diffusion-Driven Adaptation to Test-Time Corruption}



\author{
	Jin Gao\thanks{~indicates equal contribution, $^\dag$ indicates equal advising.} $^1$, Jialing Zhang$^{*1}$, Xihui Liu$^3$, Trevor Darrell$^4$, Evan Shelhamer$^{\dag 5}$, Dequan Wang$^{\dag 1,2}$
        \and
        $^1$Shanghai Jiao Tong University $^2$Shanghai Artificial Intelligence Laboratory \\
        $^3$The University of Hong Kong $^4$University of California, Berkeley $^5$DeepMind
}

\maketitle

\begin{abstract}
  Test-time adaptation harnesses test inputs to improve the accuracy of a model trained on source data when tested on shifted target data.
  Most methods update the source \emph{model} by (re-)training on each target domain.
  While re-training can help, it is sensitive to the amount and order of the data and the hyperparameters for optimization.
  We update the target \emph{data} instead, and project all test inputs toward the source domain with a generative diffusion model.
  Our diffusion-driven adaptation (DDA) method shares its models for classification and generation across all domains, training
  both on source then freezing them for all targets, to avoid expensive domain-wise re-training.
  We augment diffusion with image guidance and classifier self-ensembling to automatically decide how much to adapt.
  Input adaptation by DDA is more robust than model adaptation across a variety of corruptions, models, and data regimes on the ImageNet-C benchmark.
  With its input-wise updates, DDA succeeds where model adaptation degrades on too little data (small batches), on dependent data (correlated orders), or on mixed data (multiple corruptions).
\end{abstract}

\input{1-intro}

\input{2-related}

\input{3-method}

\input{4-experiments}

\input{5-discussion}

\clearpage
\newpage

{
	\bibliographystyle{ieee_fullname}
	\bibliography{ref}
}


\input{0-appendix}


\end{document}

%% file: 1-intro.tex
\section{Introduction}
\label{sec:intro}

Deep networks achieve state-of-the-art performance for visual recognition~\cite{he2016deep, dosovitskiy2021image, liu2021swin, liu2022convnet},
but can still falter when there is a \emph{shift} between the source data and the target data for testing~\cite{quionero2009dataset}.
Shift can result from corruption~\cite{hendrycks2019robustness,mintun2021interaction};
adversarial attack~\cite{goodfellow2015explaining};
or natural shifts between simulation and reality, different locations and times, and other such differences~\cite{peng2017visda,koh2021wilds}.
To cope with shift, adaptation and robustness techniques update predictions to improve accuracy on target data.
In this work, we consider two fundamental axes of adaptation: what to adapt---the model or the input---and how much to adapt---using the update or not.
We propose a test-time input adaptation method driven by a generative diffusion model to counter shifts due to image corruptions.

The dominant paradigm for adaptation is to train the model by joint optimization over the source and target~\cite{saenko2010adapting,torralba2011unbiased,ganin2016domain,tzeng2017adversarial,hoffman2018cycada}.
However, train-time adaptation faces a crucial issue: not knowing how the data may differ during testing.
While train-time updates can cope with known shifts, what if new and different shifts should arise during deployment?
In this case, test-time updates are needed to adapt the model (1) without the source data and (2) without halting inference.
Source-free adaptation~\cite{sun2020test,varsavsky2020test,iwasawa2021test,kundu2020universal,li2020model,liang2020we}
satisfies (1) by re-training the model on new targets without access to the source.
Test-time adaptation~\cite{schneider2020improving,sun2020test,wang2021tent,zhang2021memo} satisfies (1) and (2) by iteratively updating the model during inference.
Although updating the model can improve robustness, these updates have their own cost and risk.
Model updates may be too computationally costly, which prevents scaling to many targets (as each needs its own model), and they may be sensitive to different amounts or orders of target data, which may result in noisy updates that do not help or even hinder robustness.
In summary, most methods update the source \emph{model}, but this does not improve all deployments.

\input{fig/one4all}

We propose to update the target \emph{data} instead.
Our diffusion-driven adaptation method, DDA, learns a diffusion model on the source data during training, then projects inputs from all targets back to the source during testing.
Figure~\ref{fig:one4all} shows how just one source diffusion model enables adaptation on multiple targets. 
DDA trains a diffusion model to replace the source data, for source-free adaptation, and adapts target inputs while making predictions, for test-time adaptation.
Figure~\ref{fig:method} shows how DDA adapts the input then applies the source classifier without model updates.

Our experiments compare and contrast input and model updates on robustness to corruptions.
For input updates, we evaluate and ablate our DDA and compare it to DiffPure \cite{nie2022diffusion}, the state-of-the-art in diffusion for adversarial defense.
For model updates, we evaluate entropy minimization methods (Tent~\cite{wang2021tent} and MEMO~\cite{zhang2021memo}), the state-of-the-art for online and episodic test-time updates, and BUFR~\cite{eastwood2021source}, the state-of-the-art for source-free offline updates.
DDA achieves higher robustness than DiffPure and MEMO across ImageNet-C and helps where Tent degrades due to limited, ordered, or mixed data.
DDA is model-agnostic, by adapting the input, and improves across standard (ResNet-50) and state-of-the-art convolutional (ConvNeXt~\cite{liu2022convnet}) and attentional (Swin Transformer~\cite{liu2021swin}) architectures without re-tuning.

\vspace{2mm}
\noindent\textbf{Our contributions:}
\begin{itemize}[leftmargin=*,noitemsep]
\item We propose DDA as the first diffusion-based method for test-time adaptation to corruption and include a novel self-ensembling scheme to choose how much to adapt.
\item We identify and empirically confirm weak points for online model updates---small batches, ordered data, and mixed targets---and highlight how input updates address these natural but currently challenging regimes.
\item We experiment on the ImageNet-C benchmark to show that DDA improves over existing test-time adaptation methods across corruptions, models, and data regimes.
\end{itemize}

%% file: fig/one4all.tex
\begin{figure*}[t]
\begin{center}
\adjustbox{max width=0.9\linewidth}{
\begin{tabular}{c c c c c}
\includegraphics[width=\linewidth]{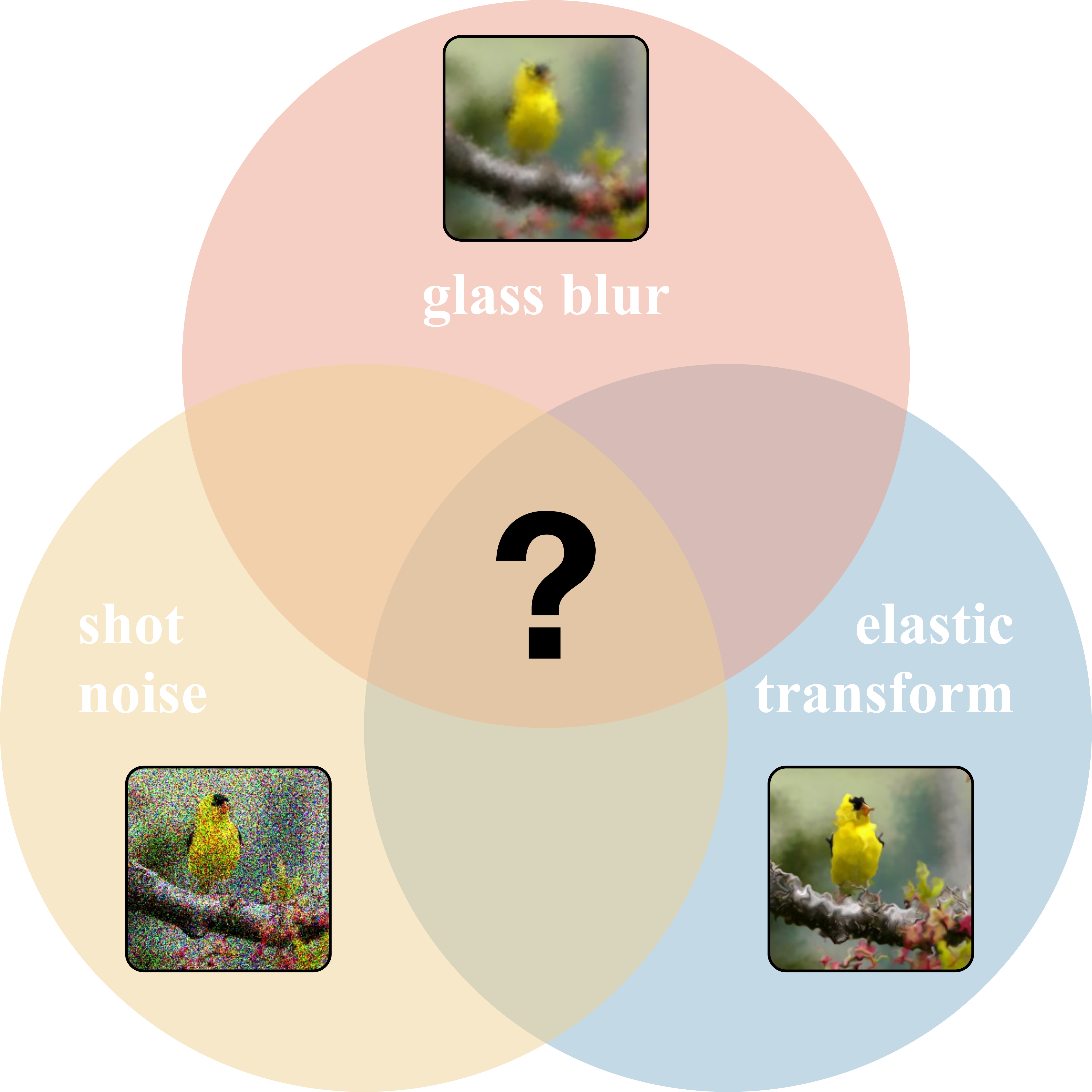} & &
\includegraphics[width=\linewidth]{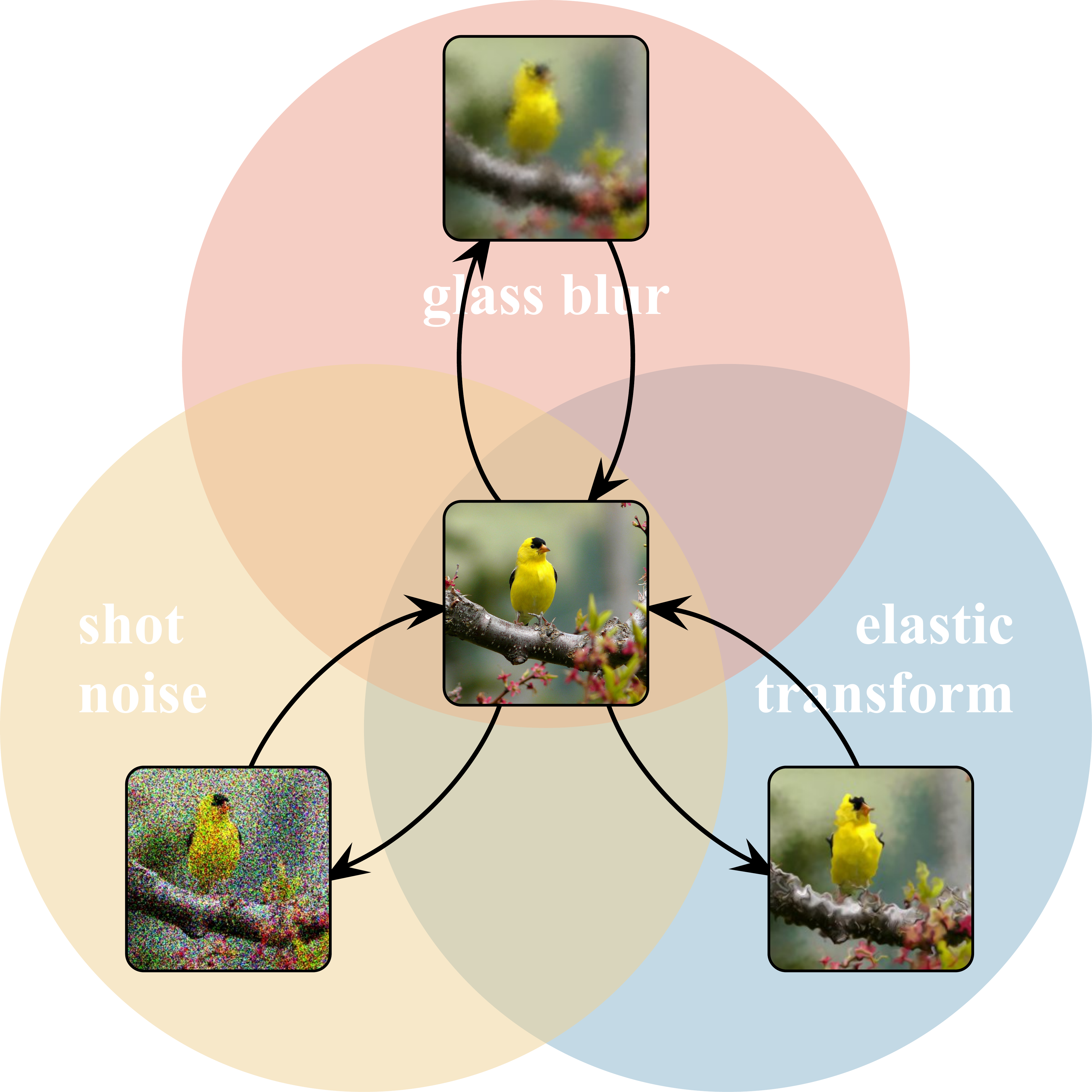} & &
\includegraphics[width=\linewidth]{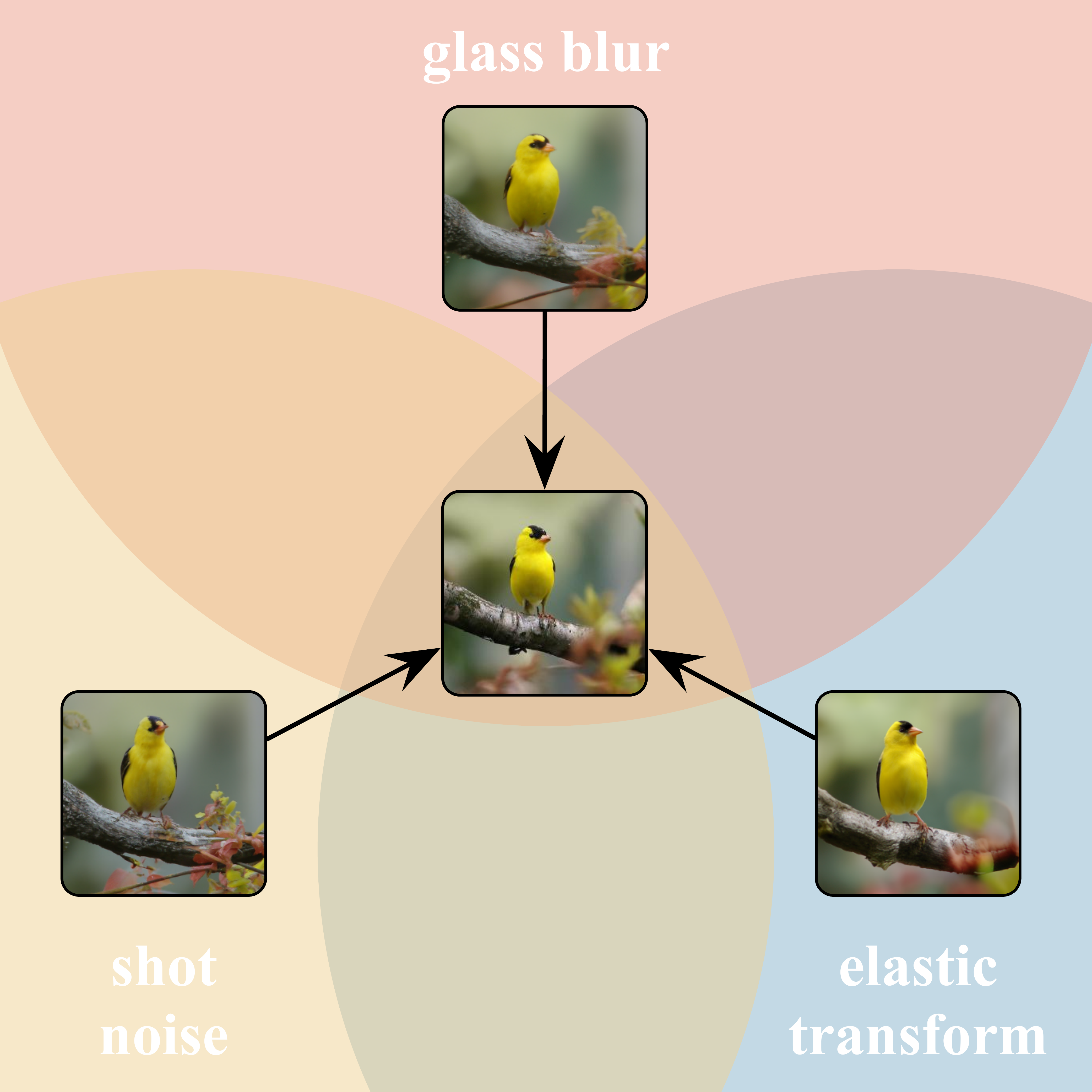} \\
& & & & \\
& & & & \\
\resizebox{!}{0.25in}{(a) Setting: Multi-Target Adaptation} & & 
\resizebox{!}{0.25in}{(b) Cycle-Consistent Paired Translation} & & 
\resizebox{!}{0.25in}{(c) DDA (ours): Many-to-One Diffusion} \\
\end{tabular}
}
\end{center}
\caption{
  \textbf{One diffusion model can adapt inputs from new and multiple targets during testing}.
  Our adaptation method, DDA, projects inputs from all target domains to the source domain by a generative diffusion model.
  Having trained on the source data alone, our source diffusion model for generation and source classification model for recognition do not need any updating, and therefore scale to multiple target domains without potentially expensive and sensitive re-training optimization.
}
\label{fig:one4all}
\vspace{-3mm}
\end{figure*}

%% file: 2-related.tex
\section{Related Work}
\label{sec:related}

\textbf{Model Adaptation} updates the source model on target data to improve accuracy.
We focus on source-free adaptation---not needing the source while adapting---and on test-time adaptation---making predictions while adapting---because DDA is a source-free and test-time method.

\emph{Source-free adaptation}~\cite{li2020model,kundu2020universal,liang2020we,sun2020test}
makes it possible to respect practical deployment constraints on computation, bandwidth, and privacy.
Nevertheless, most methods involve a certain amount of complexity and computation by altering training~\cite{li2020model,kundu2020universal,liang2020we,sun2020test,dubey2021adaptive} and interrupt testing by re-training their model(s) offline on each target~\cite{li2020model,kundu2020universal,liang2020we,eastwood2021source,Pandey_2021_CVPR}.
DDA is source-free, as it replaces the source data with source diffusion modeling.
However, it differs by updating the data rather than the model(s).
Furthermore, it does not alter the training of the classifier, as the diffusion model is trained on its own.
By keeping its models fixed, DDA handles multiple targets without halting testing for model re-training, as source-free model adaptation does.

\emph{Test-time adaptation}~\cite{sun2020test,schneider2020improving,wang2021tent,zhang2021adaptive,niu2022efficient,zhou2021bayesian,tang2023neuro,niu2023towards} simultaneously updates and predicts.

Such test-time model updates can be sensitive to their optimization hyperparameters along with the size, order, and diversity of the test data.
On the contrary, DDA updates the data, which makes it independent across inputs, and thereby invariant to batches, orders, or mixtures of the test data.
DDA can even adapt to a single test input, without augmentation, unlike test-time model adaptation.

\textbf{Input Adaptation} translates data between source and target.
DDA adapts the input from target to source by test-time diffusion.
Prior methods adapt during testing, but differ in their purpose and technique, or adapt during training, but cannot handle new target domains during testing.

During testing, translation from source to target enables the use of a source-only model.
DiffPure~\cite{nie2022diffusion} is the closest method to DDA because it applies diffusion to defense against the adversarial shift.
However, DiffPure and DDA differ in their settings of adversarial and natural shift respectively, and as a result differ in their techniques.
DDA differs in its conditioning of the diffusion updates and its self-ensembling of predictions before and after adaptation.

During training, translation from source to target
provides additional data or auxiliary losses.
Train-time translation includes style transfer~\cite{reinhard2001color, pitie2007automated, li2018closed, yoo2019photorealistic}, conditional image synthesis~\cite{zhu2017unpaired, hoffman2018cycada, huang2018multimodal, park2019semantic, park2020contrastive, jiang2020tsit, richter2022enhancing}, or adversarial generation~\cite{rusak2020simple} for robustness to shift.
CyCADA~\cite{hoffman2018cycada} adapts by translating between source and target via generation with CycleGAN~\cite{zhu2017unpaired}.
While CyCADA and DDA are generative, CyCADA needs paired source and target data for training, and cannot adapt to multiple targets during testing.
DDA only trains one model on source to adapt to multiple targets.

\input{fig/method}

\paragraph{Diffusion Modeling}
Diffusion~\cite{sohl2015deep,rombach2022high,song2019generative,song2020improved,nichol2021improved,song2021score,rombach2022high} is a strong, recent approach to generative modeling that samples by iteratively refining the input.
In essence, diffusion learns to ``reverse'' noise to generate an image by gradient updates w.r.t. the input.
The type of noise matters, and standard diffusion relies on Gaussian noise.
In this work, we investigate how a strong diffusion model can project corrupted target data toward the source data distribution, even on corruptions that are highly non-Gaussian.
We apply the denoising diffusion probabilistic model (DDPM)~\cite{ho2020denoising} in this new role of diffusion-driven adaptation.
Guided diffusion models improve generation by optimization conditioned on class labels~\cite{dhariwal2021diffusion,ho2021classifier}, text~\cite{nichol2022glide,liu2023more}, and images~\cite{choi2021ilvr}, but test-time adaptation denies the data needed for their use as-is.
DDA improves on the straightforward application of diffusion to achieve higher robustness to corruption during testing.

%% file: fig/method.tex
\begin{figure*}[t]
\centering
\includegraphics[width=0.9\linewidth]{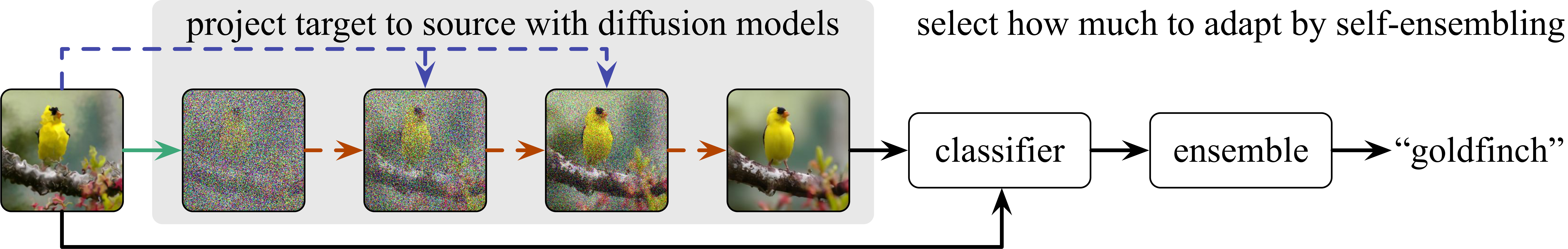}
\caption{%
  \textbf{DDA projects target inputs back to the source domain.}
  Adapting the input during testing enables direct use of the source classifier without model adaptation.
  The projection adds noise (forward diffusion, green arrow) then iteratively updates the input (reverse diffusion, red arrow)
  with conditioning on the original input (guidance, purple arrow).
  For reliability, we ensemble predictions with and without adaptation depending on their confidence.
}
\label{fig:method}
\end{figure*}

%% file: 3-method.tex
\section{Diffusion-Driven Adaptation to Corruption}
\label{sec:method}

We propose diffusion-driven adaptation (DDA) to adopt a diffusion model to counter shifts due to input corruption.
During training, we train a generation model (the diffusion model) with the source data,
and train a recognition model (the classifier) with the source data and its labels.
During inference, taking an example from the target domain as input,
the diffusion model projects it back to the source domain,
and then the classifier makes a prediction on the projected image.
Figure~\ref{fig:method} illustrates the projection and prediction steps of DDA inference.

Our DDA approach does not need any target data during training,
and is able to accept arbitrary unknown target inputs during testing.
Notably, this enables inference on a single image from the target domain.
In contrast, previous model adaptation approaches,
such as Tent~\cite{wang2021tent} and BUFR~\cite{eastwood2021source},
degrade on too little data (small batches), on dependent data (non-random order), or on mixed data (multiple corruptions).
See Sec.~\ref{sec:comparisonModelAdaptation} for our examination of these data regimes.
In this way, DDA addresses practical deployments that are not already handled by model adaptation.

\subsection{Background: Diffusion for Image Generation}
\label{subsec:diffusion}

Diffusion models
have recently achieved state-of-the-art image generation by iteratively refining noise into samples from the data distribution.
Given an image sampled from the real data distribution $x_0 \sim q(x_0)$,
the forward diffusion process defines a fixed Markov chain,
to gradually add Gaussian noise to the image $x_0$ over $T$ timesteps,
producing a sequence of noised images $x_1, x_2, \cdots, x_T$.
Mathematically, the forward process is defined as
\begin{equation}\label{eq:forward_x0}
	\begin{split}
		&q(x_{1:T}|x_{0}) := \prod_{t=1}^{T}q(x_{t}|x_{t-1}), \\
		& q\left(x_{t} \mid x_{t-1}\right):=N\left(x_{t} ; \sqrt{1-\beta_{t}} x_{t-1}, \beta_{t} \mathbf{I}\right),
	\end{split}
\end{equation}
where the sequence, $\beta_{1}, ..., \beta_{T}$,
is a fixed variance schedule to control the step sizes of the noise.

We can further sample $x_t$ from $x_0$ in a closed form,
\begin{equation}\label{eq:forward}
    q(x_t|x_0) := \sqrt{\bar{\alpha}_t}x_0+\epsilon\sqrt{1-\bar{\alpha}_t}, \epsilon \sim \mathcal{N}(0, 1),
\end{equation}
where $\alpha_{t} := 1 - \beta_{t}$ and $\overline{\alpha_{t}} := \prod_{s=1}^{t}\alpha_{s}$.

On the other hand, given the Gaussian noise sampled from the distribution $X_T \sim \mathcal{N}(0, \mathbf{I})$,
the reverse diffusion process iteratively removes the noise to generate an image in $T$ timesteps.
The reverse process is formulated as a Markov chain with Gaussian transitions:
\begin{equation}\label{eq:reverse}
	\begin{split}
		&p(x_{0:T}) := p(x_T)\prod_{t=1}^{T}p(x_{t-1}|x_{t}), \\
		&p_{\theta}\left(x_{t-1} \mid x_{t}\right):=N\left(x_{t-1} ; \mu_{\theta}\left(x_{t}, t\right), \sigma_{t}^{2}\left(x_{t}, t\right) \mathbf{I}\right).
	\end{split}
\end{equation}

Denoising diffusion probabilistic models (DDPM)~\cite{ho2020denoising}
set $\sigma_{t}\left(x_{t}, t\right)=\sigma_t\mathbf{I}$ to time-dependent constants.
$\mu_{\theta}$ is parameterized by a linear combination of $x_{t}$ and $\epsilon_{\theta}(x_{t}, t)$,
where $\epsilon_{\theta}(x_{t}, t)$ is a function that predicts the noise.
The parameters of $\mu_{\theta}\left(x_{t}, t\right)$ are optimized by the variational bound on the negative log-likelihood $\mathbb{E}[-\log p_\theta(x_0)]$.
With this parameterization and following DDPM~\cite{ho2020denoising}, the training loss $\mathcal{L}_{\text{simple}}$ simplifies to the mean-squared error between the actual noise $\epsilon \sim \mathcal{N}(0, \mathbf{I})$ in $x_t$ and the predicted noise
\begin{equation}\label{eq:loss}
\mathcal{L}_{\text{simple}} := ||\epsilon_{\theta}(x_{t}, t) - \epsilon||^2.
\end{equation}

Since their loss derives from a bound on the negative log-likelihood $\mathbb{E}[-\log p_\theta(x_0)]$,
diffusion models are optimized to learn a generative prior of the training data.

\begin{algorithm}[t]
	\caption{Diffusion-Driven Adaptation}
	\label{algo:diffusion}
	\begin{algorithmic}[1]
		\State \textbf{Input}: Reference image $x_0$
		\State \textbf{Output}: Generated image $x^g_0$
		\State $N$:  diffusion range, $\phi_{D}(\cdot):$ low-pass filter of scale $\mathrm{D}$
		\State Sample $x_{N} \sim q\left(x_{N} \mid x_0\right)$ \Comment{perturb input}
		\State $x^g_N \gets x_N$
		\For{$t \gets N$ \ldots $1$}
		\State $\hat{x}_{t-1}^{g} \sim p_{\theta}\left(x_{t-1}^{g} \mid x^g_{t}\right)$ \Comment{unconditional proposal}
        \State $\hat{x}^g_0 \leftarrow \sqrt{\frac{1}{\bar{\alpha}_t}}x^g_t - \sqrt{\frac{1}{\bar{\alpha}_t} - 1} \boldsymbol{\epsilon}_\theta(x_t^g, t)$
        \State $x^g_{t-1} \gets \hat{x}_{t-1}^{g} - \boldsymbol{w} \nabla_{x_t}\left\|\phi_{D}\left(x_{0}\right) -\phi_{D}\left(\hat{x}^{g}_0\right)\right\|_2$
		\EndFor
		\State \Return $x^g_0$
	\end{algorithmic}
\end{algorithm}

\subsection{Diffusion for Input Adaptation}
\label{subsec:guidance}

We now detail our diffusion-driven adaptation method.
A diffusion model is trained on the source domain to learn a generative prior of the input distribution for a source classifier.
Once trained, it can be applied to project single/multi-target domain data to the source domain, by
running the forward process followed by the reverse process.

Given an input image $x_0$ from the target domain and an unconditional diffusion model trained on the source domain, 
we first run the forward process (Eqn.~\ref{eq:forward}, the green arrow in Fig.~\ref{fig:method}) of the diffusion model, \ie, perturb the image with Gaussian noise.
We denote the image sequence derived by $N$ forward steps as $x_0, x_1, \cdots, x_N$,
where $N$ is a hyper-parameter (``diffusion range'') controlling the amount of noise added to the input image.
Then the reverse process (Eqn.~\ref{eq:reverse}, the red dotted arrow in Fig.~\ref{fig:method}) starts with the noised image $x_N$,
then removes noise for $N$ steps to generate the denoised image sequence $x^g_{N-1}, x^g_{N-2}, \cdots, x^g_0$.
Since the diffusion model has learned a generative prior of the source domain,
the generated image $x^g_0$ should be more likely under the distribution of the source data.

However, we notice a trade-off between preserving classes while translating domains when choosing different diffusion ranges $N$.
If $N$ is too large and too much noise is added to the image,
the diffusion model will not be able to preserve the class information in the input image.
On the contrary, if $N$ is too small and too little noise is added,
there are not enough diffusion steps to project images from the target to the source.
Our goal is to translate the domain from target to source,
while preserving the class information as much as possible.
Unfortunately, class and domain information are commonly entangled with each other,
making it difficult to find a trade-off for sufficient domain translation and class preservation.

To address this trade-off, we provide structural guidance during the reverse process.
We design an iterative latent refinement step (denoted by the purple dotted arrow in Fig.~\ref{fig:method}) conditioned on the input image in the reverse process, so that the image structure and class information can be preserved when translating images across domains.

Inspired by ILVR~\cite{choi2021ilvr}, we add a linear low-pass filter implemented by $\phi_{D}(\cdot)$,
a sequence of downsampling and upsampling operations with a scale factor of $D$, to capture the image-level structure.
We iteratively update the diffusion sample $x^g_t$ to reduce the structural difference of generated sample as measured by $D$.

At each step of reverse process, we can obtain an estimate of $x_0$, $\hat{x}^g_0$, from the noisy image at the current step $x_t^g$.

\begin{equation}\label{eq:x0}
		\hat{x}^g_0 = \sqrt{\frac{1}{\bar{\alpha}_t}}x^g_t - \sqrt{\frac{1}{\bar{\alpha}_t} - 1} \boldsymbol{\epsilon}_\theta(x_t^g, t).
\end{equation}
Therefore, we can avoid conflicting with the diffusion update by using the direction of similarity between the reference image $x_0$ and $\hat{x}^g_0$, not the one between $x_t$ and $x^g_t$.
At each step $t$ in the reverse process, we force $x^g_t$ to move in the direction that decreases the distance between $\phi_D(x_0)$ and $\phi_D(\hat{x}^g_0)$:
\begin{equation}\label{eq:guidance}
    x^g_{t-1} = \hat{x}_{t-1}^{g} - \boldsymbol{w} \nabla_{x_t}\left\|\phi_{D}\left(x_{0}\right) -\phi_{D}\left(\hat{x}^g_0\right)\right\|_2,
\end{equation}
with a scaling hyperparameter $\boldsymbol{w}$ to control the step size of guidance.
For simplicity, we neglect the difference between $t$ and $t-1$ and update $\hat{x}^g_{t-1}$ based on $x^g_t$'s gradient, and spare an extra reverse step.

In summary, we first perturb the input image from the target domain with noise in the forward process of the diffusion model,
and then in the reverse process, we adapt the input with iterative guidance to generate an image that is more like source data without altering the class information too much.
Algorithm~\ref{algo:diffusion} outlines the projection of target data back to the source by diffusion.

\subsection{Self-Ensembling Before \& After Adaptation}
\label{subsec:ensemble}
Adapting target inputs back to the source by diffusion
helps our source-trained recognition model to make more reliable predictions.
In most cases, diffusion generates an image that improves accuracy, because it has preserved the class information while projecting out the target shift (at least partially).
However, the diffusion model is not perfect, and can sometimes generate an image that is less recognizable to the classifier than the original target input.

Motivated by this possibility, we propose a self-ensembling scheme to aggregate the prediction results from the original and adapted inputs.
Since we have the test input $x_0$ and adapted input $x^g_0$ from diffusion,
we can run the classification model on both images.
We make the final prediction based on the average confidence of both,
\ie, $\argmax_c\frac{1}{2}\left(p_c + p_c^g\right)$,
where $c \in \{1, \ldots, C\}$, and the confidence of the $C$ categories is $p \in \mathbb{R}^C$ and $p^g \in \mathbb{R}^C$.

This self-ensembling scheme enables the automatic selection of how much to weigh the original and adapted inputs to further increase robustness.

%% file: 4-experiments.tex
\section{Experiments}
\label{sec:experiments}

\subsection{Setup}
\label{subsec:setup}

We summarize the data, settings, adaptation methods, and classification models studied in our experiments.
Full implementation detail is provided by the code in the supplementary material and the documentation of hyperparameters in the Appendix.

\paragraph{Datasets}
ImageNet-C (IN-C)~\cite{hendrycks2019robustness} and ImageNet-$\overline{\text{C}}$ (IN-$\bar{\text{C}}$)~\cite{mintun2021interaction} are standard benchmarks for robust large-scale image classification.
They consist of synthetic but natural corruptions
(noise, blur, digital artifacts, and different weather conditions)
applied to the ImageNet~\cite{russakovsky2015imagenet} validation set of 50,000 images.
IN-C has 15 corruption types at 5 severity levels.
IN-$\overline{\text{C}}$ has 10 more corruption types, selected for their dissimilarity to IN-C, at 5 severity levels.
We measure robustness as the top-1 accuracy of predictions on the most severe corruptions (level 5) on IN-C and IN-$\overline{\text{C}}$.
We evaluate DDA with the same hyperparameters across each dataset except as noted for ablation and analysis.

\paragraph{Adaptation Settings}
We consider two settings with more and less knowledge of the target domains.
\emph{Independent} adaptation is the standard setting for robustness experiments on ImageNet-C, where adaptation and evaluation are done independently for each corruption type.
\emph{Joint} adaptation is a more realistic and difficult setting, where adaptation and evaluation are done jointly over all corruptions by combining their data.
Experimenting with both settings allows standardized comparison with existing work and exploration of adaptation without knowledge of target domain boundaries.

\paragraph{Methods}
We compare DDA to an ablation without self-ensembling, model adaptation by MEMO~\cite{zhang2021memo} and Tent~\cite{wang2021tent}, and input adaptation by the adversarial defense DiffPure~\cite{nie2022diffusion}.
MEMO adapts to each input by augmentation and entropy minimization: it minimizes the entropy of the predictions w.r.t. the model parameters over different augmentations of the input, then resets.
By relying on data augmentation, MEMO avoids trivial solutions to optimizing so many parameters on a single input.
Tent adapts on batches of inputs by updating a small number of statistics and parameters by entropy minimization, but unlike MEMO it does not reset, and its updates compound across batches.
DiffPure and DDA rely on the same unconditional diffusion model~\cite{dhariwal2021diffusion} but differ in their reverse steps and guidance.
DiffPure simply adds a given amount of noise ($t = 150$) and then reverses to $t = 0$.

\paragraph{Classifiers}
\label{sec:models}
We experiment with multiple classifiers to assess general improvement.
We select ResNet-50~\cite{he2016deep} as a standard architecture, plus
Swin~\cite{liu2021swin} and ConvNeXt~\cite{liu2022convnet} to evaluate the state-of-the-art in attentional and convolutional architectures.
Experimenting with Swin and ConvNeXt sharpens our evaluation of adaptation as these architectures already improve robustness.

\subsection{Benchmark Evaluation: Independent Targets}
\label{subsec:source_compare}

\input{tab/source_and_imagenetc}

\paragraph{Input updates are more robust than model updates with episodic adaptation.}
We begin by evaluating source-only inference (without adaptation), model adaptation with MEMO, and input adaptation with DiffPure or our DDA.
Each method is ``episodic'', in making independent predictions for each input, for a fair comparison.
Table~\ref{tab:mCEImageNetC} summarizes each source classifier and compares the robustness of each method.
DDA achieves consistently higher robustness than MEMO and DiffPure.
On the latest
Swin-T and ConvNeXt-T models DDA still delivers a ${\sim}5$ point boost.

\paragraph{DDA consistently improves on IN-C corruption without catastrophic failure.}
Figure~\ref{fig:one_out_of_two} analyzes robustness across each corruption type of IN-C.
DDA is the most robust overall, although DDA without self-ensembling
can improve over the source-only model on most high-frequency corruptions.
As for low-frequency corruptions,
our self-ensembling automatically selects how much to adapt, and compensates for the current failures of diffusion to avoid drops on more global corruptions like fog and contrast.

\input{fig/corruption_types_imagenetc}

\input{fig/single_domain_batchsize}

Although DiffPure likewise adapts the input by diffusion, its specialization to adversarial attacks makes it unsuitable for input corruptions.
Its average accuracy on IN-C is worse than the accuracy without adaptation.
This drop underlines the need for the particular design choices of DDA that specialize it to natural shifts like corruptions, which are unlike the norm-bounded attacks DiffPure is designed for.

\paragraph{DDA is not sensitive to small batches or ordered data.}
The amount and order of the data for each corruption may vary in practical settings.
For the amount, source-free methods use the entire test set at once, while test-time methods may choose different batch sizes.
For the order of the target data, it is commonly shuffled (as done by Tent and other test-time methods).
We evaluate at different batch sizes, with and without shuffling, to understand the effect of these data regimes.
Figure.~\ref{fig:shuffle_batchsize_single} plots sensitivity to these factors.
DDA, MEMO, and DiffPure are totally unaffected, being episodic, but Tent is extremely sensitive.
Controlling the amount and order of data during deployment may not always be possible, but Tent requires it to ensure improvement.

\paragraph{DDA maintains accuracy on the corruptions of IN-$\overline{\text{C}}$.}
Table \ref{tab:mCEImageNetCBARmulti} compares input adaptation by DiffPure and DDA on IN-$\overline{\text{C}}$.
These corruption types are more difficult, as they are designed and selected to differ from natural images and the corruptions of IN-C.
While DDA does not improve robustness in this case, it averts the large drops caused by DiffPure, which are even larger than its drops on IN-C.
\input{tab/mCEImageNetCBAR.tex}

\subsection{Challenge Evaluation: Joint Targets}
\label{sec:comparisonModelAdaptation}

\input{tab/mCEImageNetCmulti}

The joint adaptation setting combines the data for all corruption types to present a new challenge.
The amount, order, and mixture of the data can be varied to complicate adaptation for methods that depend on the batching or ordering of domains.
DDA and MEMO can both address small batches, ordered data, and mixed domains, because they are episodic methods, which adapt to each input independently.
However, non-episodic methods like Tent have no such guarantee, because of its cumulative updates across inputs.

\paragraph{DDA is more robust on joint targets where Tent and other cumulative updates degrade.}
Table~\ref{tab:mCEImageNetCmulti} compares episodic adaptation by DDA, MEMO, and DiffPure with cumulative adaptation by Tent in the joint setting.
The reported results are an average under multiple experiments to avoid randomness though we find that there is almost no difference among different seeds.

While the episodic methods are all invariant to the joint setting, this is not the case for Tent.
Tent can adapt the best when its assumptions of large enough batches and randomly ordered data are met, but it can otherwise harm robustness.
In contrast, the accuracy of DDA is independent of batch size and data order, and helps robustness in each setting.

For further comparison to model updates in the joint setting, we evaluate batch normalization (BN) on the target data~\cite{schneider2020improving} and source-free adaptation of feature histograms by BUFR~\cite{eastwood2021source}.
We evaluate with ResNet-50, as it is a standard architecture for IN-C, and the focus of~\cite{schneider2020improving}.
Although BN is competitive in the independent setting, sharing the mean and variance across all corruptions in the joint setting cannot adapt well: it achieves worse than source model performance at $10.3\%$ accuracy vs. the $29.7\%$ accuracy of DDA.
BUFR does not report results with ImageNet scale, nor with ResNet-50, and our tuning could not achieve better than source-only accuracy.

\subsection{Analysis \& Ablation of Diffusion Updates}
\label{subsec:ablation}

\paragraph{Timing}
As diffusion models are computationally intense, we compare the time for model adaptation by MEMO and input adaptation by DiffPure and DDA.
We measure the wall clock time for single input inference with ResNet-50 on the same hardware (GeForce RTX 2080 Ti) and average over the test set of $50,000$ inputs.
Table~\ref{tab:flops} reports our profiling.
While this experimentally verifies the current cost of diffusion modeling, it underlines the importance of design choices: DDA is more robust to corruption and faster than DiffPure.
Furthermore, we confirm the potential for speed-up by applying accelerated sampling with DEIS~\cite{zhang2022fast}.

\input{tab/time}

\paragraph{Ablation}
We ablate the different diffusion steps that update the input.
As described in Sec.~\ref{subsec:guidance}, our diffusion-driven adaptation method is composed of a forward process, reverse process, and guidance.
We experiment with three settings as follows:
(1) We first run the forward process (\ie, add Gaussian noise) on the input image and then run the reverse process of the diffusion model to denoise, without our iterative guidance.
This setting is denoted as ``\textit{forward+reverse}''.
(2) We start from a random noise and run the reverse process of the diffusion model with iterative guidance, which we denote as ``\textit{reverse+guidance}''.
(3) Our DDA model combines both, \ie, we run the forward process on the input image and then run the reverse process of the diffusion model with iterative guidance.
Figure~\ref{fig:diffusion_module_ablation} shows the performance of ``forward-reverse'', ``reverse-guidance'', and our DDA approach which includes the forward process, reverse process, and iterative guidance.
The results demonstrate that each step contributes to the robustness of adaptation.

\input{fig/diffusion_module_ablation}

%% file: tab/source_and_imagenetc.tex
\begin{table}
  \caption{
    \textbf{DDA is more robust in the episodic setting.}
    Episodic inference is independent across inputs, and includes the source-only model without adaptation, model updates by MEMO, and input updates by DiffPure and DDA (ours).
    We evaluate accuracy on standard ImageNet and the corruptions of ImageNet-C. 
  }
  \label{tab:mCEImageNetC}
  \vspace{-5mm}
  \begin{center}
  \adjustbox{max width=\linewidth}{
  \renewcommand\arraystretch{1.2}
  \begin{tabular}{l|c|cccc>{\columncolor{lightgray}}c}
    &
    IN &
    \multicolumn{4}{c}{ImageNet-C Accuracy}
     \\
    Model &
    Acc. & 
    Source-Only &
    MEMO &
    DiffPure &
    DDA
     \\
    \midrule
    ResNet-50 & 
    76.6 &18.7 & 24.7 & 16.8 & \bf 29.7 \\
    Swin-T & 
    81.2 & 33.1 & 29.5 & 24.8 & \bf 40.0  \\
    ConvNeXt-T & 
    82.1 & 39.3 & 37.8 & 28.8 & \bf 44.2  \\
    Swin-B & 
    83.4 & 40.5 & 37.0 & 28.9 &  \bf 44.5\\
    ConvNeXt-B & 
    83.9 & 45.6 & 45.8 & 32.7 & \bf 49.4 \\
    
    \bottomrule
  \end{tabular}
  }
  \end{center}
\end{table}

%% file: fig/corruption_types_imagenetc.tex
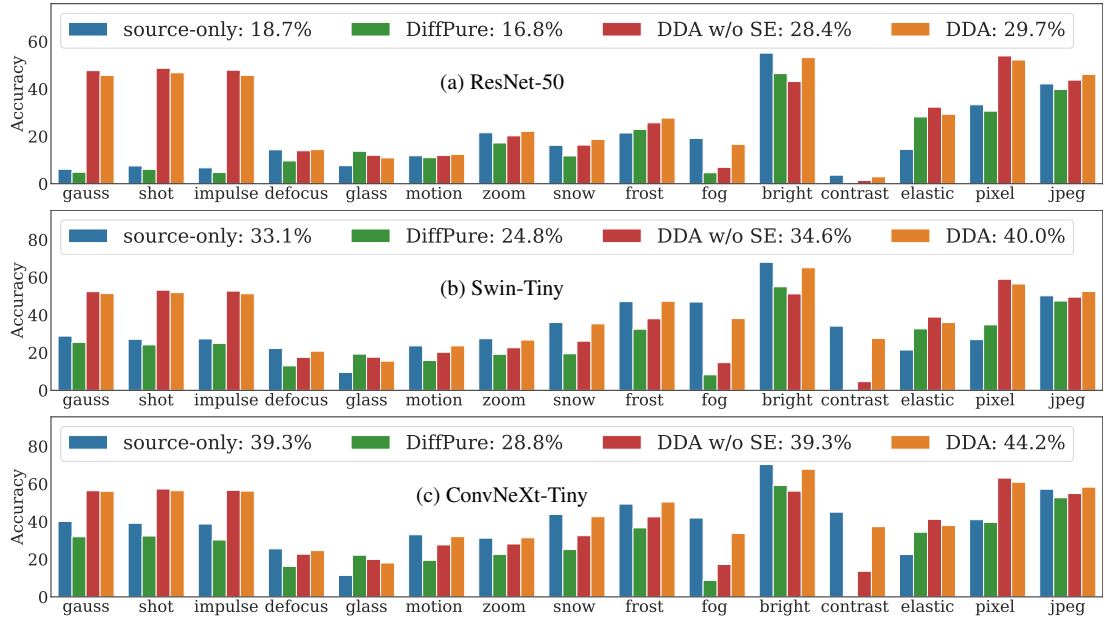
\begin{figure*}[t]
\vspace{-5pt}
\begin{center}
\adjustbox{max width=0.86\linewidth}{

    \begin{tabular}{c}
      \begin{tikzpicture}
        \node [
          above right,
          inner sep=0
        ] (image) at (0,0) {\includegraphics[width=\textwidth]{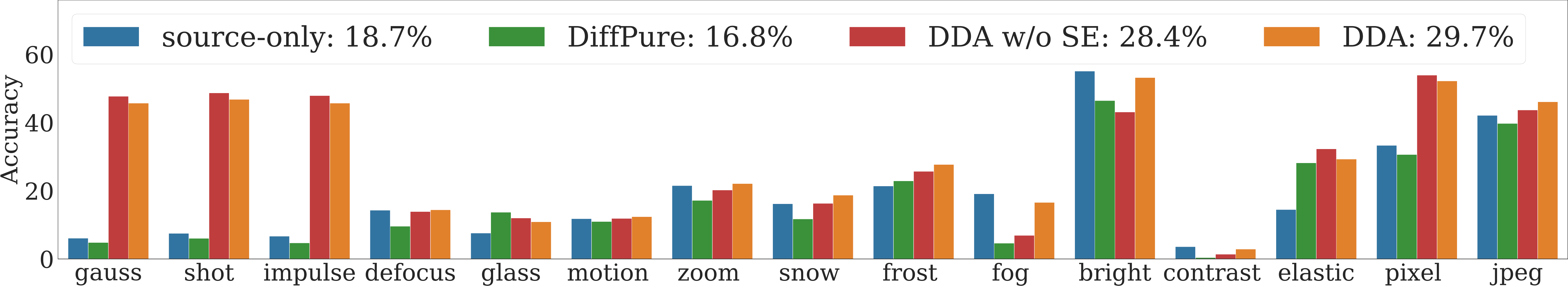}};

        \begin{scope}[
        x={($0.1*(image.south east)$)},
        y={($0.1*(image.north west)$)}]

        \node[above,black] at (4.5,5.1) {\normalsize (a) ResNet-50};

        \end{scope}
      \end{tikzpicture} \\
      \begin{tikzpicture}
        \node [
          above right,
          inner sep=0
        ] (image) at (0,0) {\includegraphics[width=\textwidth]{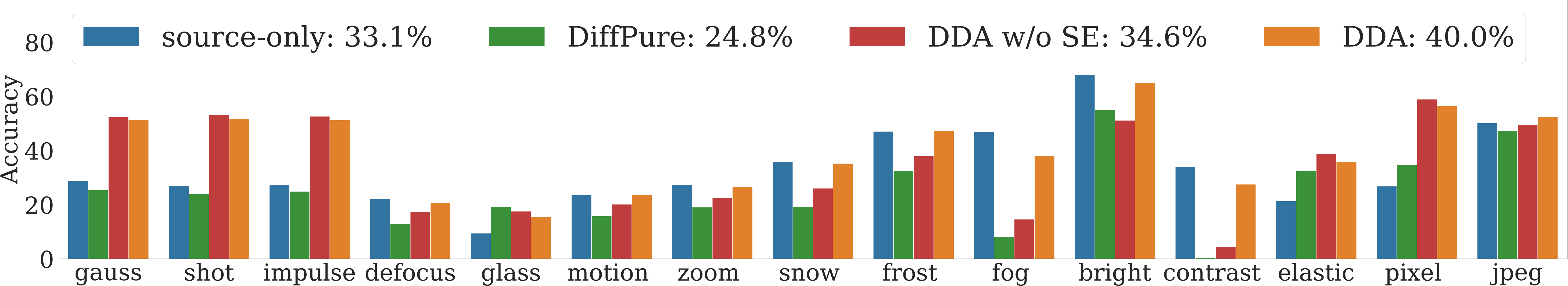}};

        \begin{scope}[
        x={($0.1*(image.south east)$)},
        y={($0.1*(image.north west)$)}]

        \node[above,black] at (4.5,5.1) {\normalsize (b) Swin-Tiny};

        \end{scope}
      \end{tikzpicture} \\
      \begin{tikzpicture}
        \node [
          above right,
          inner sep=0
        ] (image) at (0,0) {\includegraphics[width=\textwidth]{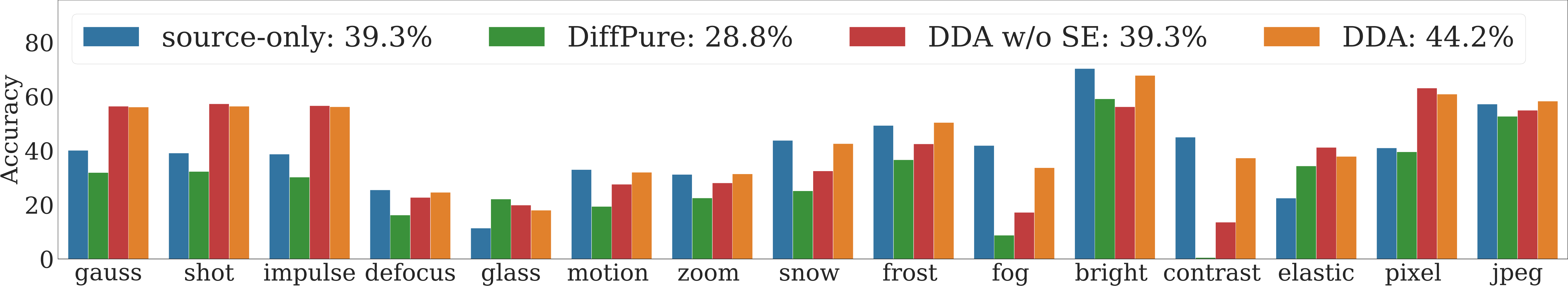}};

        \begin{scope}[
        x={($0.1*(image.south east)$)},
        y={($0.1*(image.north west)$)}]

        \node[above,black] at (4.5,5.1) {\normalsize (c) ConvNeXt-Tiny};

        \end{scope}
      \end{tikzpicture} \\
  \end{tabular}
}
\end{center}
\vspace{-4mm}
\caption{%
  \textbf{DDA reliably improves robustness across corruption types.}
  We compare DDA with the source-only model, state-of-the-art diffusion for adversarial defense (DiffPure), and a simple ablation of DDA (DDA w/o Self-Ensembling (SE)).
  DDA is the best on average, strictly improves on DiffPure, and improves on simple diffusion in most cases.
  Our self-ensembling prevents catastrophic drops (on fog or contrast, for example).
}
\label{fig:one_out_of_two}
\end{figure*}

%% file: fig/single_domain_batchsize.tex
\begin{figure*}[t]
  \begin{center}
    \adjustbox{max width=0.86\linewidth}{
      \begin{tabular}{c c c}
        \begin{tikzpicture}
          \node [
          above right,
          inner sep=0
          ] (image) at (0,0) {\includegraphics[width=\textwidth]{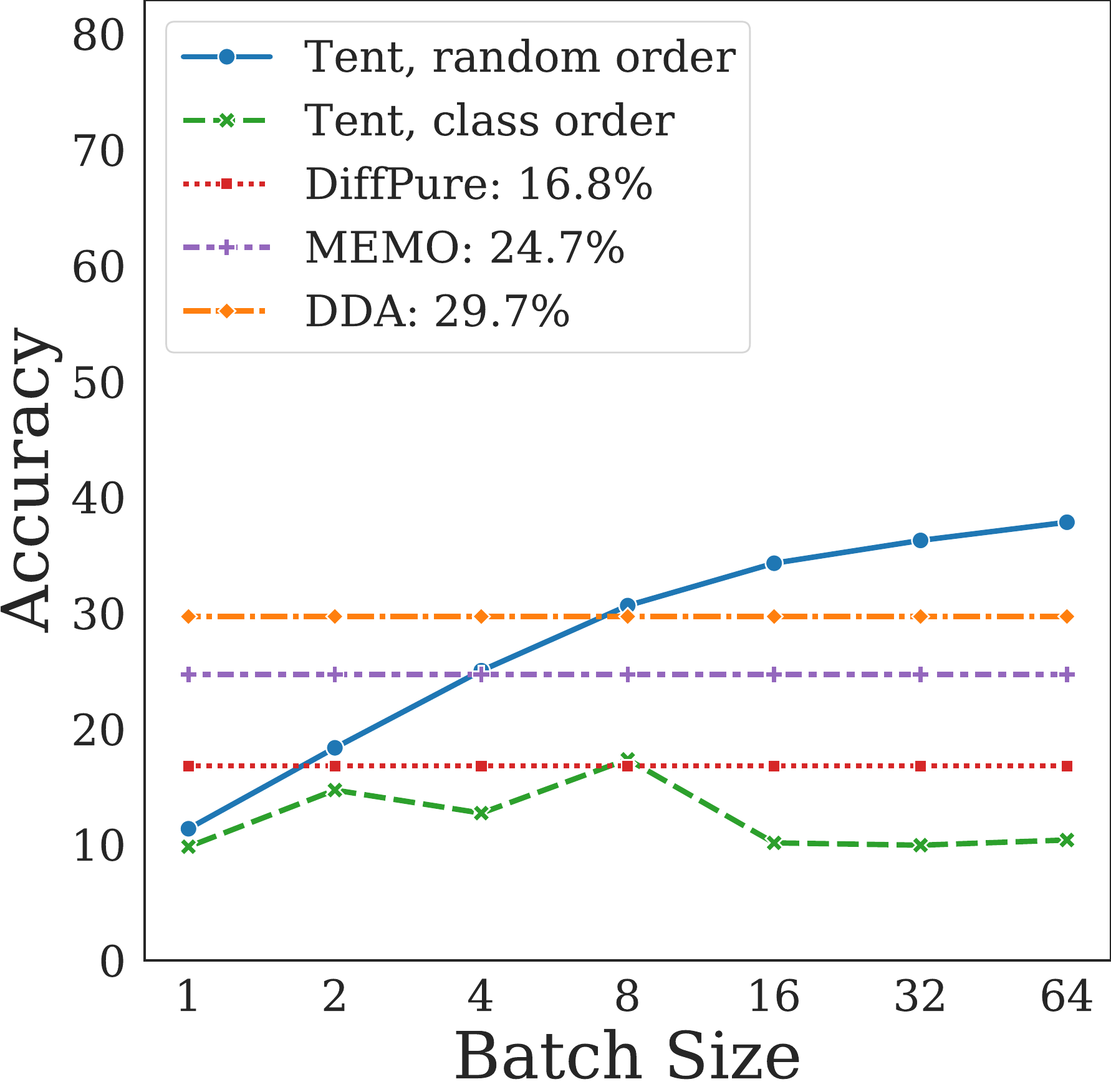}};

          \begin{scope}[
          x={($0.1*(image.south east)$)},
          y={($0.1*(image.north west)$)}]

          \node[above,black] at (3.5,1.5) {\Huge (a) ResNet-50};

          \end{scope}
        \end{tikzpicture} &
        \begin{tikzpicture}
          \node [
          above right,
          inner sep=0
          ] (image) at (0,0) {\includegraphics[width=\textwidth]{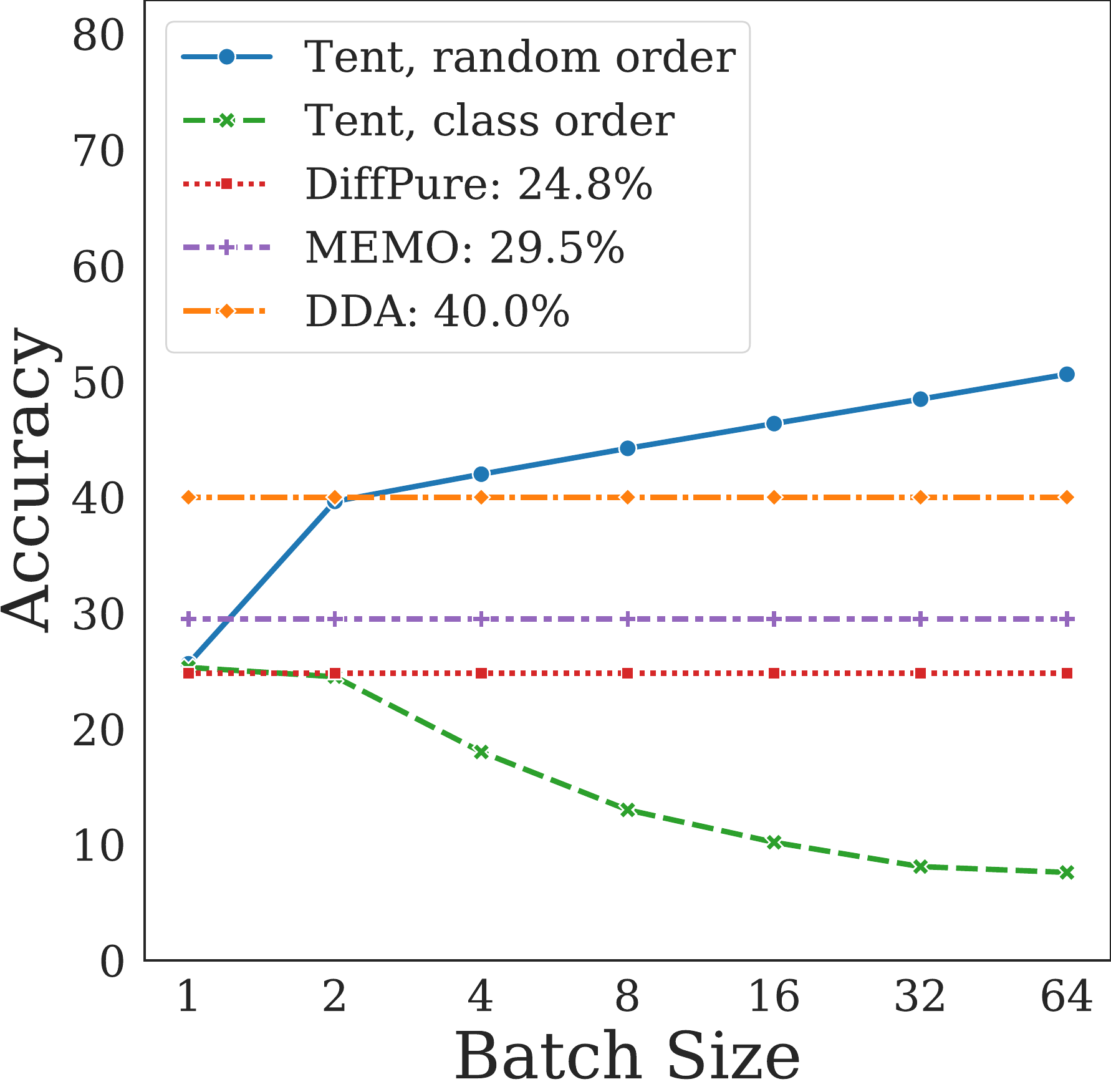}};

          \begin{scope}[
          x={($0.1*(image.south east)$)},
          y={($0.1*(image.north west)$)}]

          \node[above,black] at (3.5,1.5) {\Huge (b) Swin-Tiny};

          \end{scope}
        \end{tikzpicture} &
        \begin{tikzpicture}
          \node [
          above right,
          inner sep=0
          ] (image) at (0,0) {\includegraphics[width=\textwidth]{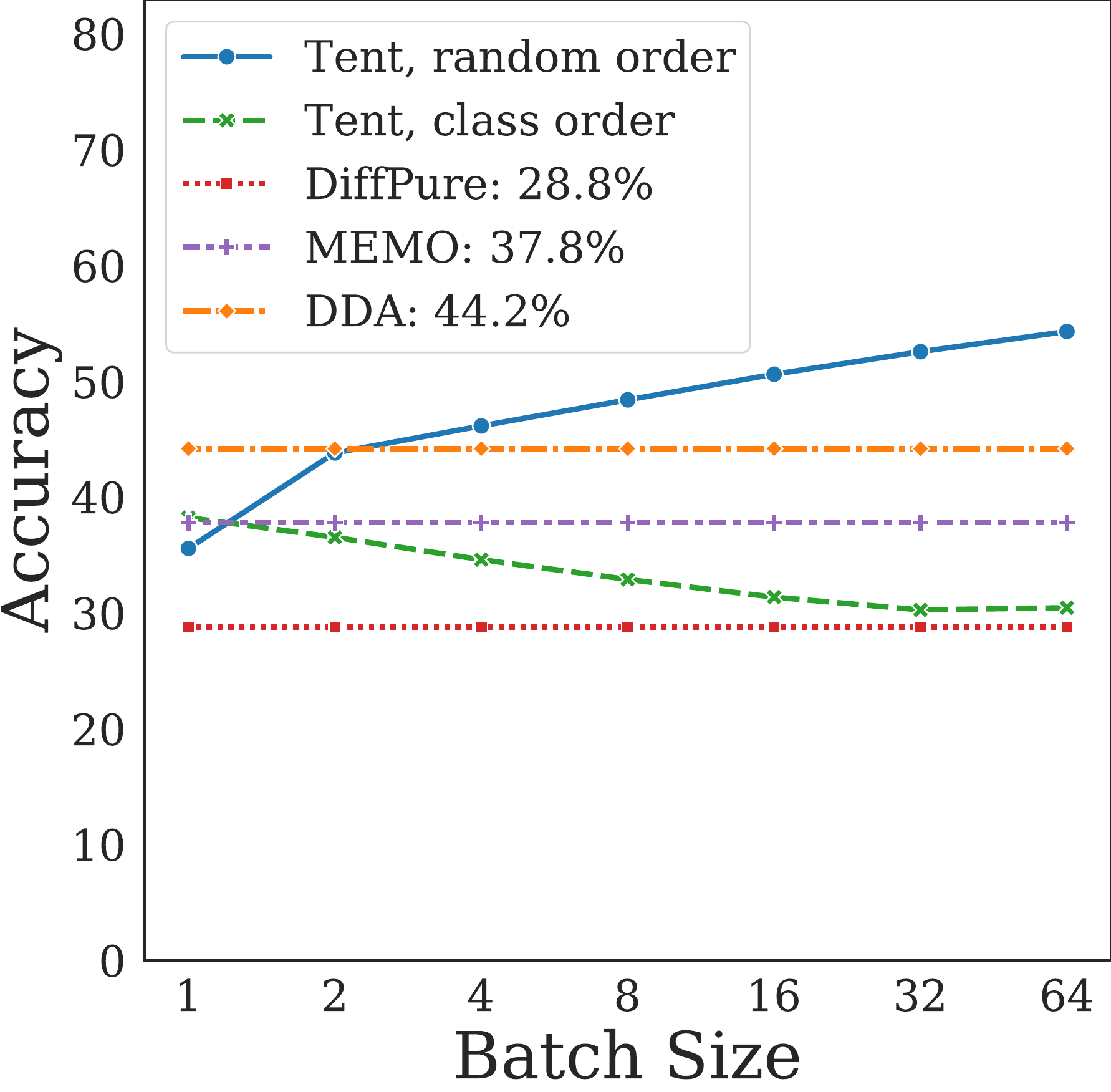}};

          \begin{scope}[
          x={($0.1*(image.south east)$)},
          y={($0.1*(image.north west)$)}]

          \node[above,black] at (3.5,1.5) {\Huge (c) ConvNeXt-Tiny};

          \end{scope}
        \end{tikzpicture} \\
      \end{tabular}
    }
  \end{center}
  \vspace{-4mm}
  \caption{%
    \textbf{DDA is invariant to batch size and data order while Tent is extremely sensitive.}
    To analyze sensivity to the amount and order of the data we measure the average robustness of independent adaptation across corruption types.
    DDA does not depend on these factors and consistently improves on MEMO.
    Tent fails on class-ordered data without shuffling and degrades at small batch sizes.
  }
  \vspace{-4mm}
  \label{fig:shuffle_batchsize_single}
\end{figure*}
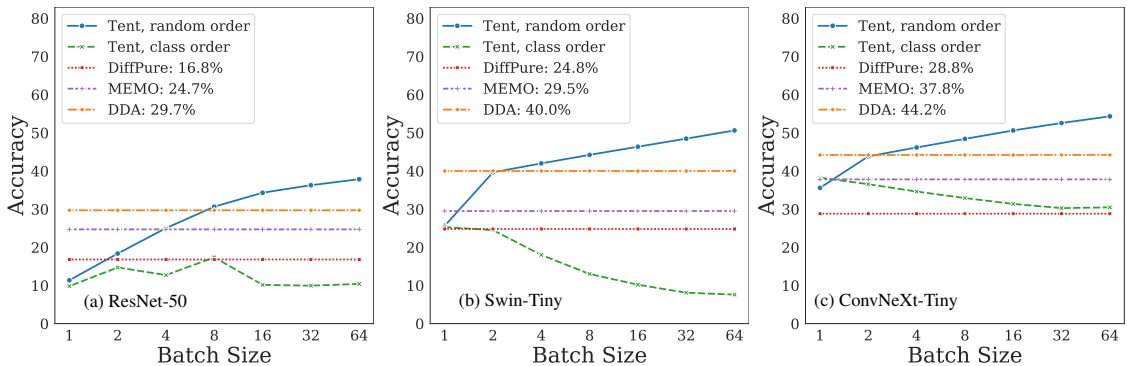

%% file: tab/mCEImageNetCBAR.tex
\begin{table}
  \caption{%
    \textbf{Diffusion vs. Other Corruptions.}
    We measure robustness to corruption on ImageNet-$\overline{\text{C}}$, which is designed to differ from ImageNet-C, by accuracy at maximum severity (level 5). 
  }
  \vspace{-3mm}
  \label{tab:mCEImageNetCBARmulti}
  \begin{center}
  \adjustbox{max width=0.8\linewidth}{
  \huge
  \renewcommand\arraystretch{1.2}
    \begin{tabular}{c|ccc}
      \toprule
      Method &
      ResNet-50 &
      Swin-T &
      ConvNeXt-T \\
      \midrule
      Source-Only & 25.8  &  44.2 & 47.2\\
      DiffPure~\cite{nie2022diffusion} & 19.8  & 28.5   & 32.1   \\
      \rowcolor{lightgray}
      DDA (ours) & 29.4  &  43.8  &  46.3 \\
      \bottomrule
  \end{tabular}}
  \end{center}
\end{table}

%% file: tab/mCEImageNetCmulti.tex
\begin{table}
  \caption{%
    \textbf{DDA is reliably more robust when the target data is limited, ordered, or mixed.}
    Deployment may supply target data in various ways. 
    To explore these regimes, we vary batch size and whether or not the data is ordered by class or mixed across corruption types.
    We compare episodic adaptation by input updates with DDA (ours) and by model updates with MEMO against cumulative adaptation with Tent.
    DDA and MEMO are invariant to these differences in the data.
    However, Tent is highly sensitive to batch size and order, and fails in the more natural data regimes.
  }
  \label{tab:mCEImageNetCmulti}
  \vspace{-3mm}
  \begin{center}
  \adjustbox{max width=\linewidth}{
  \huge
  \renewcommand\arraystretch{1.2}
    \begin{tabular}{c|cc|c|ccc}
      Method &
      \pbox{3em}{Mixed \\ Classes} &
      \pbox{3em}{Mixed \\ Types} &
      \pbox{3em}{Batch \\ Size} &
      ResNet-50 &
      Swin-T &
      ConvNeXt-T \\
      \midrule
      Source-Only & \multicolumn{2}{c|}{\multirow{4}{*}{N/A}} & \multirow{4}{*}{N/A} & 18.7 & 33.1 & 39.3 \\ 
      MEMO~\cite{zhang2021memo} & & & & 24.7 & 29.5 & 37.8 \\ 
      DiffPure~\cite{nie2022diffusion} & & &  & 16.8 & 24.8 & 28.8 \\ 
      \rowcolor{lightgray}
      DDA (ours) & & & &  \bf 29.7 &  \bf 40.0 & \bf 44.2 \\ 
      \midrule
      \multirow{4}{*}{Tent~\cite{wang2021tent}} & \xmark & \xmark & 1 / 64 & 2.2 / 0.4  & 0.2 / 0.2  & 0.1 / 1.4 \\
      & \xmark & \cmark & 1 / 64 & 1.6 / 0.5  & 0.2 / 0.5 & 0.3 / 1.6 \\
      & \cmark & \xmark & 1 / 64 & 3.0 / \bf{7.6} & 0.1 / 43.3  & 0.2 / 48.8 \\
      & \cmark & \cmark & 1 / 64 & 2.3 / 3.9  & 0.3 / \bf{44.1} & 0.3 / \bf{51.9}  \\

      \bottomrule
  \end{tabular}}
  \end{center}
\end{table}

%% file: tab/time.tex
\begin{table}
\caption{%
\textbf{DDA balances time and robustness.}
Diffusion by DDA or DiffPure is slower than entropy minimization by MEMO, 
but DDA is the most robust and faster than DiffPure.
Accelerating diffusion by DEIS can trade time and robustness for DDA.
}
\label{tab:flops}
\vspace{-5mm}
\begin{center}
\adjustbox{max width=\linewidth}{
    \begin{tabular}{lcccc}
      
      & MEMO~\cite{zhang2021memo} &
      DiffPure~\cite{nie2022diffusion} &
      DDA (ours) & DDA (+DEIS) \\
      \midrule
      Runtime (s) & 
      0.7 & 
      31.7 & 
      13.5 & 
       2.4
      \\ 
      IN-C Acc. (\%) &
      24.7 &
      16.8 &
      29.7 & 
     27.0
      \\
      \bottomrule
    \end{tabular}
}
\end{center}
\end{table}

%% file: fig/diffusion_module_ablation.tex
\begin{table}
\caption{%
    \textbf{Ablation of diffusion updates justifies each step.}
    We ablate the forward, reverse, and refinement updates of DDA.
    We omit self-ensembling to focus on the input updates.
    Forward adds noise, reverse denoises by diffusion, and refinement guides the reverse updates.
    DDA is best with all steps, but forward and reverse or reverse and refinement help on their own.
    }
    \label{tab:diffusion_module_ablation}
    \adjustbox{max width=\linewidth}{
      \renewcommand\arraystretch{1.15}
      \begin{tabular}{ll|ccc}
      &
      & ResNet-50
      & Swin-T
      & ConvNeXt-T
      \\
      \midrule
      \multirow{3}{*}{\rotatebox{90}{(a) elastic}}
      & forward+reverse
      & 24.5  & 24.9  & 25.8
      \\
      & reverse+guidance
      & 17.7  & 23.0 &  24.8
      \\
      & DDA (ours)
      & \bf 32.3  &  \bf 38.9  & \bf 41.2 
      \\
      \midrule
      \multirow{3}{*}{\rotatebox{90}{(b) blur}}
      & forward+reverse
      & 13.9  & 14.4  & 15.0
      \\
      & reverse+guidance
      &  7.6 &  11.5  &  13.4 
      \\
      & DDA (ours)
      &  \bf 12.0 & \bf 17.6  &  \bf 19.9 
      \\
      \midrule
      \multirow{3}{*}{\rotatebox{90}{(c) noise}}
      & forward+reverse
      & 19.5  & 20.2  & 21.0
      \\
      & reverse+guidance
      &   20.7  &  24.0   &  26.9 
      \\
      & DDA (ours)
      &  \bf 48.7  &\bf 53.2  &  \bf 57.3
      \\
      \bottomrule
      \end{tabular}
    }  
\end{table}

\begin{figure}[t]
    \centering
    \adjustbox{max width=\linewidth}{
      \begin{tabular}{c c c c c c}
        & \resizebox{!}{0.35in}{corrupted} & \resizebox{!}{0.35in}{forward} & \resizebox{!}{0.3in}{reverse+} & \resizebox{!}{0.33in}{DDA} & \resizebox{!}{0.35in}{original} \\
        & \resizebox{!}{0.35in}{image} & \resizebox{!}{0.3in}{+reverse} & \resizebox{!}{0.35in}{refinement} & \resizebox{!}{0.35in}{(both)} & \resizebox{!}{0.35in}{image} \\
        \resizebox{0.5in}{!}{\rotatebox{90}{(a) elastic}} &
        \includegraphics[width=\linewidth]{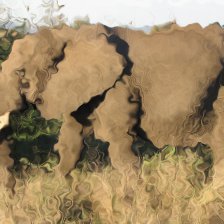} &
        \includegraphics[width=\linewidth]{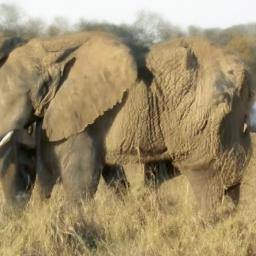} &
        \includegraphics[width=\linewidth]{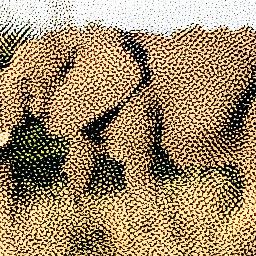} &
        \includegraphics[width=\linewidth]{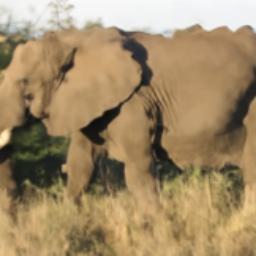} &
        \includegraphics[width=\linewidth]{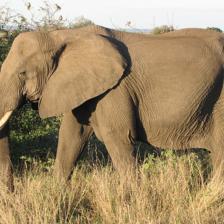} \\
        & & & \\
        \resizebox{0.5in}{!}{\rotatebox{90}{(b) glass blur}} &
        \includegraphics[width=\linewidth]{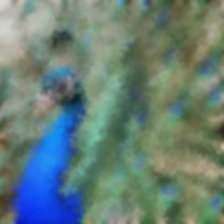} &
        \includegraphics[width=\linewidth]{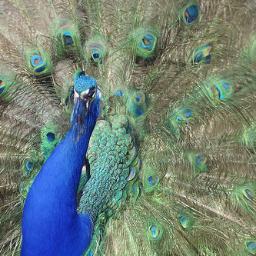} &
        \includegraphics[width=\linewidth]{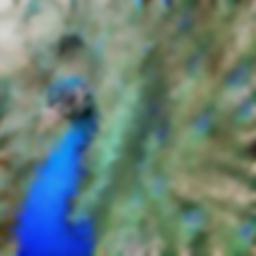} &
        \includegraphics[width=\linewidth]{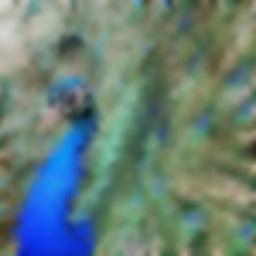} &
        \includegraphics[width=\linewidth]{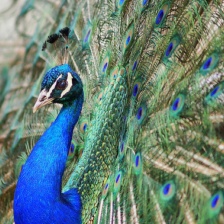} \\
        & & & \\
        \resizebox{0.5in}{!}{\rotatebox{90}{(c) shot noise}} &
        \includegraphics[width=\linewidth]{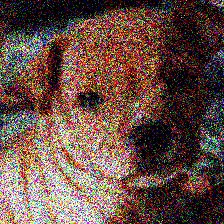} &
        \includegraphics[width=\linewidth]{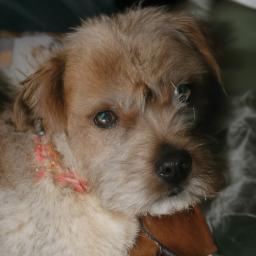} &
        \includegraphics[width=\linewidth]{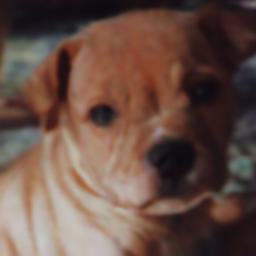} &
        \includegraphics[width=\linewidth]{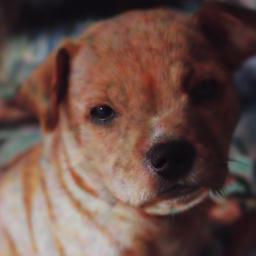} &
        \includegraphics[width=\linewidth]{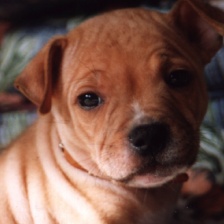} \\
      \end{tabular}
    }
  \caption{
    \textbf{Visualization of updates for ablations of diffusion.}
    }
  \label{fig:diffusion_module_ablation}
  \vspace{-3mm}
\end{figure}

%% file: 5-discussion.tex
\section{Discussion}
\label{sec:discussion}

DDA mitigates shift by test-time input adaptation with diffusion modeling.
Our experiments on ImageNet-C confirm that diffusing target data back to the source domain improves robustness.
In contrast to test-time model adaptation, which can struggle with scarce, ordered, and mixed data, our method is able to reliably boost accuracy in these regimes.
In contrast to source-free model adaptation, which can require re-training to each target, we are able to freely scale adaption to multiple targets by keeping our source models fixed.
These practical differences are due to our conceptual shift from model adaptation to input adaptation and our adoption of diffusion modeling.

Having examined whether to adapt by input updates or model updates, we expect that reconciling the two will deliver more robust generalization than either alone.

\paragraph{Limitations}
The strengths and weaknesses of input adaptation complement those of modal adaptation.
Although our method can adapt to a single target input, it must adapt from scratch on each input, and so its computation cannot be amortized across the deployment.
In contrast, model adaptation by TTT~\cite{sun2020test} or Tent~\cite{wang2021tent} can update on each batch while cumulatively adapting the model more and more.
Although diffusion can project many targets to the source data, and does so without expensive model re-training, it can fail on certain shifts.
If these shifts arise gradually, then model adaptation could gradually update too~\cite{kumar2020understanding}, but our fixed diffusion model cannot.

We rely on diffusion, and so we are bound to the quality of generation by diffusion.
Diffusion does have its failure modes, even though our positive results demonstrate its present use and future potential.
In particular, diffusion models may not only translate domain attributes but other image content, given their large model capacity.
Our use of image guidance helps avoid this, but at the cost of restraining adaptation on certain corruptions.
New diffusion architectures or new guidance techniques specific to adaption could address these shortcomings.

At present, diffusion takes more computation time than classification, so ongoing work to accelerate diffusion is needed to reduce inference time~\cite{salimans2021progressive}.
Our design choices bring DDA to $19\times$ the time as MEMO, while DiffPure takes $\sim45\times$ the time, but both diffusion methods are still slower than model updates.
Accelerating diffusion sampling by DEIS~\cite{zhang2022fast} reduces the time to ${<}4\times$ but sacrifices ${\sim}3$ points of robustness.
Further speed-up may require more fundamental changes to diffusion sampling and training.

\paragraph{Societal Impact}
While our work seeks to mitigate dataset shift, we must nevertheless remain aware of dataset bias.
Because our diffusion model is trained entirely on the source data, biases in the data may be reflected or amplified by the learned model.
Having learned from biased data, the diffusion model is then liable to project target data to whatever biases are present, and may in the process lose important or sensitive attributes of the target data.
While this is a serious concern, diffusion-driven adaptation at least allows for interpretation and monitoring of the translated images, since it adapts the input rather than the model.
Even so, making good use of this capacity requires diligence and more research into automated analyses of generated images.

%% file: 0-appendix.tex
\clearpage
\newpage

\appendix

\section{Ablation}

\subsection{Ablation on Self-Ensembling}

Since diffusion models themselves are not omnipotent, we propose a prediction fusion mechanism (Section \ref{subsec:ensemble}) to automatically choose the better candidate between an image pair, the vanilla test image, and the diffusion model's generation.
Entropy and confidence seem to be proper signals used for prediction fusion.
According to Tent~\cite{wang2021tent}, they are amazing signals indicating the potential of a prediction.
To some extent, the lower (higher) a prediction's entropy (confidence) is, the higher accuracy it would obtain.
Based on the entropy (confidence) of a prediction, we study some possibilities to utilize the image pair better.
Our exploration includes two parts, hard selection, and soft fusion,
which simply selects an image from the image pair and fuses the image pair into one new image, respectively.
The soft fusion can be operated on both pixel and logit levels.

\paragraph{Hard Selection}
Since entropy (confidence) can reflect the real accuracy of a prediction, a simple idea is to pick the image, logits of which have lower entropy (higher confidence) to make the final prediction.

\paragraph{Early Fusion}
Apart from the hard selection which selects an image from two,
we can fuse two images on the pixel level into a new one according to entropy (confidence).
We simply fuse two images, $X_1$ and $X_2$, using the weighted sum $F(a, b; f)$, where weights are from the entropy (confidence) of two images' logits, $y_1$ and $y_2$. We use a softmax operation to ensure the sum of two weights is one.

\begin{equation}
	F(a, b; f) = [a * f(X_1) + b * f(X_2)] / [f(X_1) + f(X_2)]
\end{equation}
It is worth noting that when using confidence, an image's weight is from its confidence, $conf(\cdot)$. The new image is $X_{new} = F(X_1, X_2; confidence)$. As for the entropy weight, it is from the other image's entropy, $ent(\cdot)$. The corresponding new image is $X_{new} = F(X_2, X_1; entropy)$.

\paragraph{Late Fusion}
Logits are also an excellent perspective for prediction fusion. We can take a similar strategy as the early fusion to fuse the logits, rather than image pixels.
As for confidence fusion, the new image is $y_{new} = F(y_1, y_2; confidence)$. As for the entropy fusion, the new image is $y_{new} = F(y_2, y_1; entropy)$.

As can be seen in Table~\ref{tab:ablationImageNetC2in1}, late fusion shows a better performance in all six models. Experiments have shown that the prediction fusion can effectively combine information from both the test image and the diffusion model's generation, and make a more accurate precision.

\input{tab/ablationImageNetC2in1}

\input{fig/one_out_of_two_appendix}

\paragraph{The additional evaluation of Self-Ensembling}
Figure~\ref{fig:one_out_of_two_appendix} evaluates the self-ensembling among DDA and state-of-the-art diffusion for adversarial defense (DiffPure).
It is shown that the ensembling leads to better performance for most corruption types,
and that the overall performance improvement is consistent for different image classification models.
DDA is the best on average, even DDA without self-ensembling improves on DiffPure with self-ensembling, which shows the generality of our method, since both a strong domain shift and a weak one can be covered by DDA.

\subsection{Ablation on Diffusion Models}

We investigate different choices for scaling factor $D$ and guidance scale $\boldsymbol{w}$  for latent guidance, related results of which are presented in Table~\ref{appendixTab:diffusionscalingfactor} and Table~\ref{appendixTab:diffusionrefinementrange}.
if $\boldsymbol{w}$ is too small, the guidance of the target image will not take effect. On the contrary, if $\boldsymbol{w}$ is too large, the shifting domain will bring a side effect.
The refinement range $D$ has the same effect as $\boldsymbol{w}$.
If $D$ is too small, the corruption on the target image will interfere the sampling process.
If the scaling factor $D$ is too large, the semantics information will be filtered by the low-pass filtering operation.
The results in Table~\ref{appendixTab:diffusionscalingfactor} and Table~\ref{appendixTab:diffusionrefinementrange} confirm the effect of the hyper-parameters $D$ and $\boldsymbol{w}$.

\input{tab/appendix_scalingfactor}
\input{tab/appendix_refinementrange}

\section{Tent, MEMO, BUFR, and DiffPure}

\subsection{Implementation}

Model adaptation, including Tent, MEMO, and BUFR, is sensitive to optimization hyper-parameter, especially learning rate and optimizer type. We also compare with the input adaptation Methods DiffPure~\cite{nie2022diffusion}  We introduce the hyper-parameter in this part for the four methods.

\paragraph{Tent}
\label{sec:tent_hyperparameters}
We augment the entropy loss $\mathcal{L}_{ent}$ from Tent~\cite{wang2021tent} with the additional diversity loss $\mathcal{L}_{div}$,
following the practice of SHOT~\cite{liang2020we}.
The test-time training objective is the linear combination of these two losses.
 $\mathcal{L}=\mathcal{L}_{ent} + \mathcal{L}_{div}$:
\begin{equation}
\begin{split}
    \mathcal{L}_{ent}&=-\Sigma_c p(\hat{y}_c)\log(p(\hat{y}_c) \\
    \mathcal{L}_{div}&=\kld{\hat{y}}{\frac{1}{C}\mathbf{1}_{C}} - \log(C) \\
\end{split}
\end{equation}
where $C$ is the number of classes.
$p(\hat{y}_c)$ denotes the c-th category probability in prediction $\hat{y}$.
$\mathbf{1}_{C}$ is an all-one vector with $C$ dimensions.
Therefore, $\frac{1}{C}\mathbf{1}_{C}$ indicates that every class has the same evenly distributed $\frac{1}{C}$ probabilities.
$D_{\text{KL}}$ is the notation of Kullback-Leibler divergence.

Since most recent architectures, such as ViT~\cite{dosovitskiy2021image},
do not have BatchNorm layers anymore,
we thus extend the training parameter to the whole parameter except the final classification layer.
As for ResNet-like backbones, such as ResNet-50~\cite{he2016deep},
we choose SGD as an optimizer with a learning rate $0.001$, momentum $0.9$, and weight decay $0.0001$.
As for Transformer-like backbones, such as Swin-T~\cite{liu2021swin} and ConvNeXt-T~\cite{liu2022convnet},
we choose AdamW as an optimizer with a learning rate of $0.00001$, weight decay $0.05$.
The reason behind the difference in optimizer is to follow the optimizer choice of the corresponding ImageNet training recipe.

Since the purpose of test-time adaptation is to equip the recognition model with a simple yet effective way to adapt itself,
we do not change the hyper-parameter for either single or mixed domain settings. Also, there is no prior knowledge of what test domain is before the model's deployment. We cannot choose the best hyper-parameter to accompany the test-time sampling policy, batch size, \etc.
We observe that model adaptation is highly sensitive to the choice of optimization hyper-parameters.
The ablation of learning rate and optimizer type could be found in Figure~\ref{fig:lr_sgd_adamw_single} and Figure~\ref{fig:lr_sgd_adamw_multi}.

\input{fig/lr_sgd_adamw}

\paragraph{MEMO}
Following the official repository of MEMO~\cite{zhang2021memo}, we choose SGD as the optimizer with a learning rate of $0.00025$. We have also explored the optimizer type and lr for different models and experiments show that the official setting is the best.

\paragraph{BUFR}
Since the  official repository of BUFR~\cite{eastwood2021source} did not conduct the experiments on ImageNet-C with ResNet-50, we choose SGD as the optimizer with a learning rate of $0.001$. The epochs per block is 3. Trained with ordered data, the average accuracy for BUFR is 2.8\% / 17.9\% accuracy when batch size is 1 / 64. When the class order is fixed, and the batch size is 64, BUFR achieved 4.2\% / 3.4\% with mixed/unmixed types. When the class order is shuffled, and the batch size is 64, BUFR achieved 4.3\% / 8.3\% with mixed/unmixed types. The results demonstrate that the limited, ordered, or mixed data does affect the training process and the classification accuracy.

\paragraph{DiffPure}
DiffPure~\cite{nie2022diffusion} simply adds a given amount of noise and then reverses to $t = 0$.
Following the official repository of DiffPure, we set the hyperparameter $t=150$

\subsection{Results}

\paragraph{Benchmark Evaluation (Independent Adaptation): Tent and DDA}
Table~\ref{tab:mCEImageNetCsingle} depicts the performance of Tent\cite{wang2021tent} and our DDA in 8 models.
The first models, ResNet50, Swin-T, and ConvNeXt-T, are already mentioned in Sec~\ref{sec:models}.
Here we additionally provide more experiments in much larger models, Swin transformer Base (Swin-B) and ConvNeXt Base (ConvNeXt-B).
It is worth noting that we provide two versions of base models: 1) trained with ImageNet-1K only (denote as Swin-B and ConvNeXt-B), 2) pretrained with ImageNet-21K first and then finetuned with ImageNet-1K (denote as Swin-B* and ConvNeXt-B*).
When the batch size equals one, DDA can beat Tent easily according to its advantage in tackling insufficient sampling.
When the categories of test images are not shuffled, DDA still has much better performance,
even with the larger state-of-the-art architectures.

\input{tab/mCEImageNetCsingle}

\input{tab/appendix_mCEImageNetCmulti}
\paragraph{Challenge Exploration (Joint Adaptation): Offline Tent}
In Table~\ref{tab:appendix_mCEImageNetCmulti}, we provide a more detailed comparison between episodic, online, and offline model adaptation performance.
In the first group of rows, we evaluate the episodic setting, in which adaptation and prediction are independent across inputs.
In particular, the episodic source-only, MEMO, DiffPure, and DDA methods do not depend on any data except for each input in isolation.
In the second group of rows we evaluate model adaptation by Tent in the online setting, in which adaptation and prediction are done per batch, and adaptation updates persist across batches.
(This is the setting reported in Table~\ref{tab:mCEImageNetCmulti} of the main paper, as it is the recommended setting for Tent.)
In the third group of rows, we evaluate model adaptation by Tent in the offline setting, in which the method first adapts to the entire test set, and then makes predictions for each input.
In this setting, Tent first learns from all of the data but then makes predictions with a single model that cannot specifically adapt to each batch.
Whether online or offline, Tent is sensitive to the order of the data, and fails when data arrive one by one (with a batch size of one) or when classes are not mixed by shuffling.

\input{fig/multi_domain_batch}
\paragraph{Challenge Exploration (Joint Adaptation): fixed and shuffled class order}
As shown in Figure~\ref{fig:shuffle_batchsize_multi_shuffle} and Figure~\ref{fig:shuffle_batchsize_multi_noshuffle},
the shuffled class order is essential to Tent, especially for models with low capacity.
Even for big models, the potential of Tent as the batch size increases is suppressed largely.

\section{More datasets}

\paragraph{ImageNet-W}

\input{fig/appendix_ImageNetW}

ImageNet-W~\cite{li2022whacamole} is an evaluation set based on ImageNet for the watermark. The authors found that the watermark as a shortcut affects nearly every modern vision model. In our experimental settings, the watermark can be regarded as a common corruption by human activities, as a supplement dataset of the unexpected corruptions, ImageNet-C (IN-C)~\cite{hendrycks2019robustness} and ImageNet-$\overline{\text{C}}$ (IN-$\bar{\text{C}}$)~\cite{mintun2021interaction}.
Figure~\ref{fig:appendix_imagenetw} shows that DDA can effectively remove the watermark to avoid shortcut reliance.
Our experiments, conducted on ImagenNet-W using ResNet-50, showed that DDA achieved 58.3\% accuracy, a significant improvement over the baseline accuracy of 47.4\%. Both the visualizations and model performance demonstrate that DDA can enhance the robustness of the model against watermark corruption.

\paragraph{ImageNet-R}

ImageNet-R~\cite{hendrycks2021many} is an evaluation set based on ImageNet for rendition, containing cartoons, embroidery, graphics, paintings, sketches, tattoos, toys, and so on.
Our qualitative experiments in Figure~\ref{fig:appendix_imagenetr} depict the performance of DDA on rendition to real images. Although the adapted images still reserve the original background, real-world characteristics have been added to the main part of the picture. DDA easily concentrates on the core domain shift in the renditions, especially for the gap between the 2D and 3D features.

\input{fig/appendix_ImageNetR}

\section{Visualization}

\paragraph{Progressive generation on ImageNet-C and ImageNet}

\input{fig/appendix_timestep}

Figure~\ref{appendixFig:timestep} illustrates how diffusion models denoise and reconstruct the given corrupted images.
Here we take elastic transformation, glass blur, and shot noise as the representative corruption for the digital artifact, natural blur, and common noise.
We observe that diffusion models are able to clean up the \say{local}, high-frequency noises, \ie Gaussian noise, pixelate, etc.
As for \say{global}, low-frequency corruption, \ie fog, snow, \etc, diffusion models failed to recover the original version.
One reason behind this phenomenon could be these low-frequency corrupted images are treated as natural samples during ImageNet training.
In other words, the diffusion model is trained on ImageNet, which may cover several augmentations including these low-frequency corruptions.

\input{fig/timestep_origin}

Figure~\ref{fig:timestep_origin} visualizes the procedure given the original ImageNet validation images.
We observe that images after diffusion are almost the same as the original ones.
In general, the reconstructions from diffusion models look similar to the original ImageNet validation images, which indicates the effectiveness of the leveraged generative models.
It is worth noting the comparison between the output and the original in the second row, peacock. Diffusion models hallucinate more details on the left that do not exist in the original input image.

\paragraph{Qualitative results on success and failure Cases}

\input{fig/corruption_positive}
\input{fig/corruption_negative}

We visualize success and failure cases across corruptions on ImageNet-C as shown in Figure~\ref{fig:corruption_positive} and Figure~\ref{fig:corruption_negative}.
DDA forces the model to preserve the global structural information to avoid semantic drift,
leading to a drawback that it may fail to adapt images from certain domains.
While DDA performs well when projecting most high-frequency/local corruptions (\eg, Gaussian noise, impulse noise, jpeg encoding, \ldots),
it fails for a few low-frequency/global corruptions (\eg, frost, fog, brightness adjustment, \ldots).
However, our self-ensembling scheme effectively detects these cases to avoid significant drops in accuracy.

%% file: tab/ablationImageNetC2in1.tex
\begin{table}[t]
  \caption{
  \textbf{Ablation on the design choices of selection module.} 
    We report accuracy on corruption benchmark ImageNet-C at severity level 5 (most severe). 
    Higher is better.
  }
  \label{tab:ablationImageNetC2in1}
  \begin{center}
    \adjustbox{max width=\linewidth}{
    \renewcommand\arraystretch{1.2}

    \begin{tabular}{l|ccccc}
      \toprule
      & ResNet-50
      & Swin-T
      & ConvNeXt-T
      & Swin-B
      & ConvNeXt-B
      \\
      \midrule
      corruption
      & 18.7  & 33.1  & 39.3  & 40.5  & 45.6
      \\
      diffusion
      & 28.4  & 34.6  & 39.3  & 38.6  & 42.8
      \\
      \midrule
      entropy
      & 29.7  & 39.7  & 43.9  & 43.9  & 49.2 
      \\
      confidence
      & 29.6  & 39.8  & 44.0  & 44.1  & 49.2
      \\
      \midrule
      entropy fuse
      & 23.8  & 38.2  & 42.8  & 44.0  & 48.0 
      \\
      confidence fuse
      & 23.8  & 38.2  & 42.8  & 44.0  & 48.0
      \\
      \midrule
      entropy sum
      & 29.7  & 40.0  & 44.2  & 44.5  & 49.4
      \\
      confidence sum
      & 29.7  & 39.9  & 44.2  & 44.4  & 49.4
      \\
      \rowcolor{lightgray}
      sum (ours)
      & 29.7  & 40.0  & 44.2  & 44.5  & 49.4
      \\
	  \midrule
      original test 
      & 76.6 
      & 81.2 
      & 82.1
      & 83.4
      & 83.9
      \\
      \bottomrule
    \end{tabular}}

  \end{center}
\end{table}

%% file: fig/one_out_of_two_appendix.tex
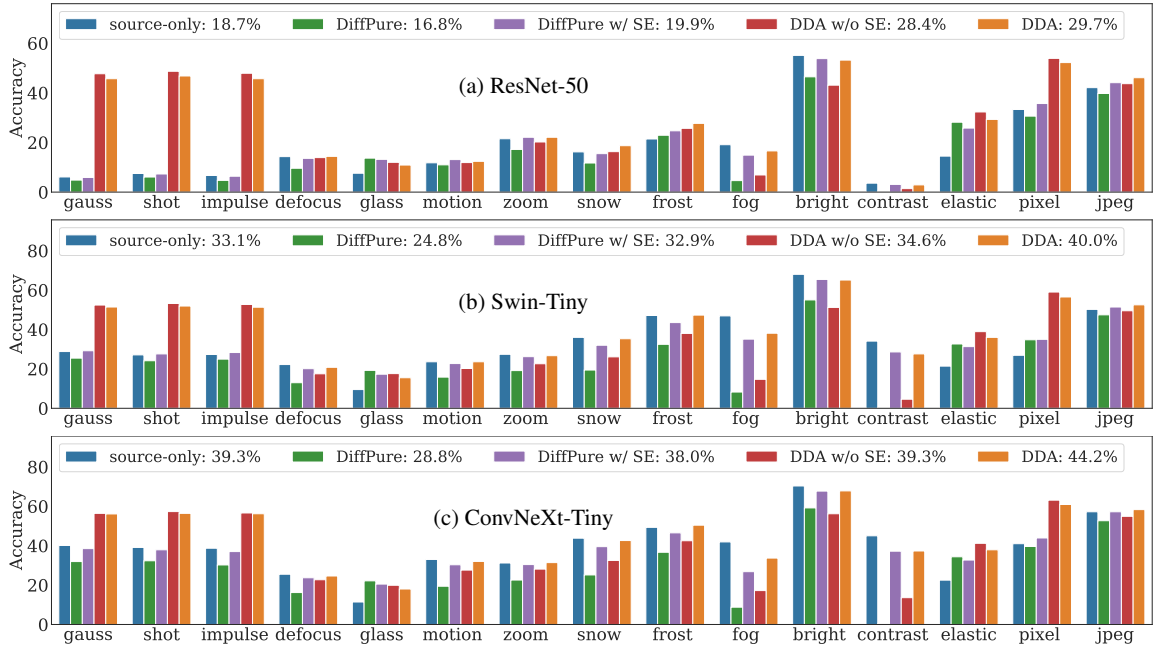
\begin{figure*}[t]
\vspace{-5pt}
\begin{center}
\adjustbox{max width=0.9\linewidth}{

    \begin{tabular}{c}
      \begin{tikzpicture}
        \node [
          above right,
          inner sep=0
        ] (image) at (0,0) {\includegraphics[width=\textwidth]{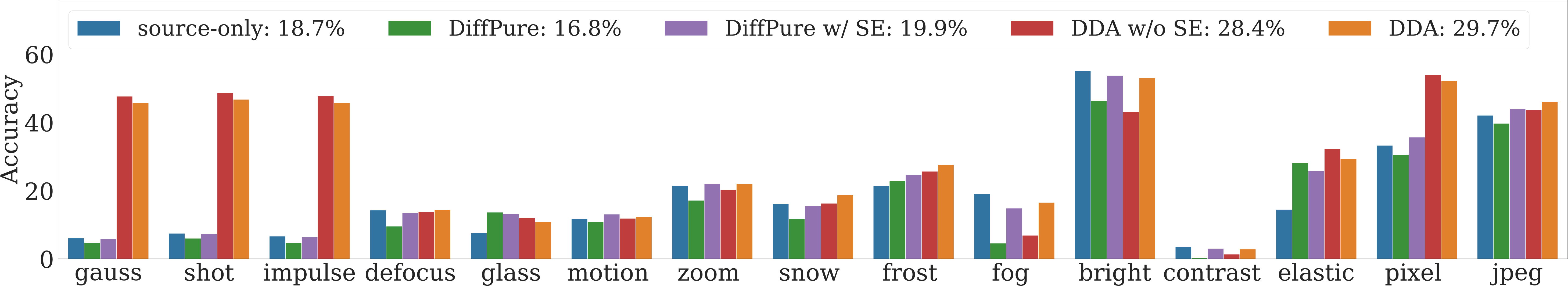}};

        \begin{scope}[
        x={($0.1*(image.south east)$)},
        y={($0.1*(image.north west)$)}]

        \node[above,black] at (4.5,5.1) {\normalsize (a) ResNet-50};

        \end{scope}
      \end{tikzpicture} \\
      \begin{tikzpicture}
        \node [
          above right,
          inner sep=0
        ] (image) at (0,0) {\includegraphics[width=\textwidth]{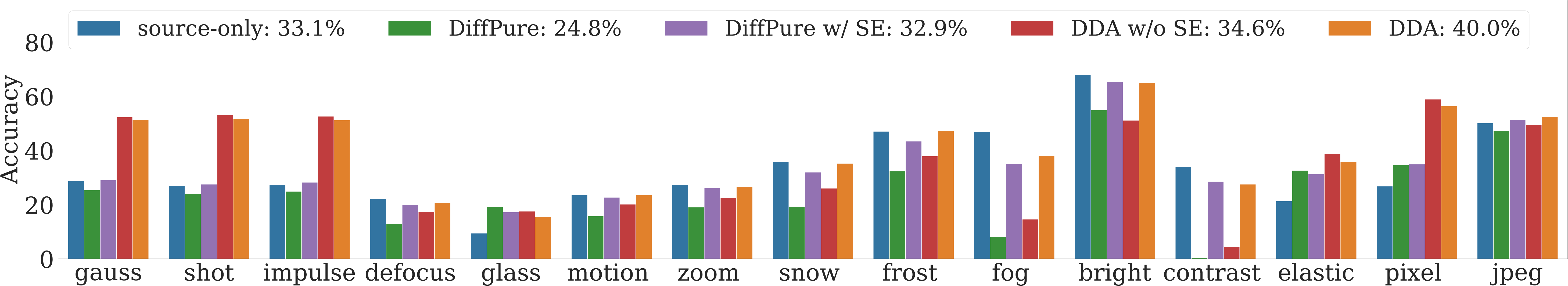}};

        \begin{scope}[
        x={($0.1*(image.south east)$)},
        y={($0.1*(image.north west)$)}]

        \node[above,black] at (4.5,5.1) {\normalsize (b) Swin-Tiny};

        \end{scope}
      \end{tikzpicture} \\
      \begin{tikzpicture}
        \node [
          above right,
          inner sep=0
        ] (image) at (0,0) {\includegraphics[width=\textwidth]{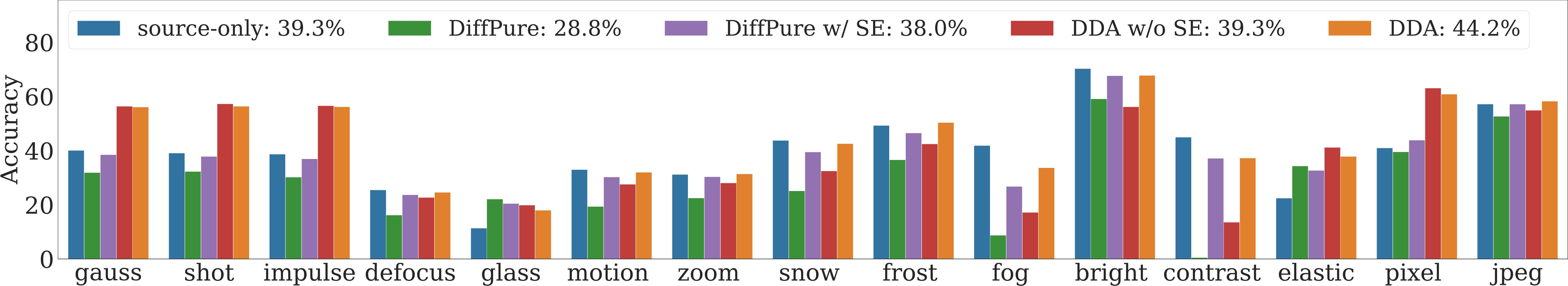}};

        \begin{scope}[
        x={($0.1*(image.south east)$)},
        y={($0.1*(image.north west)$)}]

        \node[above,black] at (4.5,5.1) {\normalsize (c) ConvNeXt-Tiny};

        \end{scope}
      \end{tikzpicture} \\
  \end{tabular}
}
\end{center}
\vspace{-4mm}
\caption{%
    \textbf{Self-Ensembling reliably improves robustness across corruption types.}
    We evaluate the self-ensembling among DDA and state-of-the-art diffusion for adversarial defense (DiffPure). Self-Ensembling prevents catastrophic drops (on fog or contrast, for example) and improves the performance on DDA and DiffPure. DDA is the best on average, even DDA without self-ensembling improves on DiffPure with self-ensembling.
}

\label{fig:one_out_of_two_appendix}
\end{figure*}

%% file: tab/appendix_scalingfactor.tex
\begin{table}[t]
  \caption{
	\textbf{Ablation on choices of scaling factor $D$ with fixed $\boldsymbol{w}$.} 
	Our default hyper-parameter is in the shadow row.
	The empirical results confirm the choice of hyper-parameters in DDA.
  }
  \label{appendixTab:diffusionscalingfactor}
  \begin{center}
   		\adjustbox{max width=\linewidth}{
   			\renewcommand\arraystretch{1.15}
   			\begin{tabular}{ll|ccc}
   				& $\boldsymbol{w}=6$
   				& ResNet-50
   				& Swin-T
   				& ConvNeXt-T
   				\\
   				\midrule
   				& 2
                    & 22.2    & 29.9     & 31.2
   				\\
   				\rowcolor{lightgray} (a) elastic transformation & 4
                    & 32.3    & 38.9     & 41.2
   				\\
   				& 8
                   & 41.6     & 45.2  & 48.0
   				\\
   				\midrule
   				& 2
                    & 9.5      & 13.4  & 14.9
   				\\
   				\rowcolor{lightgray} (b) glass blur & 4
                    & 12.0     & 17.6  & 19.9
   				\\
   				& 8
                    & 19.7     & 26.0  & 29.8  
   				\\
   				\midrule
   				& 2
                    & 50.4     & 54.5  & 58.7
   				\\
   				\rowcolor{lightgray} (c) shot noise & 4
                    & 48.7     & 53.2  & 57.3
   				\\
   				& 8
                    & 40.4     & 44.6  & 48.7 
   				\\
   				\bottomrule
   			\end{tabular}
   		}
    \end{center}
\end{table}

%% file: tab/appendix_refinementrange.tex
\begin{table}[t]
  \caption{
	\textbf{Ablation on choices of refinement range $\boldsymbol{w}$ with fixed $D$.} 
	Our default hyper-parameter is in the shadow row.
	The empirical results confirm the choice of hyper-parameters in DDA.
  }
  \label{appendixTab:diffusionrefinementrange}
  \begin{center}
   		\adjustbox{max width=\linewidth}{
   			\renewcommand\arraystretch{1.15}
   			\begin{tabular}{ll|ccc}
   				& $D=4$
   				& ResNet-50
   				& Swin-T
   				& ConvNeXt-T
   				\\
   				\midrule
   				& 3
                    & 38.0     & 42.3  & 44.9
   				\\
   				\rowcolor{lightgray} (a) elastic transformation & 6
                    & 32.3     & 38.9  & 41.2
   				\\
   				& 9
                    & 27.8     & 35.2  & 37.9
   				\\
   				\midrule
   				& 3
                    & 17.9     & 24.7  & 28.7
   				\\
   				\rowcolor{lightgray} (b) glass blur & 6
                    & 12.0     & 17.6  & 19.9
   				\\
   				& 9
                    & 9.2      & 13.4  & 15.2
   				\\
   				\midrule
   				& 3
                    & 48.2     & 50.5  & 54.1
   				\\
   				\rowcolor{lightgray} (c) shot noise & 6
                    & 48.7     & 53.2  & 57.3   
   				\\
   				& 9
                    & 40.7     & 49.1  & 54.4
   				\\
   				\bottomrule
   			\end{tabular}
   		}
    \end{center}
\end{table}

%% file: fig/lr_sgd_adamw.tex
\begin{figure*}[t]
    \begin{center}
    	\adjustbox{max width=0.95\linewidth}{
            \begin{tabular}{c c c c}
                \resizebox{!}{0.4in}{\hspace{3.3mm} (a) ResNet-50} & 
                \resizebox{!}{0.4in}{\hspace{3.3mm} (b) Swin-Tiny} & 
                \resizebox{!}{0.4in}{\hspace{3.3mm} (c) ConvNeXt-Tiny} \\
                & & \\
                & & \\
                \includegraphics[width=\linewidth]{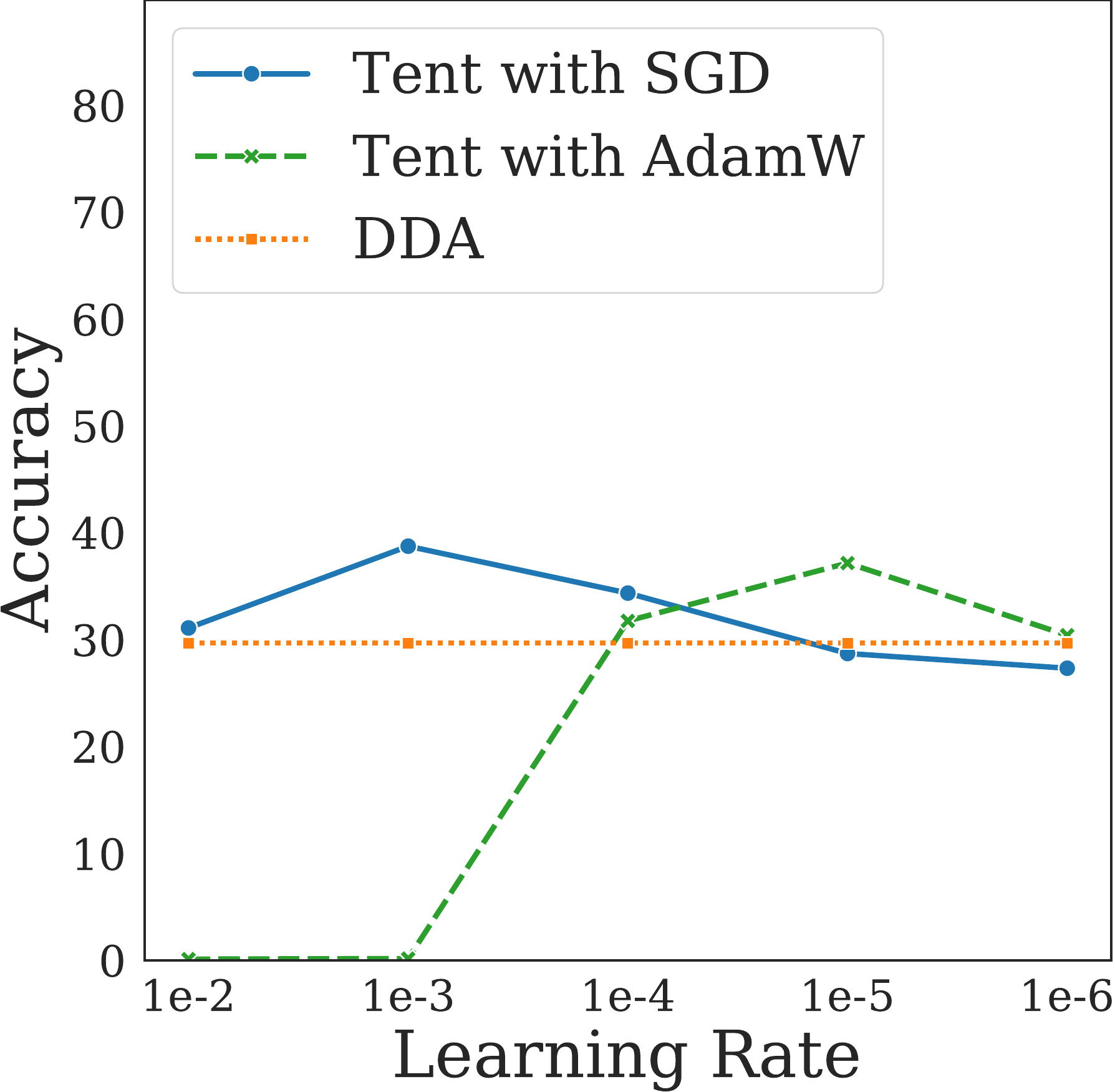} &
                \includegraphics[width=\linewidth]{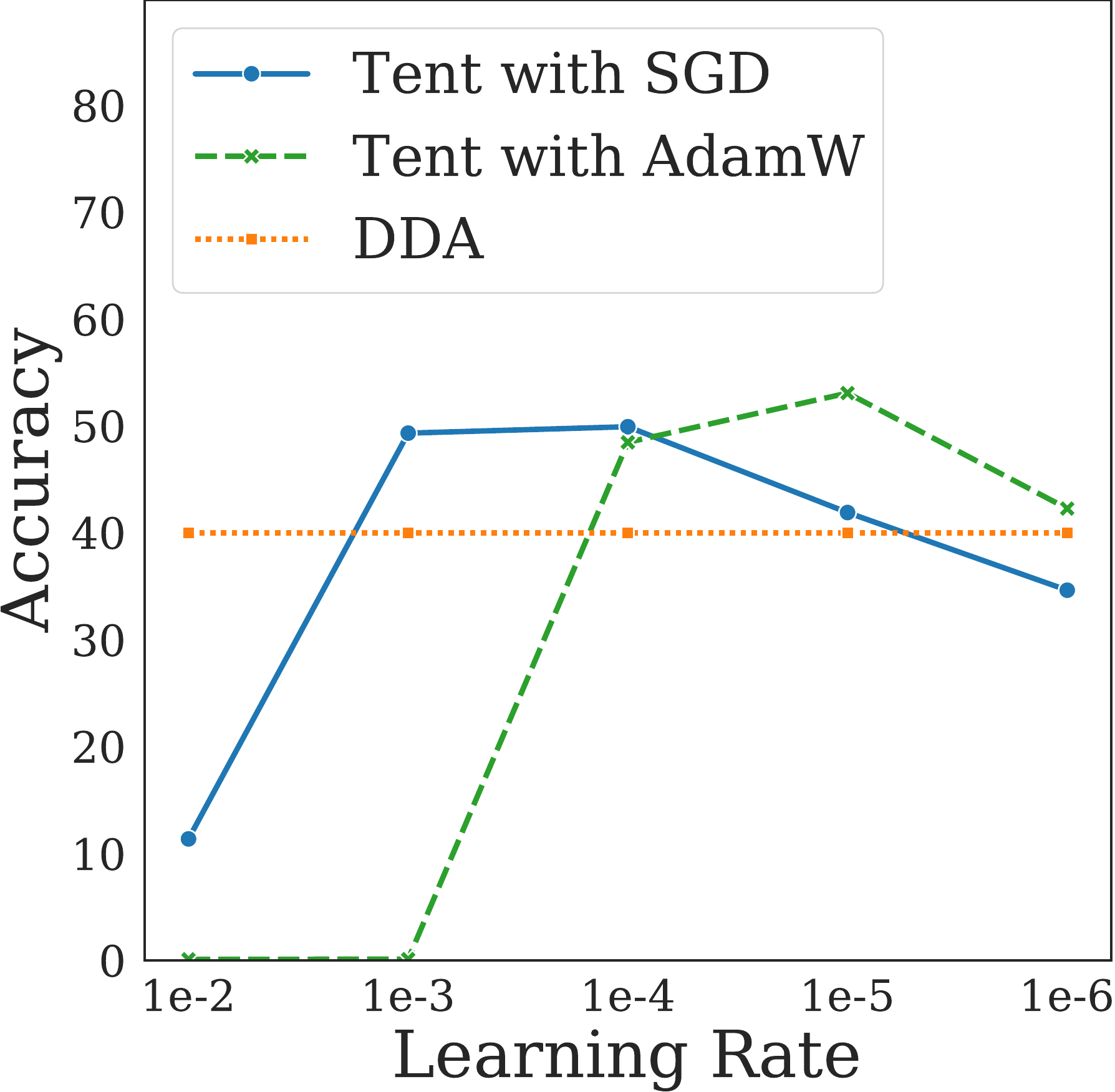} &
                \includegraphics[width=\linewidth]{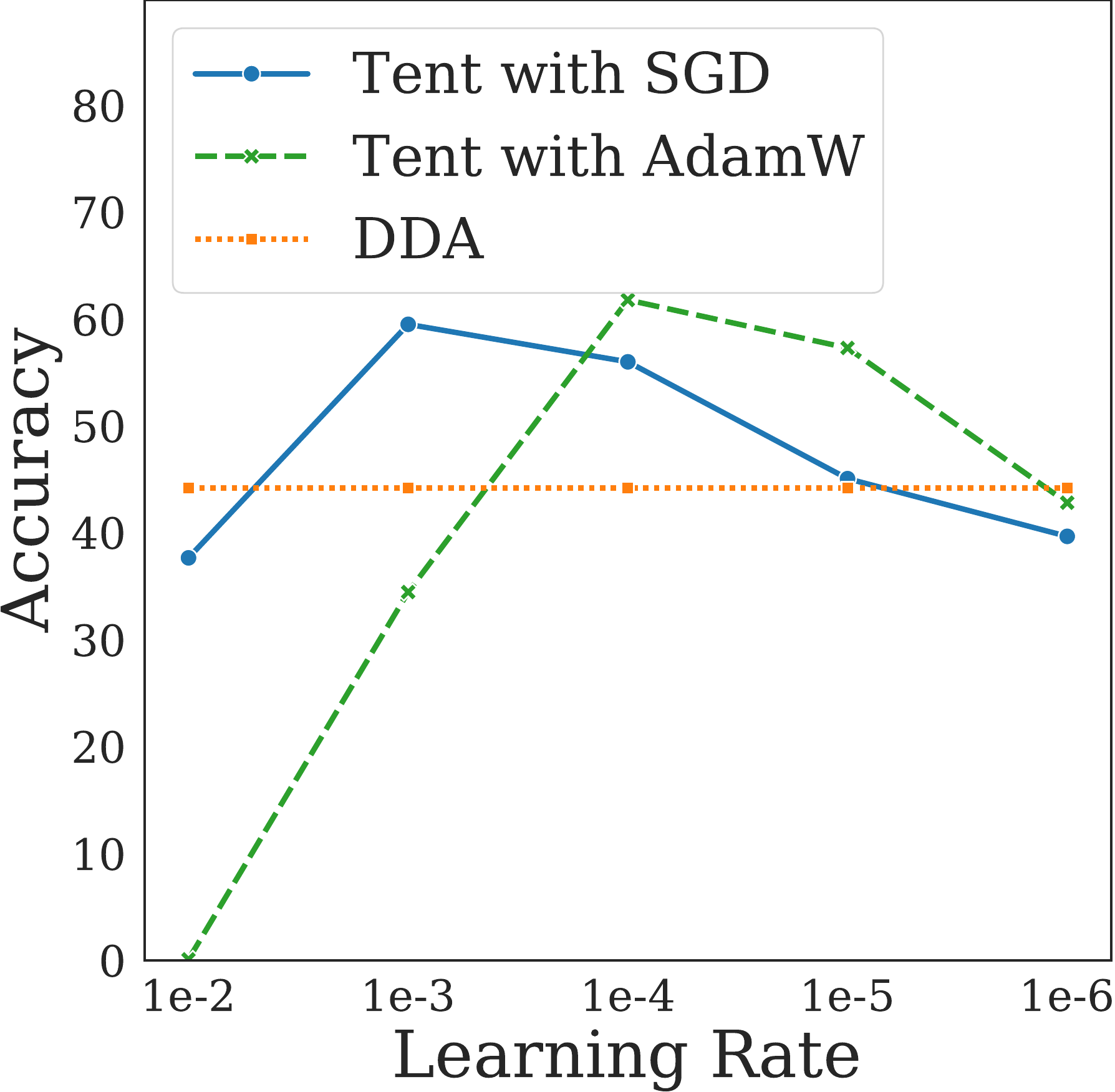} \\
            \end{tabular}
    	}
	\end{center}
	\caption{
        \textbf{DDA is invariant to learning rate while Tent is extremely sensitive on benchmark evaluation (independent adaptation).}
        To analyze sensitivity to the learning rate we measure the average robustness of independent adaptation across corruption types.
        DDA does not depend on these factors while Tent fails without the proper tuning on learning rate.
    }
	\label{fig:lr_sgd_adamw_single}
\end{figure*}

\begin{figure*}[t]
    \begin{center}
    	\adjustbox{max width=0.95\linewidth}{
            \begin{tabular}{c c c c}
                \resizebox{!}{0.4in}{\hspace{3.3mm} (a) ResNet-50} & 
                \resizebox{!}{0.4in}{\hspace{3.3mm} (b) Swin-Tiny} & 
                \resizebox{!}{0.4in}{\hspace{3.3mm} (c) ConvNeXt-Tiny} \\
                & & \\
                & & \\
                \includegraphics[width=\linewidth]{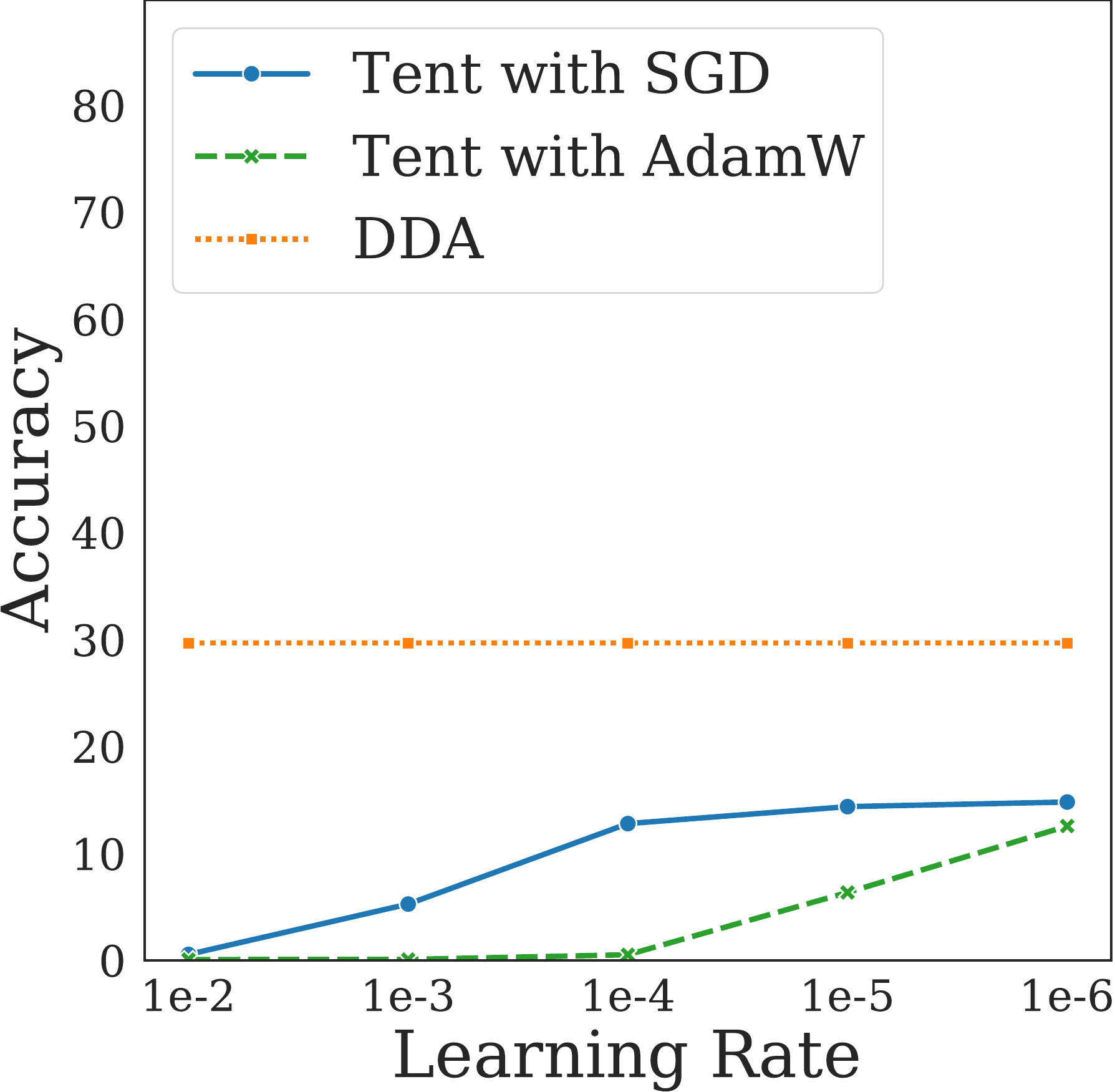} &
                \includegraphics[width=\linewidth]{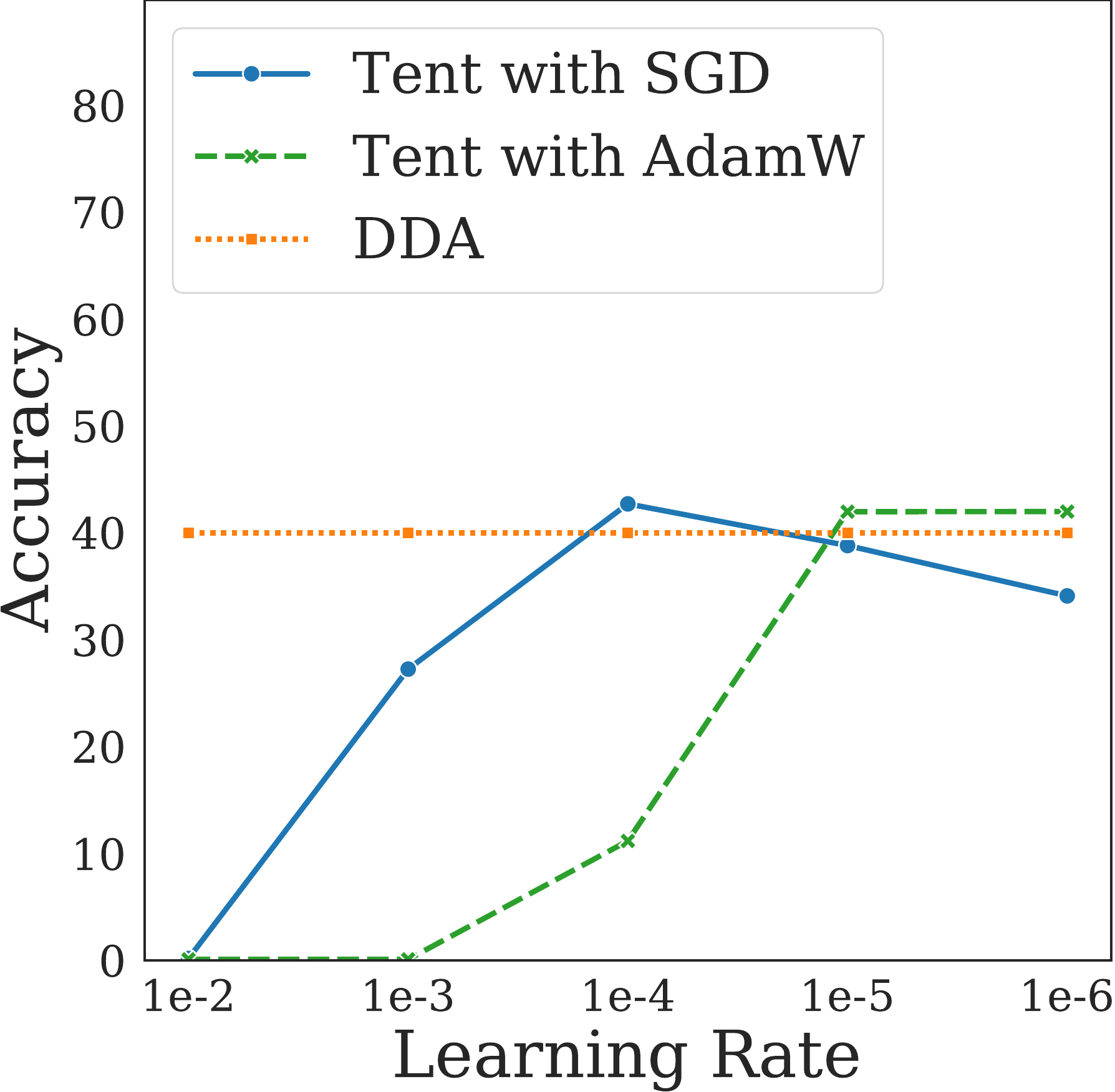} &
                \includegraphics[width=\linewidth]{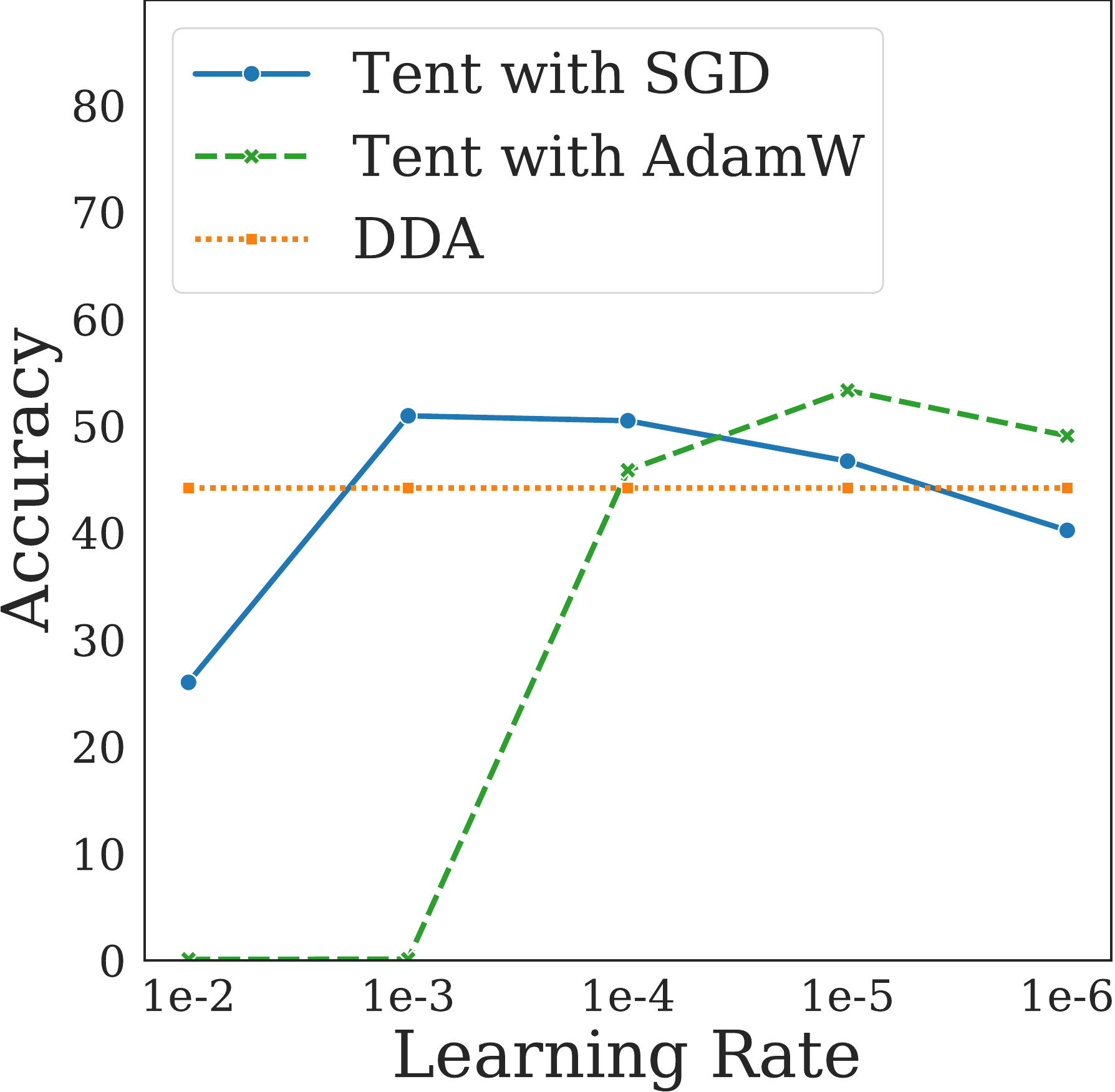} \\
            \end{tabular}
    	}
	\end{center}
	\caption{
        \textbf{DDA is invariant to learning rate while Tent is extremely sensitive on challenge exploration (joint adaptation).}
    }
	\label{fig:lr_sgd_adamw_multi}
\end{figure*}

%% file: tab/mCEImageNetCsingle.tex
\begin{table}[t]
\caption{
\textbf{DDA is reliably more robust on benchmark evaluation (independent adaptation) with fixed and shuffled class order.}
Deployment may supply target data in various ways.
To explore these regimes, we vary batch size and whether or not the data is ordered by class.
We compare episodic adaptation by input updates with DDA (ours) and source-only test baseline against cumulative adaptation with Tent.
DDA and source-only are invariant to these differences in the data.
However, Tent is highly sensitive to batch size and order, and fails in the more natural data regimes.
}
\label{tab:mCEImageNetCsingle}
\begin{center}
\adjustbox{max width=\linewidth}{
\renewcommand\arraystretch{1.2}
\begin{tabular}{c|c|c|cc|cc|cc}
  \toprule
  Class Order & Batch Size 
  & ResNet-50 &
  Swin-T & ConvNeXt-T &
  Swin-B & ConvNeXt-B &
  Swin-B* & ConvNeXt-B* \\
  \midrule
  \multirow{2}{*}{\xmark} & 1 
  & 9.8 & 25.3 & 38.2 & 34.9 & 34.5 & 44.3 & 46.3 \\
   & 64 
   & 10.4 & 7.6 & 30.5 & 7.3 & 34.5 & 18.8 & 44.3 \\
  \midrule
  \multirow{2}{*}{\cmark} & 1 
  & 11.4 & 25.6 & 35.6 & 34.5 & 45.8 & 43.9 & 46.3 \\
   & 64 
   & 37.8 & 50.6 & 54.3 & 57.5 & 59.1  & 64.3 & 67.7 \\
  \midrule
  Source-Only & \multirow{2}{*}{N/A} 
  & 18.7 & 33.1 & 39.3 & 40.5 & 45.6 & 51.6 & 51.7 \\
  DDA (ours) & 
  & 29.7 & 40.0 & 44.2 & 44.5 & 49.4 & 54.5 & 55.4 \\
  \bottomrule
\end{tabular}
}
\end{center}
\end{table}

%% file: tab/appendix_mCEImageNetCmulti.tex
\begin{table}
  \caption{%
    \textbf{DDA is reliably more robust when the target data is limited, ordered, or mixed.}
    Deployment may supply target data in various ways.
    To explore these regimes, we vary batch size and whether or not the data is ordered by class or mixed across corruption types.
    We compare episodic adaptation by input updates with DDA (ours) and by model updates with MEMO along with cumulative adaptation by Tent.
    DDA and MEMO are invariant to these differences.
    However, Tent is highly sensitive to batch size and order, and fails on ordered classes and mixed types.
  }
  \label{tab:appendix_mCEImageNetCmulti}
  \begin{center}
  \adjustbox{max width=\linewidth}{
  \huge
  \renewcommand\arraystretch{1.2}
    \begin{tabular}{c|cc|c|c|cc}
      \toprule
      Method &
      \pbox{3em}{Mixed \\ Classes} &
      \pbox{6em}{Mixed \\ Types} &
      Batch Size &
      ResNet-50 &
      Swin-T &
      ConvNeXt-T \\
      \midrule
      Source-Only & \multicolumn{2}{c|}{\multirow{4}{*}{N/A}} & \multirow{4}{*}{N/A} 
      & 18.7 & 33.1 & 39.3 \\ 
      MEMO~\cite{zhang2021memo} & & & 
      & 24.7 & 29.5 & 37.8 \\ 
      DiffPure~\cite{nie2022diffusion} & & &  
      & 16.8 & 24.8 & 28.8 \\
      \rowcolor{lightgray}
      DDA (ours) & & & 
      & {\bf 29.7} & {\bf 40.0} & {\bf 44.2} \\ 
      \midrule
      \multirow{4}{*}{Tent (Online)} & \xmark & \xmark & 1 / 64 
      & 0.1  / 0.4  & 2.8  / 2.3  & 10.5  / 9.6 \\
      & \xmark & \cmark & 1 / 64 
      & 0.1  / 0.3  & 8.0  / 2.2  & 18.8  / 6.5 \\
      & \cmark & \xmark & 1 / 64 
      & 0.1  / \bf{22.6}  & 3.0  / \bf{41.0}  & 11.0  / \bf{50.1} \\
      & \cmark & \cmark & 1 / 64 
      & 0.1  / 6.5  & 8.5  / 36.9  & 18.9  / 47.4  \\ 
      \midrule
      \multirow{4}{*}{Tent (Offline)} & \xmark & \xmark & 1 / 64 
      & 2.2 / 0.4 & 0.2 / 0.2 & 0.1 / 1.4 \\ 
      & \xmark & \cmark & 1 / 64 
      & 1.6 / 0.5 & 0.2 / 0.5 & 0.3 / 0.5 \\ 
      & \cmark & \xmark & 1 / 64 
      & 3.0 / {\bf 7.6} & 0.1 / 43.3 & 0.2 / 48.8 \\ 
      & \cmark & \cmark & 1 / 64 
      & 2.3 / 3.9 & 0.3 / {\bf 44.1} & 0.3 / {\bf 51.9} \\ 
      \midrule
      \bottomrule
  \end{tabular}}
  \end{center}
\end{table}

%% file: fig/multi_domain_batch.tex
\begin{figure*}[t]
	\begin{center}
		\adjustbox{max width=0.95\linewidth}{
			\begin{tabular}{c c c}
				\resizebox{!}{0.4in}{\hspace{3.3mm} (a) ResNet-50} & 
				\resizebox{!}{0.4in}{\hspace{3.3mm} (b) Swin-Tiny} & 
				\resizebox{!}{0.4in}{\hspace{3.3mm} (c) ConvNeXt-Tiny} \\
				& & \\
				\includegraphics[width=\linewidth]{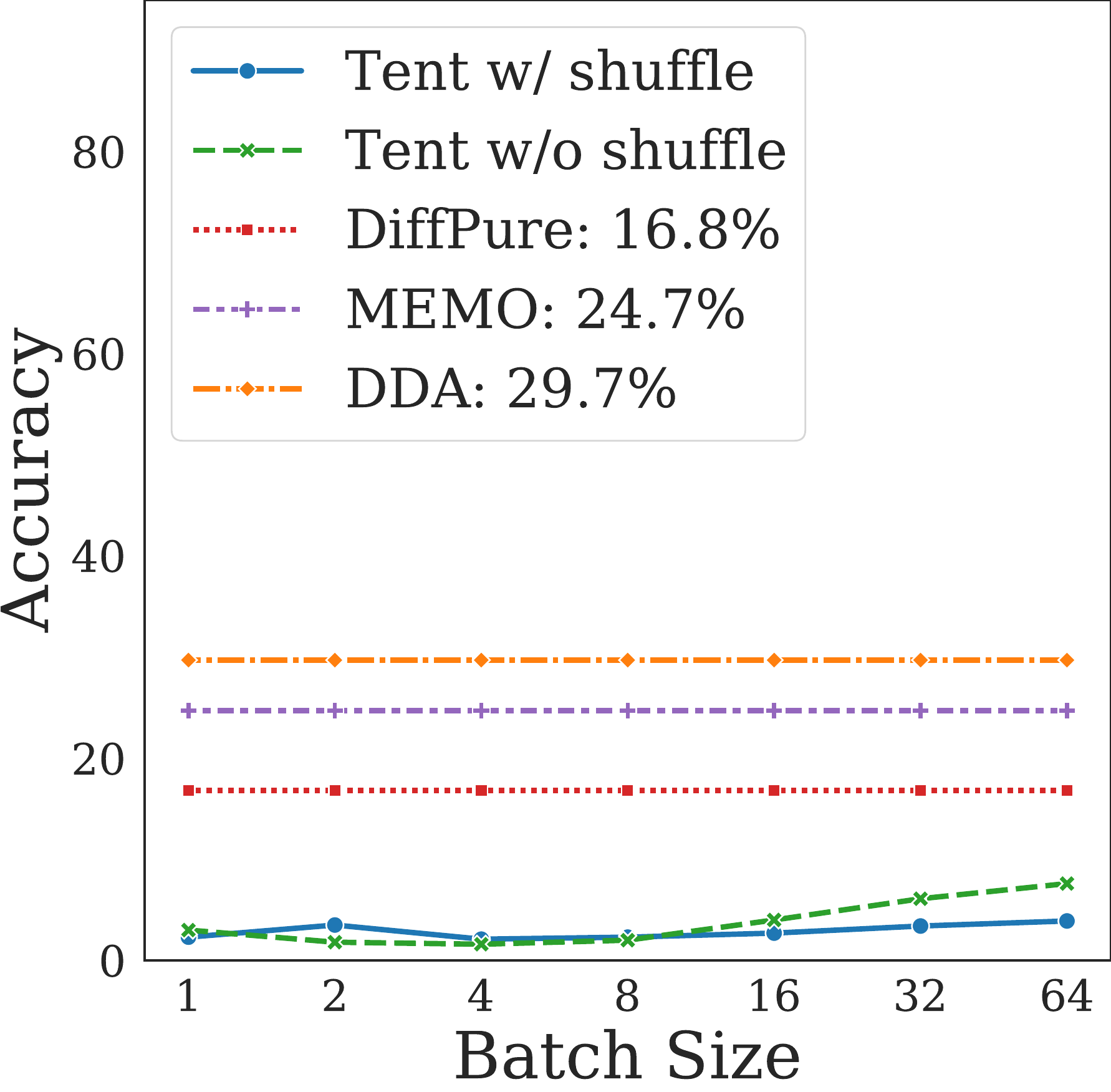} &
				\includegraphics[width=\linewidth]{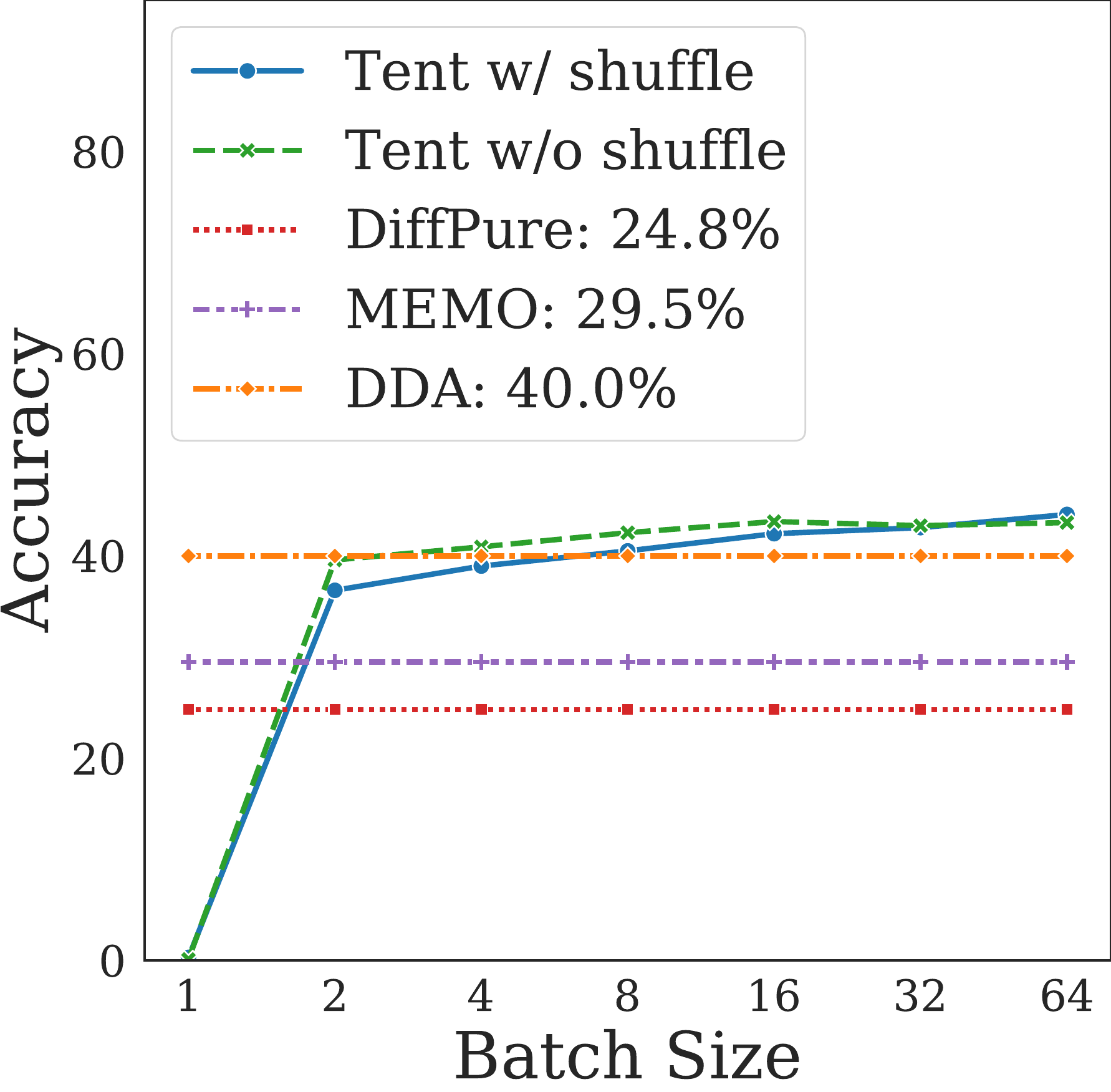} &
				\includegraphics[width=\linewidth]{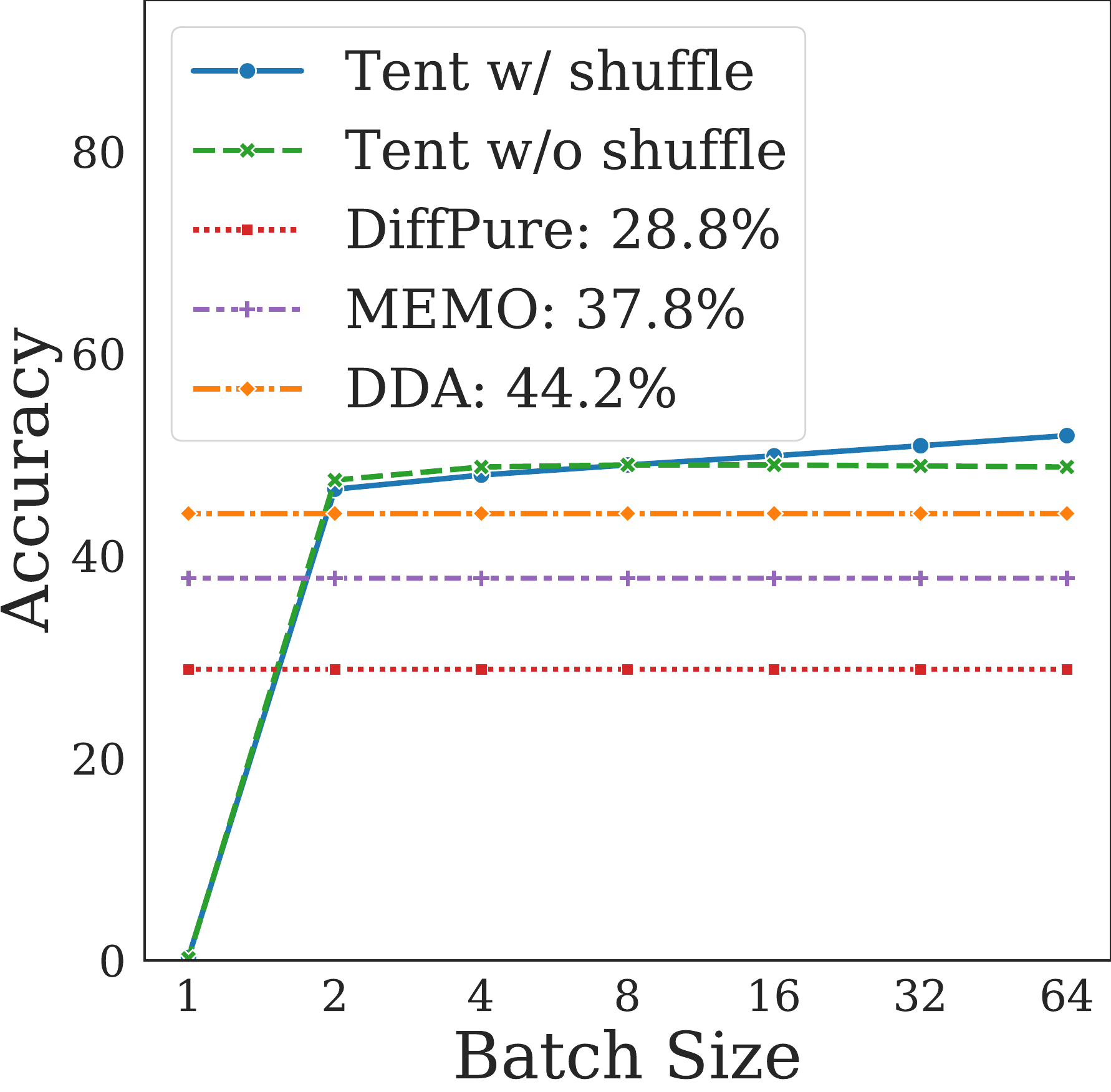} \\               
			\end{tabular}
		}
	\end{center}
	\caption{
		\textbf{DDA is reliably more robust on challenge exploration (joint adaptation) with shuffled class order.}
		Deployment may supply target data in various ways.
		To explore these regimes, we vary the batch size and whether or not the data is ordered by class or mixed across corruption types.
		We compare episodic adaptation by input updates with DDA (ours), DiffPure, and by model updates with MEMO against cumulative adaptation with offline and online Tent.
		DDA, DiffPure, and MEMO are invariant to these differences in the data.
		However, Tent is highly sensitive to batch size and order, and fails in the more natural data regimes.
	}
	\label{fig:shuffle_batchsize_multi_shuffle}
\end{figure*}

\begin{figure*}[t]
	\begin{center}
		\adjustbox{max width=0.95\linewidth}{
			\begin{tabular}{c c c}
				\resizebox{!}{0.4in}{\hspace{3.3mm} (a) ResNet-50} & 
				\resizebox{!}{0.4in}{\hspace{3.3mm} (b) Swin-Tiny} & 
				\resizebox{!}{0.4in}{\hspace{3.3mm} (c) ConvNeXt-Tiny} \\
				& & \\
				\includegraphics[width=\linewidth]{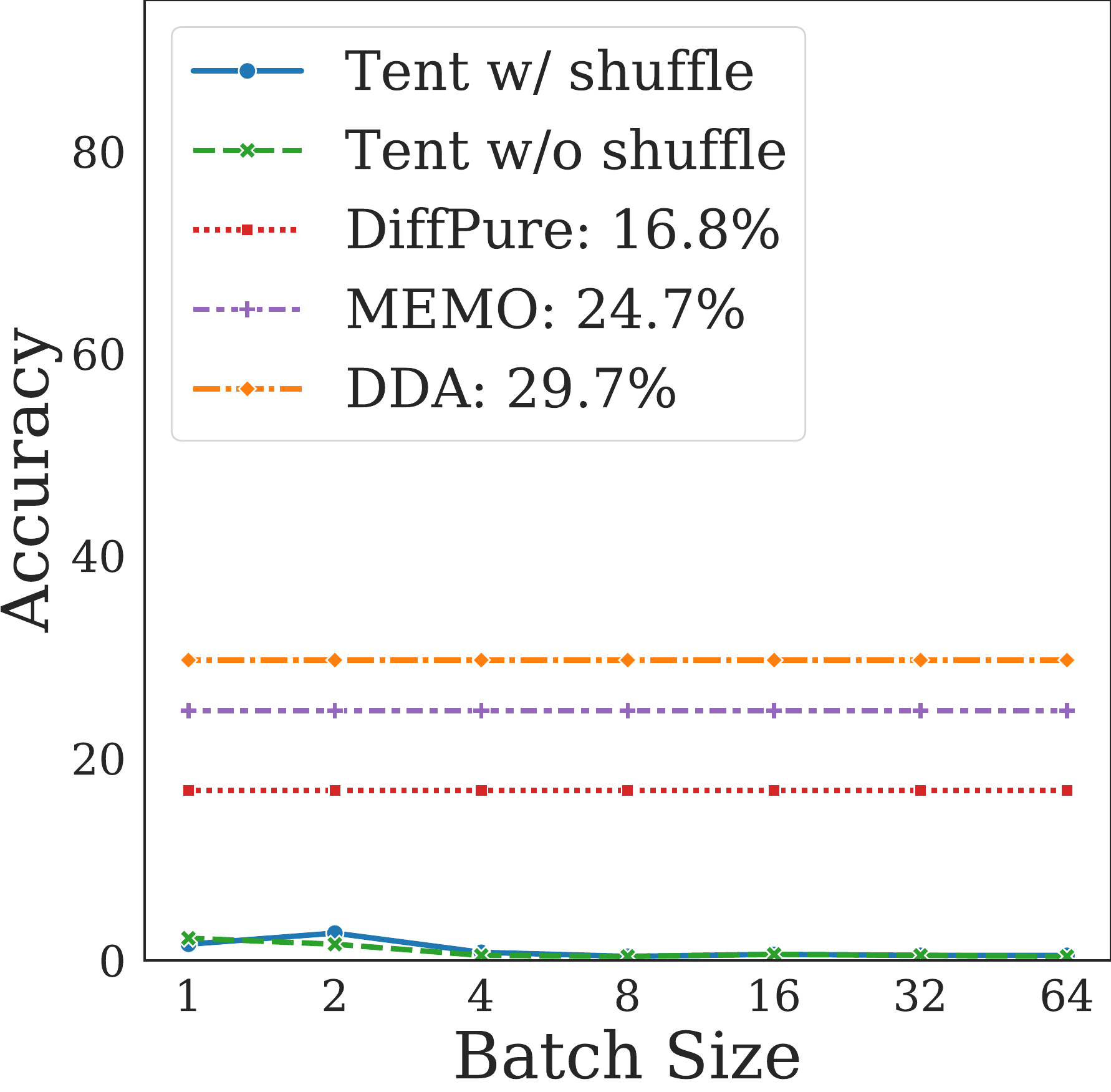} &
				\includegraphics[width=\linewidth]{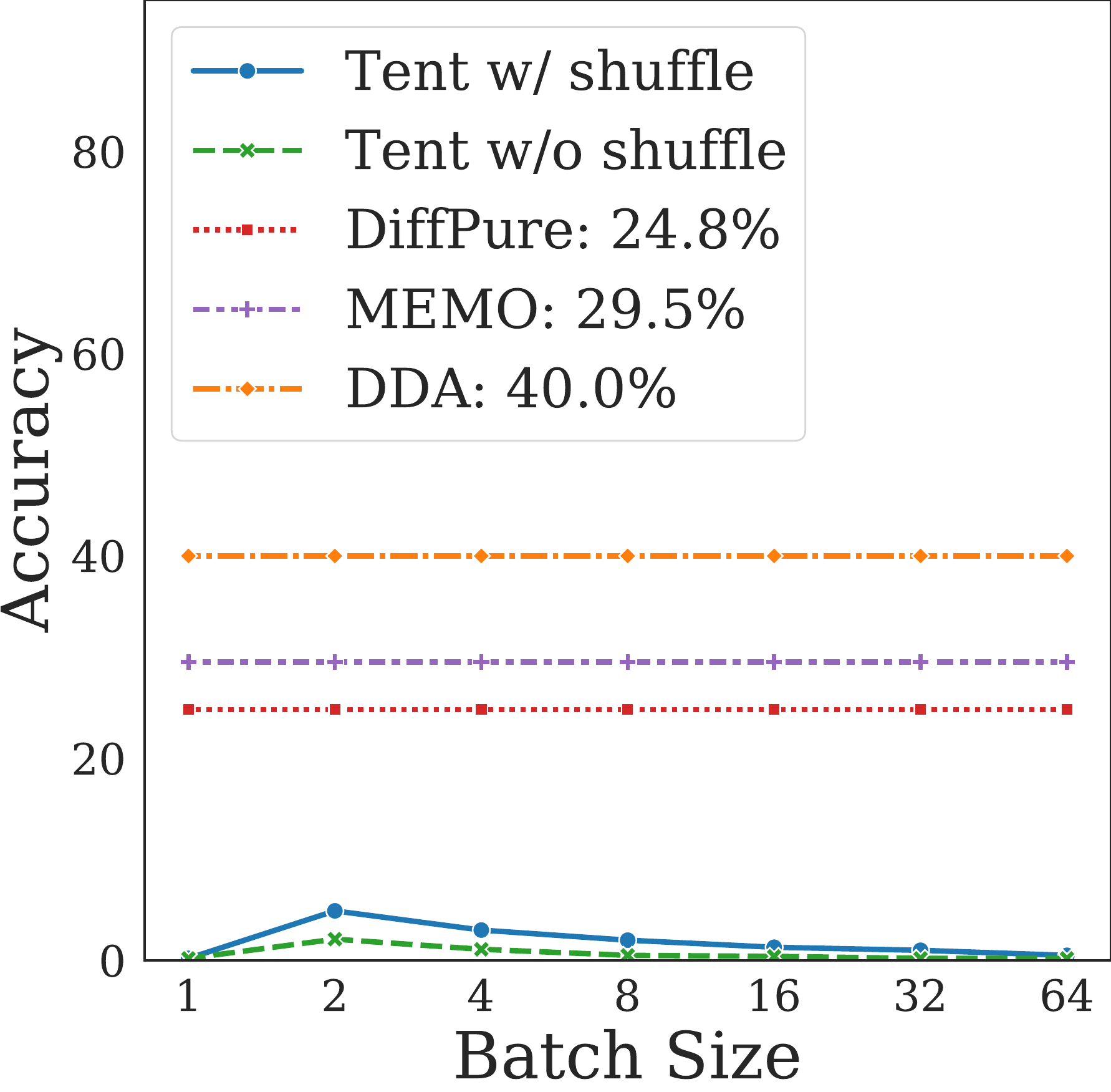} &
				\includegraphics[width=\linewidth]{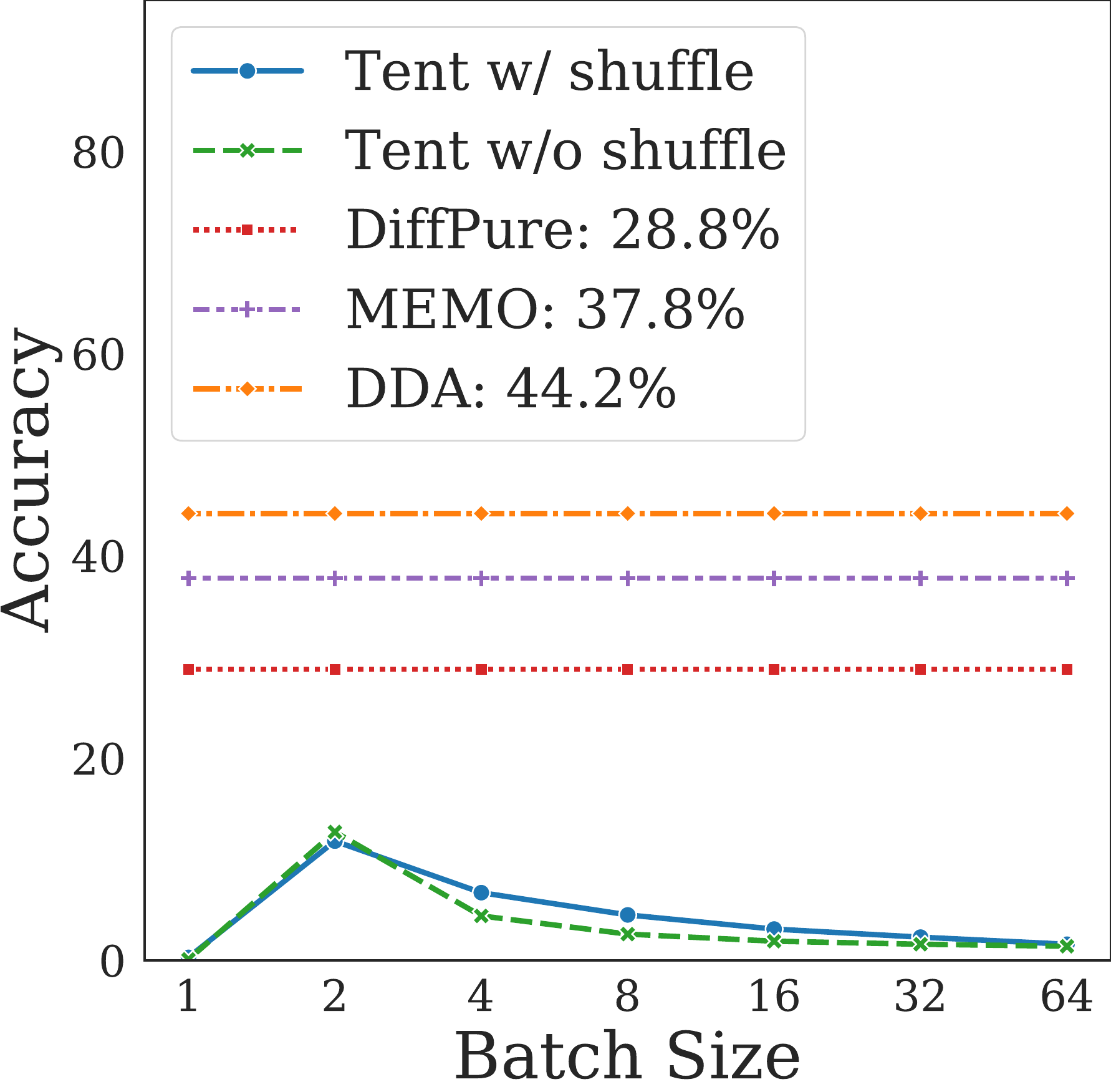} \\
			\end{tabular}
		}
	\end{center}
	\caption{
		\textbf{DDA is reliably more robust on challenge exploration (joint adaptation) with fixed class order.}
	}
	\label{fig:shuffle_batchsize_multi_noshuffle}
\end{figure*}

%% file: fig/appendix_ImageNetW.tex
\begin{figure*}[htbp]
    \begin{center}
    	\adjustbox{max width=\linewidth}{
            \begin{tabular}{c c c c c c c c c}
               
                \resizebox{0.75in}{!}{\rotatebox{90}{ImageNet-W}} &
                \includegraphics[width=\linewidth]{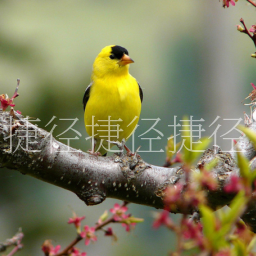} &
                \includegraphics[width=\linewidth]{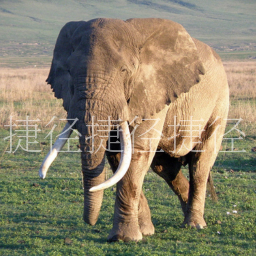} &
                \includegraphics[width=\linewidth]{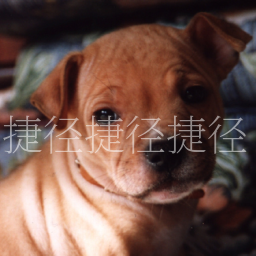} &
                \includegraphics[width=\linewidth]{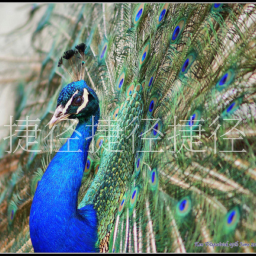} &
                \includegraphics[width=\linewidth]{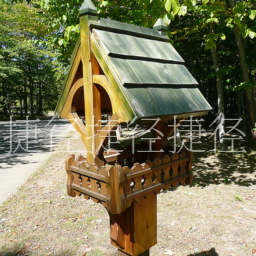} &
                \includegraphics[width=\linewidth]{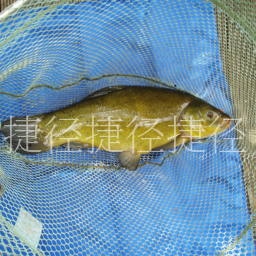} &
                \includegraphics[width=\linewidth]{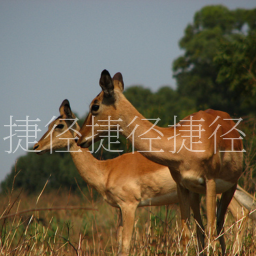} &
                \includegraphics[width=\linewidth]{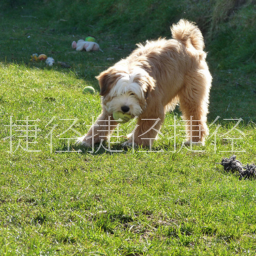} \\

                & & & & & & & \\
                \resizebox{0.75in}{!}{\rotatebox{90}{DDA Output}} &
                \includegraphics[width=\linewidth]{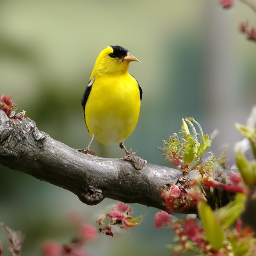} &
                \includegraphics[width=\linewidth]{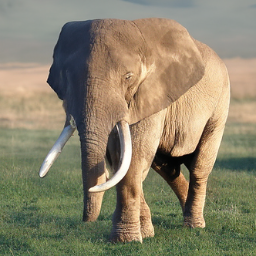} &
                \includegraphics[width=\linewidth]{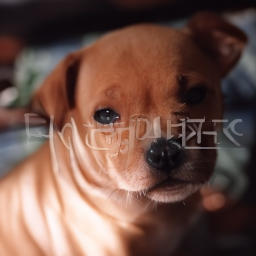} &
                \includegraphics[width=\linewidth]{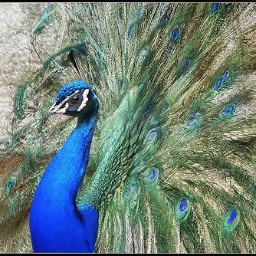} &
                \includegraphics[width=\linewidth]{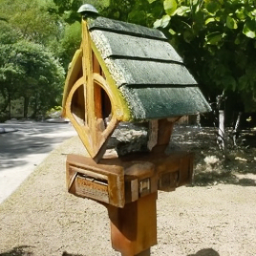} &
                \includegraphics[width=\linewidth]{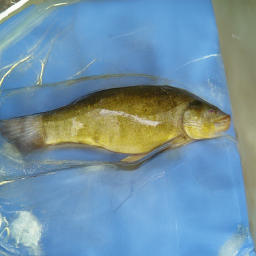} &
                \includegraphics[width=\linewidth]{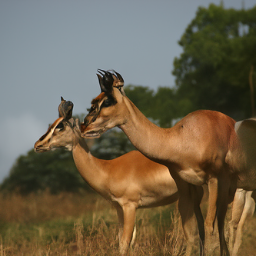} &
                \includegraphics[width=\linewidth]{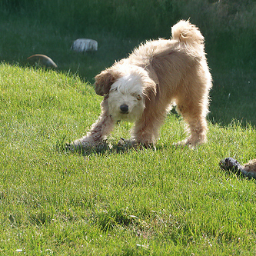} \\

            \end{tabular}
    	}

	\end{center}
	\caption{Visualization of generated images with DDA output on ImageNet-W.}
	\label{fig:appendix_imagenetw}
\end{figure*}

%% file: fig/appendix_ImageNetR.tex
\begin{figure*}[htbp]
    \begin{center}
    	\adjustbox{max width=\linewidth}{
            \begin{tabular}{c c c c c c c c c}
                & \resizebox{!}{0.6in}{cartoon} & \resizebox{!}{0.7in}{misc} & \resizebox{!}{0.7in}{painting} & \resizebox{!}{0.7in}{sketch} & \resizebox{!}{0.7in}{videogame} & \resizebox{!}{0.7in}{embroidery} & \resizebox{!}{0.6in}{tattoo} & \resizebox{!}{0.7in}{graphic} \\
                & & & & & & & & \\
                \resizebox{0.75in}{!}{\rotatebox{90}{ImageNet-R}} &
                \includegraphics[width=\linewidth]{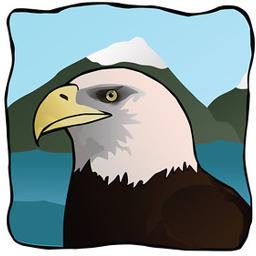} &
                \includegraphics[width=\linewidth]{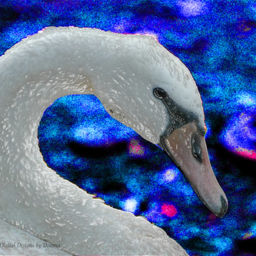} &
                \includegraphics[width=\linewidth]{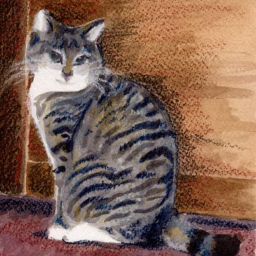} &
                \includegraphics[width=\linewidth]{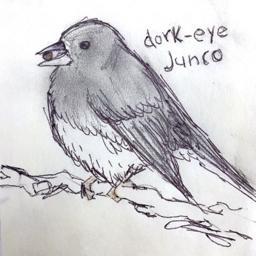} &
                \includegraphics[width=\linewidth]{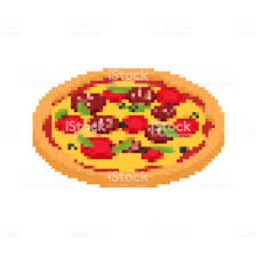} &
                \includegraphics[width=\linewidth]{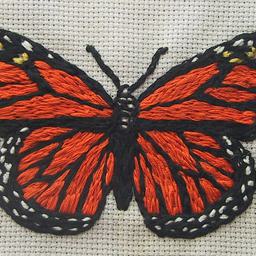} &
                \includegraphics[width=\linewidth]{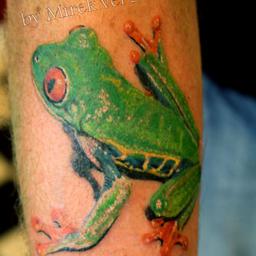} &
                \includegraphics[width=\linewidth]{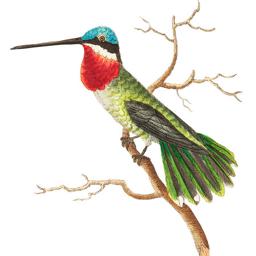} \\
                
                & & & & & & & & \\
                \resizebox{0.75in}{!}{\rotatebox{90}{DDA Output}} &
                \includegraphics[width=\linewidth]{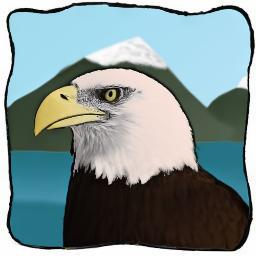} &
                \includegraphics[width=\linewidth]{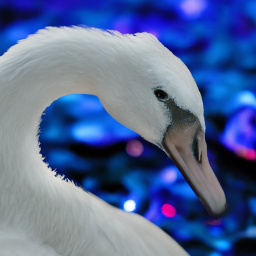} &
                \includegraphics[width=\linewidth]{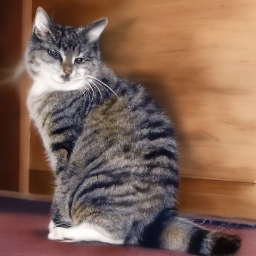} &
                \includegraphics[width=\linewidth]{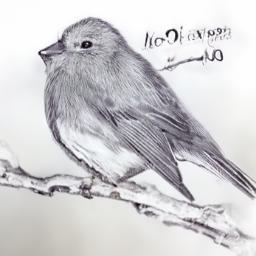} &
                \includegraphics[width=\linewidth]{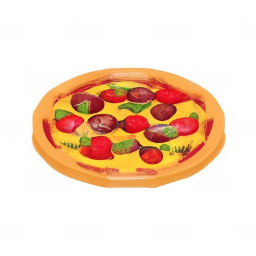} &
                \includegraphics[width=\linewidth]{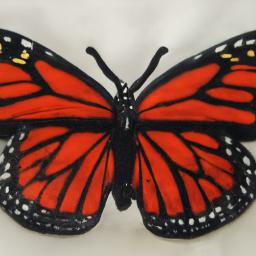} &
                \includegraphics[width=\linewidth]{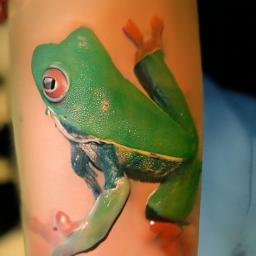} &
                \includegraphics[width=\linewidth]{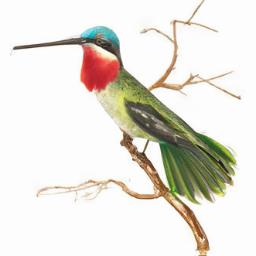} \\

            \end{tabular}
    	}

	\end{center}
	\caption{Visualization of generated images with DDA on ImageNet-R.}
	\label{fig:appendix_imagenetr}
\end{figure*}

%% file: fig/appendix_timestep.tex
\begin{figure*}[ht]
    \begin{center}
    	\adjustbox{max width=0.78\linewidth}{
            \begin{tabular}{c c c c c c c c}
                \resizebox{!}{0.7in}{corrupted} & \resizebox{!}{0.7in}{input} & \resizebox{!}{0.7in}{20\%} & \resizebox{!}{0.7in}{40\%} & \resizebox{!}{0.7in}{60\%} & \resizebox{!}{0.7in}{80\%} & \resizebox{!}{0.6in}{output} & \resizebox{!}{0.7in}{original} \\
                & & & & & & & \\
                & & & & & & & \\
                \includegraphics[width=\linewidth]{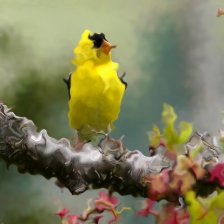} &
                \includegraphics[width=\linewidth]{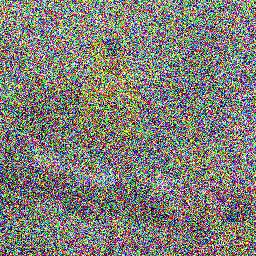} &
                \includegraphics[width=\linewidth]{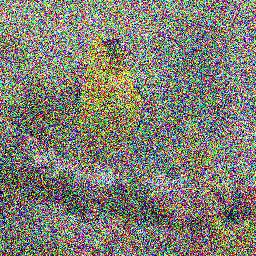} &
                \includegraphics[width=\linewidth]{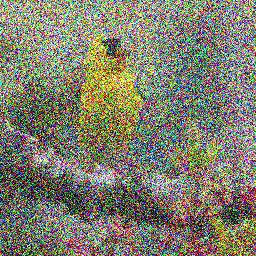} &
                \includegraphics[width=\linewidth]{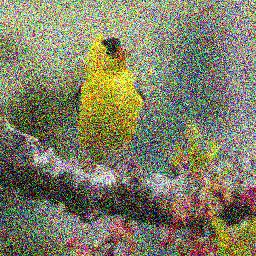} &
                \includegraphics[width=\linewidth]{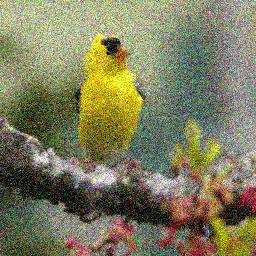} &
                \includegraphics[width=\linewidth]{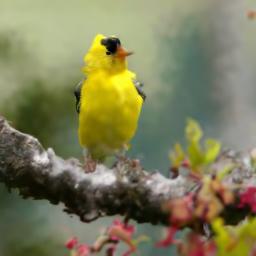} &
                \includegraphics[width=\linewidth]{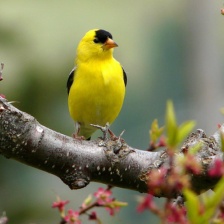} \\
                & & & & & & & \\
                \includegraphics[width=\linewidth]{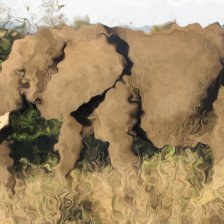} &
                \includegraphics[width=\linewidth]{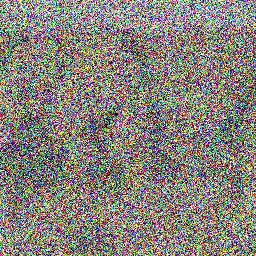} &
                \includegraphics[width=\linewidth]{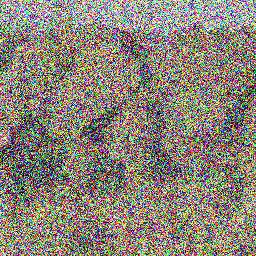} &
                \includegraphics[width=\linewidth]{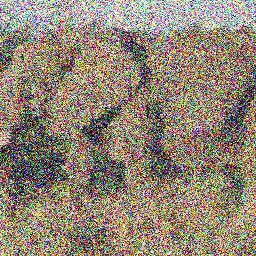} &
                \includegraphics[width=\linewidth]{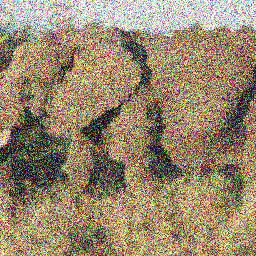} &
                \includegraphics[width=\linewidth]{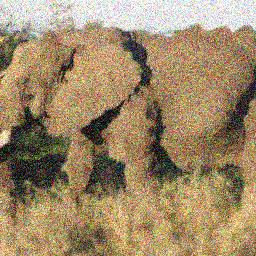} &
                \includegraphics[width=\linewidth]{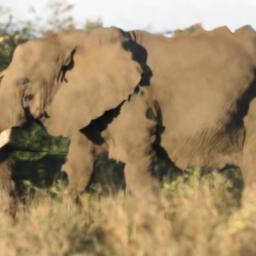} &
                \includegraphics[width=\linewidth]{ablation/origin/ILSVRC2012_val_00046757} \\
                & & & & & & & \\
                \includegraphics[width=\linewidth]{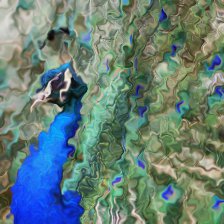} &
                \includegraphics[width=\linewidth]{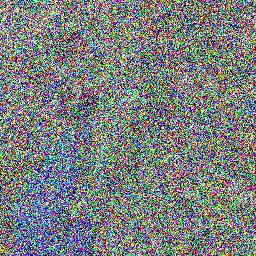} &
                \includegraphics[width=\linewidth]{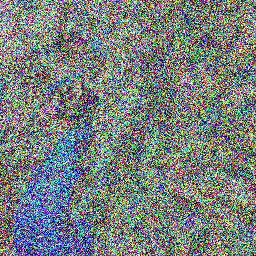} &
                \includegraphics[width=\linewidth]{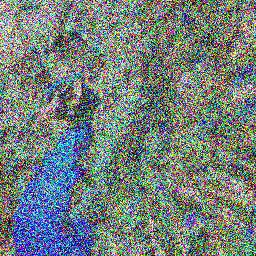} &
                \includegraphics[width=\linewidth]{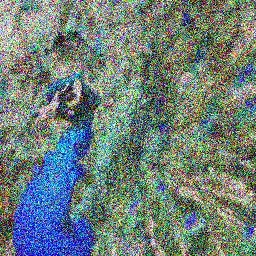} &
                \includegraphics[width=\linewidth]{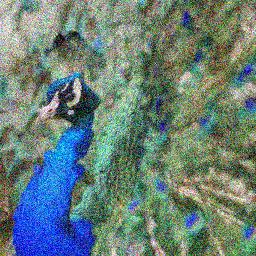} &
                \includegraphics[width=\linewidth]{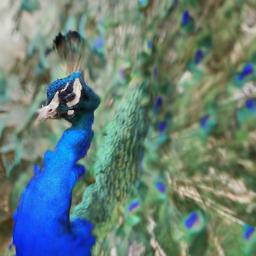} &
                \includegraphics[width=\linewidth]{timestep/ILSVRC2012_val_00028159_r256c224} \\
                & & & & & & & \\
                \includegraphics[width=\linewidth]{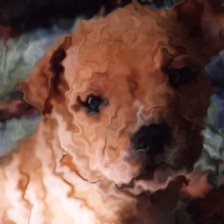} &
                \includegraphics[width=\linewidth]{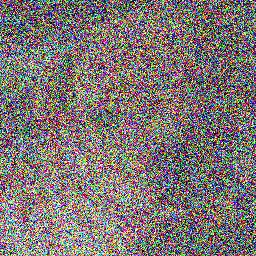} &
                \includegraphics[width=\linewidth]{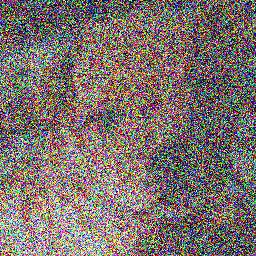} &
                \includegraphics[width=\linewidth]{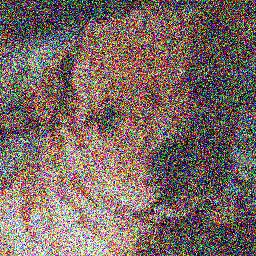} &
                \includegraphics[width=\linewidth]{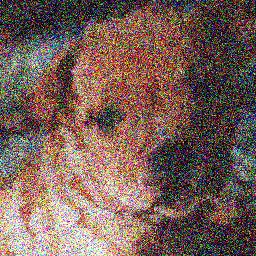} &
                \includegraphics[width=\linewidth]{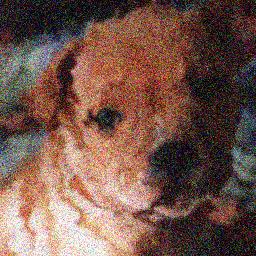} &
                \includegraphics[width=\linewidth]{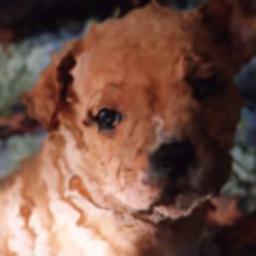} &
                \includegraphics[width=\linewidth]{timestep/ILSVRC2012_val_00046547_r256c224} \\
                & & & & & & & \\
                \midrule[3mm]
                & & & & & & & \\
                \includegraphics[width=\linewidth]{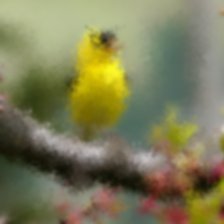} &
                \includegraphics[width=\linewidth]{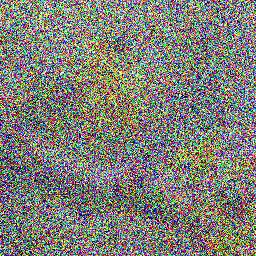} &
                \includegraphics[width=\linewidth]{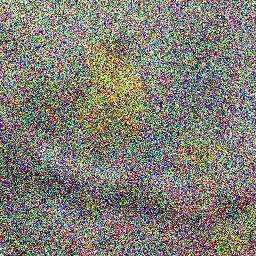} &
                \includegraphics[width=\linewidth]{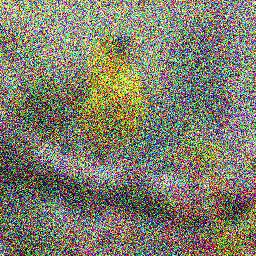} &
                \includegraphics[width=\linewidth]{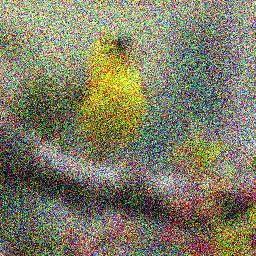} &
                \includegraphics[width=\linewidth]{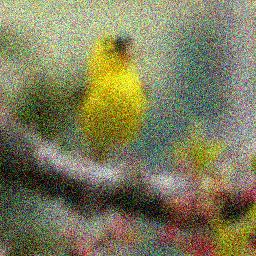} &
                \includegraphics[width=\linewidth]{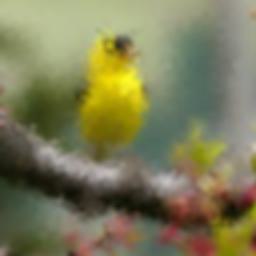} &
                \includegraphics[width=\linewidth]{timestep/ILSVRC2012_val_00005567_r256c224} \\
                & & & & & & & \\
                \includegraphics[width=\linewidth]{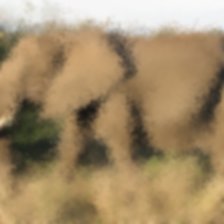} &
                \includegraphics[width=\linewidth]{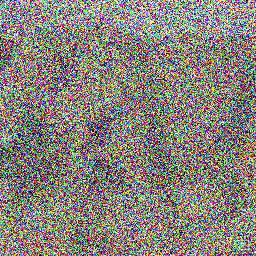} &
                \includegraphics[width=\linewidth]{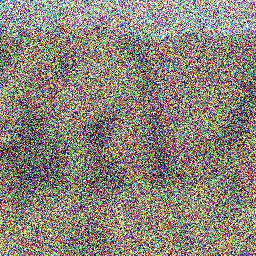} &
                \includegraphics[width=\linewidth]{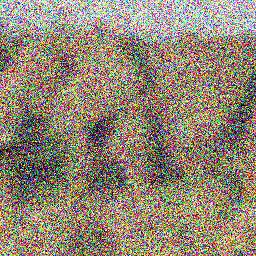} &
                \includegraphics[width=\linewidth]{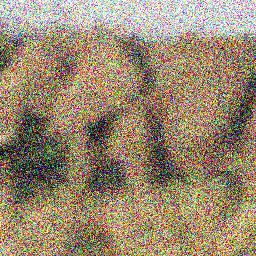} &
                \includegraphics[width=\linewidth]{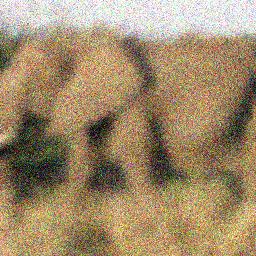} &
                \includegraphics[width=\linewidth]{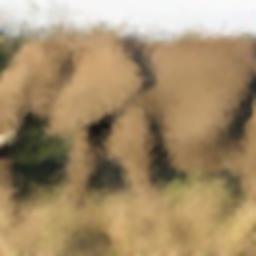} &
                \includegraphics[width=\linewidth]{ablation/origin/ILSVRC2012_val_00046757} \\
                & & & & & & & \\
                \includegraphics[width=\linewidth]{imagenet-c/glass_blur/5/n01806143/ILSVRC2012_val_00028159} &
                \includegraphics[width=\linewidth]{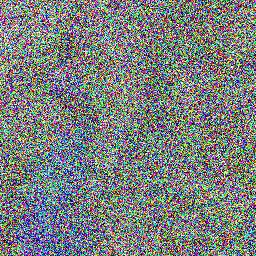} &
                \includegraphics[width=\linewidth]{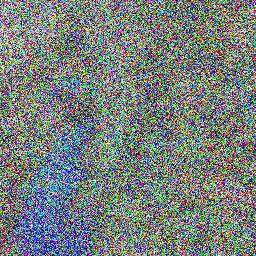} &
                \includegraphics[width=\linewidth]{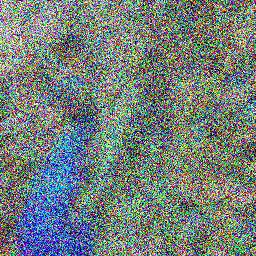} &
                \includegraphics[width=\linewidth]{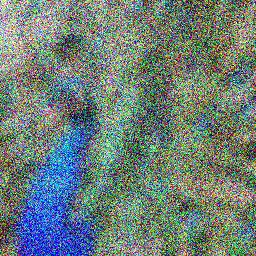} &
                \includegraphics[width=\linewidth]{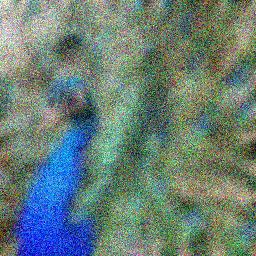} &
                \includegraphics[width=\linewidth]{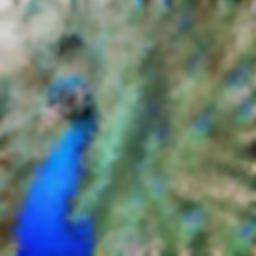} &
                \includegraphics[width=\linewidth]{timestep/ILSVRC2012_val_00028159_r256c224} \\
                & & & & & & & \\
                \includegraphics[width=\linewidth]{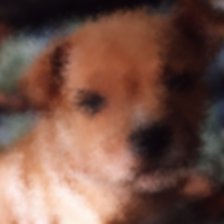} &
                \includegraphics[width=\linewidth]{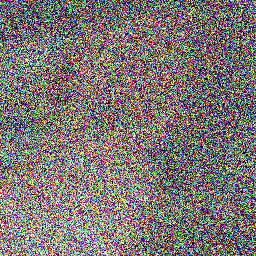} &
                \includegraphics[width=\linewidth]{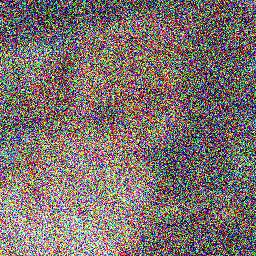} &
                \includegraphics[width=\linewidth]{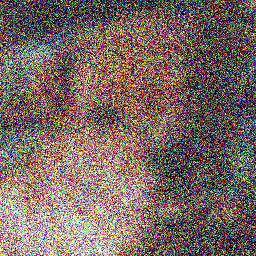} &
                \includegraphics[width=\linewidth]{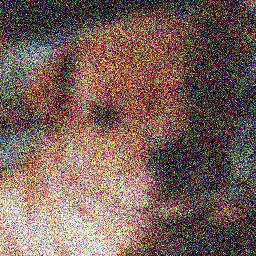} &
                \includegraphics[width=\linewidth]{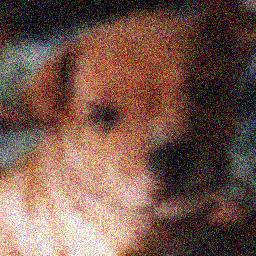} &
                \includegraphics[width=\linewidth]{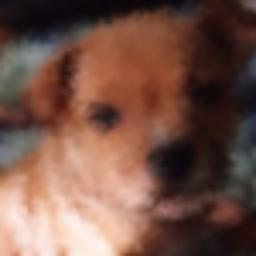} &
                \includegraphics[width=\linewidth]{timestep/ILSVRC2012_val_00046547_r256c224} \\
                & & & & & & & \\
                \midrule[3mm]
                & & & & & & & \\
                \includegraphics[width=\linewidth]{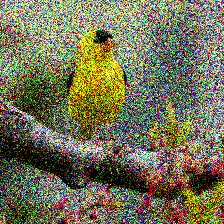} &
                \includegraphics[width=\linewidth]{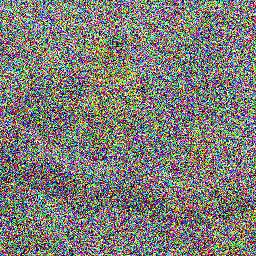} &
                \includegraphics[width=\linewidth]{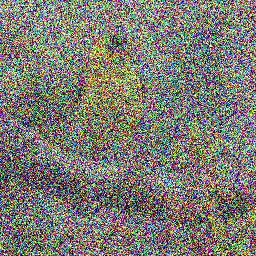} &
                \includegraphics[width=\linewidth]{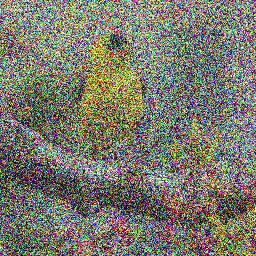} &
                \includegraphics[width=\linewidth]{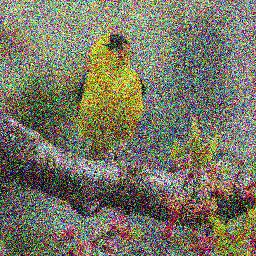} &
                \includegraphics[width=\linewidth]{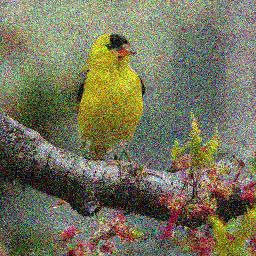} &
                \includegraphics[width=\linewidth]{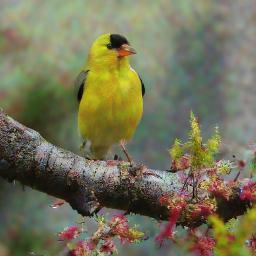} &
                \includegraphics[width=\linewidth]{timestep/ILSVRC2012_val_00005567_r256c224} \\
                & & & & & & & \\
                \includegraphics[width=\linewidth]{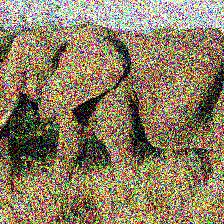} &
                \includegraphics[width=\linewidth]{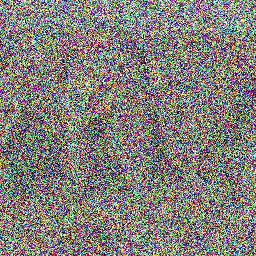} &
                \includegraphics[width=\linewidth]{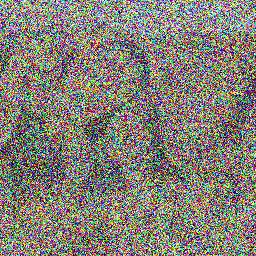} &
                \includegraphics[width=\linewidth]{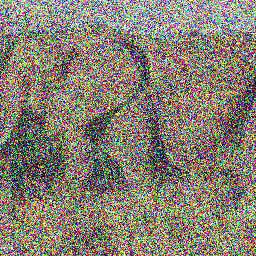} &
                \includegraphics[width=\linewidth]{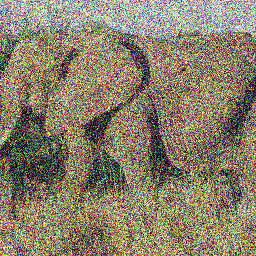} &
                \includegraphics[width=\linewidth]{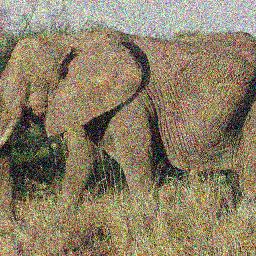} &
                \includegraphics[width=\linewidth]{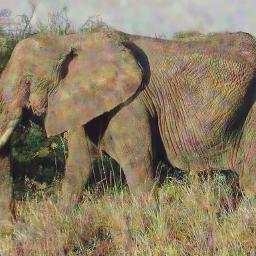} &
                \includegraphics[width=\linewidth]{ablation/origin/ILSVRC2012_val_00046757} \\
                & & & & & & & \\
                \includegraphics[width=\linewidth]{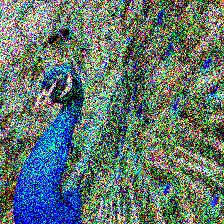} &
                \includegraphics[width=\linewidth]{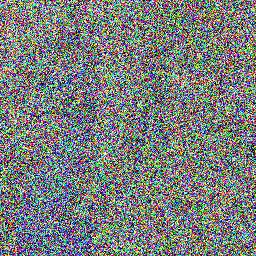} &
                \includegraphics[width=\linewidth]{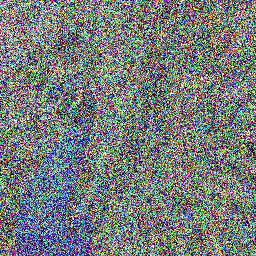} &
                \includegraphics[width=\linewidth]{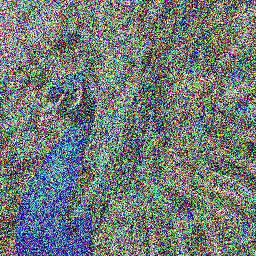} &
                \includegraphics[width=\linewidth]{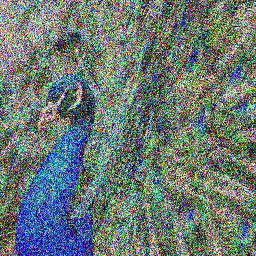} &
                \includegraphics[width=\linewidth]{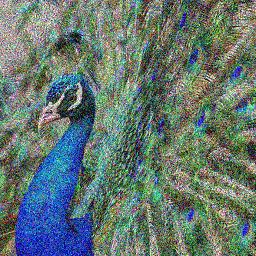} &
                \includegraphics[width=\linewidth]{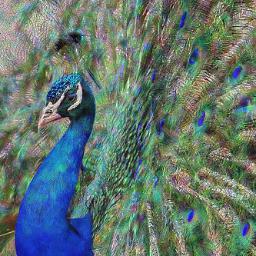} &
                \includegraphics[width=\linewidth]{timestep/ILSVRC2012_val_00028159_r256c224} \\
                & & & & & & & \\
                \includegraphics[width=\linewidth]{imagenet-c/shot_noise/5/n02093256/ILSVRC2012_val_00046547} &
                \includegraphics[width=\linewidth]{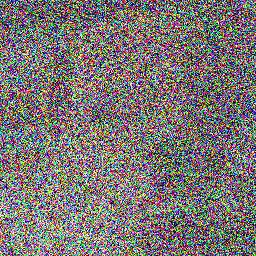} &
                \includegraphics[width=\linewidth]{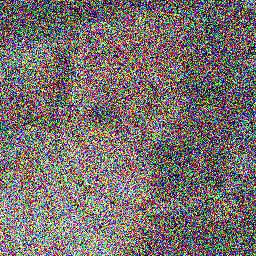} &
                \includegraphics[width=\linewidth]{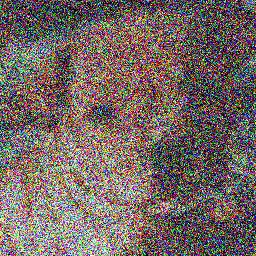} &
                \includegraphics[width=\linewidth]{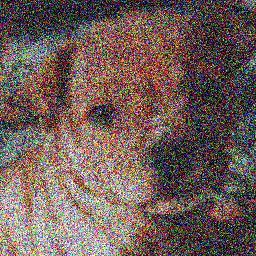} &
                \includegraphics[width=\linewidth]{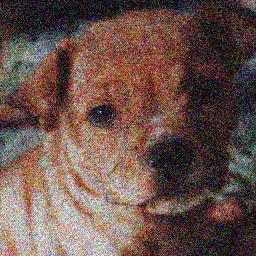} &
                \includegraphics[width=\linewidth]{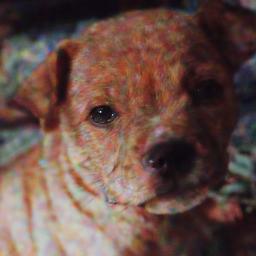} &
                \includegraphics[width=\linewidth]{timestep/ILSVRC2012_val_00046547_r256c224} \\
            \end{tabular}
    	}

	\end{center}
	\caption{Visualization of generated images with diffusion models, given highest severity corrupted images during the test time.}
	\label{appendixFig:timestep}
\end{figure*}

%% file: fig/timestep_origin.tex
\begin{figure*}[ht]
    \begin{center}
    	\adjustbox{max width=\linewidth}{
            \begin{tabular}{c c c c c c c c}
                \resizebox{!}{0.7in}{original} & \resizebox{!}{0.7in}{input} & \resizebox{!}{0.7in}{20\%} & \resizebox{!}{0.7in}{40\%} & \resizebox{!}{0.7in}{60\%} & \resizebox{!}{0.7in}{80\%} & \resizebox{!}{0.6in}{output} & \resizebox{!}{0.7in}{original} \\
                & & & & & & & \\
                & & & & & & & \\
                \includegraphics[width=\linewidth]{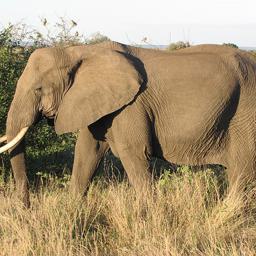} &
                \includegraphics[width=\linewidth]{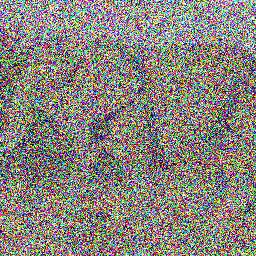} &
                \includegraphics[width=\linewidth]{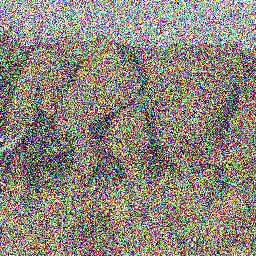} &
                \includegraphics[width=\linewidth]{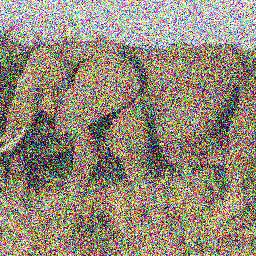} &
                \includegraphics[width=\linewidth]{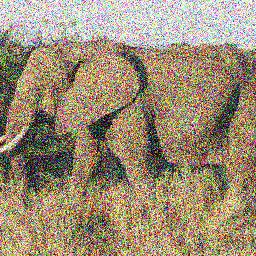} &
                \includegraphics[width=\linewidth]{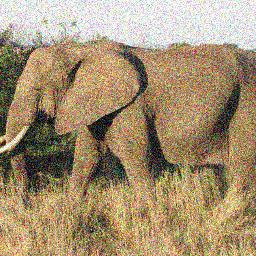} &
                \includegraphics[width=\linewidth]{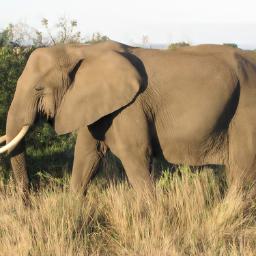} &
                \includegraphics[width=\linewidth]{ablation/origin/ILSVRC2012_val_00046757r256} \\
                & & & & & & & \\
                \includegraphics[width=\linewidth]{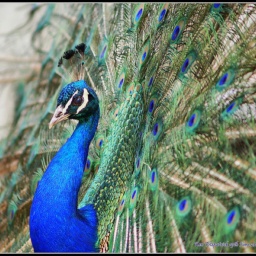} &
                \includegraphics[width=\linewidth]{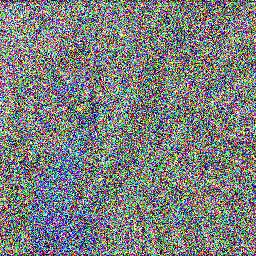} &
                \includegraphics[width=\linewidth]{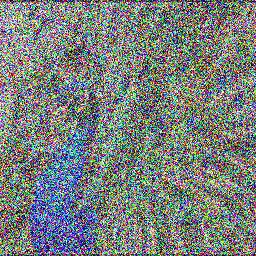} &
                \includegraphics[width=\linewidth]{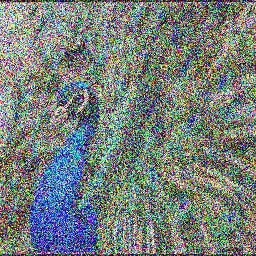} &
                \includegraphics[width=\linewidth]{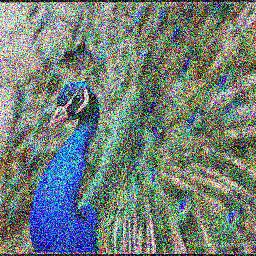} &
                \includegraphics[width=\linewidth]{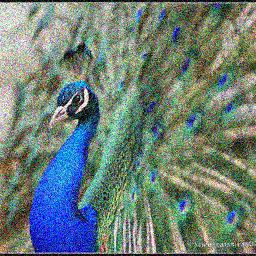} &
                \includegraphics[width=\linewidth]{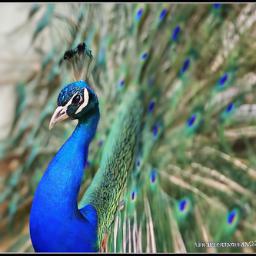} &
                \includegraphics[width=\linewidth]{timestep/ILSVRC2012_val_00028159_r256c256} \\
                & & & & & & & \\
                \includegraphics[width=\linewidth]{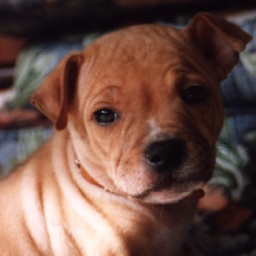} &
                \includegraphics[width=\linewidth]{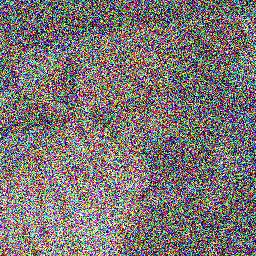} &
                \includegraphics[width=\linewidth]{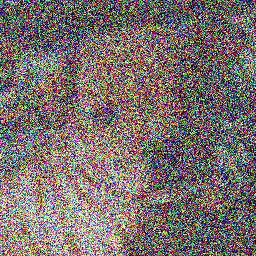} &
                \includegraphics[width=\linewidth]{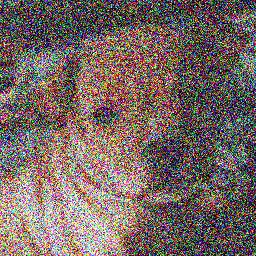} &
                \includegraphics[width=\linewidth]{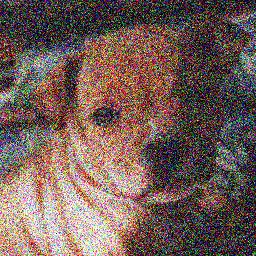} &
                \includegraphics[width=\linewidth]{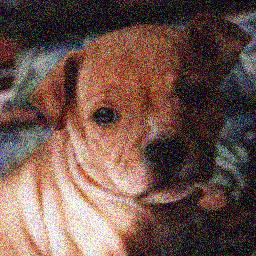} &
                \includegraphics[width=\linewidth]{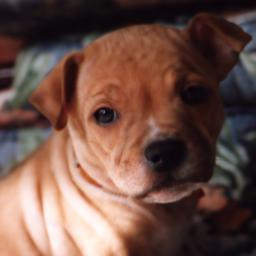} &
                \includegraphics[width=\linewidth]{timestep/ILSVRC2012_val_00046547_r256c256} \\
            \end{tabular}
    	}

	\end{center}
	\caption{Visualization of the progress of image generation with diffusion models, given the original images during the test time. }
	\label{fig:timestep_origin}
\end{figure*}

%% file: fig/corruption_positive.tex
\begin{figure*}[ht]
    \begin{center}
            \adjustbox{max width=\linewidth}{
            \begin{tabular}{c c c c c c c c}
                &
                \resizebox{!}{14.7mm}{gauss} & 
                \resizebox{!}{18.7mm}{shot} & 
                \resizebox{!}{17.7mm}{impulse} & 
                \resizebox{!}{17.7mm}{glass} & 
                \resizebox{!}{17.7mm}{elastic} & 
                \resizebox{!}{17.7mm}{pixelate} & 
                \resizebox{!}{17.7mm}{jpeg} \\
                \resizebox{0.8in}{!}{\rotatebox{90}{(a) corruption}} &
                \includegraphics[width=\linewidth]{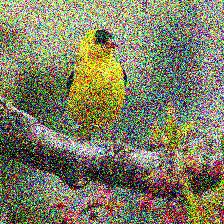} &
                \includegraphics[width=\linewidth]{imagenet-c/shot_noise/5/n01531178/ILSVRC2012_val_00005567} &
                \includegraphics[width=\linewidth]{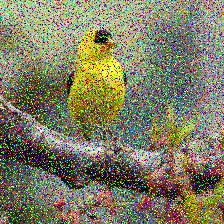} &
                \includegraphics[width=\linewidth]{imagenet-c/glass_blur/5/n01531178/ILSVRC2012_val_00005567} &
                \includegraphics[width=\linewidth]{imagenet-c/elastic_transform/5/n01531178/ILSVRC2012_val_00005567} &
                \includegraphics[width=\linewidth]{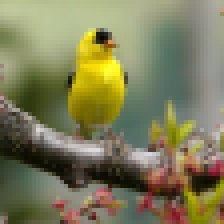} &
                \includegraphics[width=\linewidth]{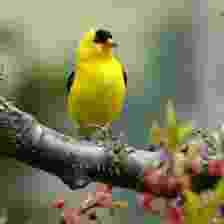} \\
                \resizebox{0.8in}{!}{\rotatebox{90}{(b) diffusion}} &
                \includegraphics[width=\linewidth]{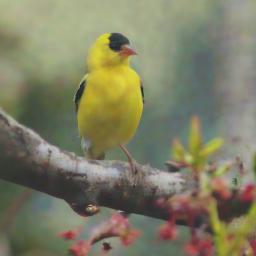} &
                \includegraphics[width=\linewidth]{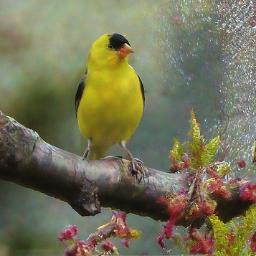} &
                \includegraphics[width=\linewidth]{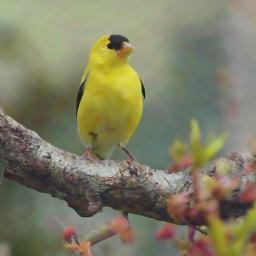} &
                \includegraphics[width=\linewidth]{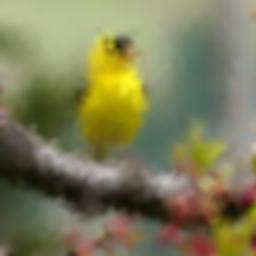} &
                \includegraphics[width=\linewidth]{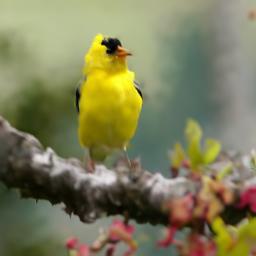} &
                \includegraphics[width=\linewidth]{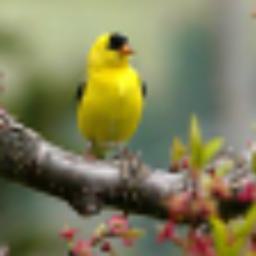} &
                \includegraphics[width=\linewidth]{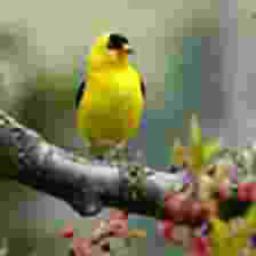} \\
            \end{tabular}} \\
	\end{center}
	\caption{Visualization of \emph{positive} generated images with diffusion models, given highest severity corrupted images during the test time.}
	\label{fig:corruption_positive}
\end{figure*}

%% file: fig/corruption_negative.tex
\begin{figure*}[ht]
    \begin{center}
            \adjustbox{max width=\linewidth}{
            \begin{tabular}{c c c c c c c c c}
                &
                \resizebox{!}{17.7mm}{defocus} & 
                \resizebox{!}{17.7mm}{motion} & 
                \resizebox{!}{14.7mm}{zoom} & 
                \resizebox{!}{14.7mm}{snow} & 
                \resizebox{!}{17.7mm}{frost} & 
                \resizebox{!}{17.7mm}{fog} & 
                \resizebox{!}{17.7mm}{bright} & 
                \resizebox{!}{15.7mm}{contrast} \\
                \resizebox{0.8in}{!}{\rotatebox{90}{(a) corruption}} &
                \includegraphics[width=\linewidth]{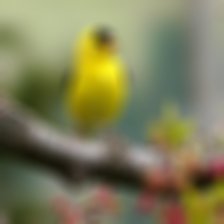} &
                \includegraphics[width=\linewidth]{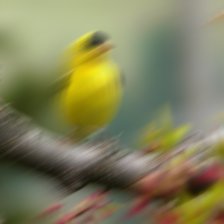} &
                \includegraphics[width=\linewidth]{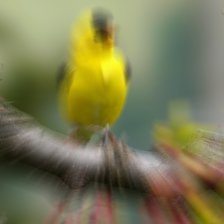} &
                \includegraphics[width=\linewidth]{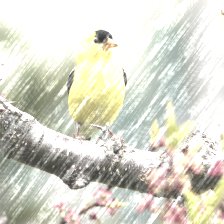} &
                \includegraphics[width=\linewidth]{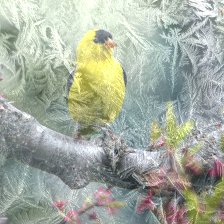} &
                \includegraphics[width=\linewidth]{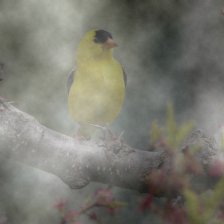} &
                \includegraphics[width=\linewidth]{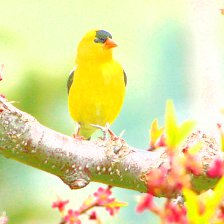} &
                \includegraphics[width=\linewidth]{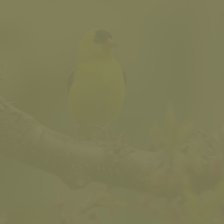} \\
                \resizebox{0.8in}{!}{\rotatebox{90}{(b) diffusion}} &
                \includegraphics[width=\linewidth]{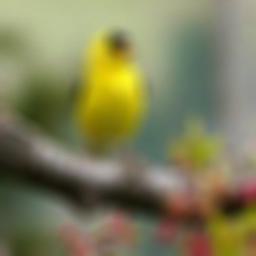} &
                \includegraphics[width=\linewidth]{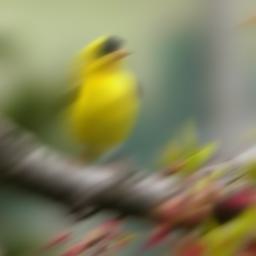} &
                \includegraphics[width=\linewidth]{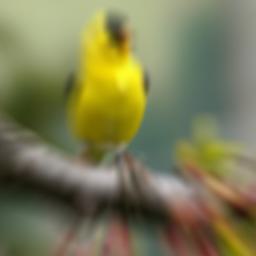} &
                \includegraphics[width=\linewidth]{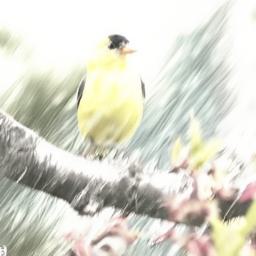} &
                \includegraphics[width=\linewidth]{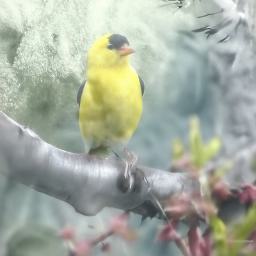} &
                \includegraphics[width=\linewidth]{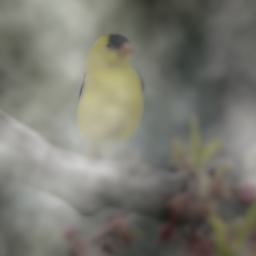} &
                \includegraphics[width=\linewidth]{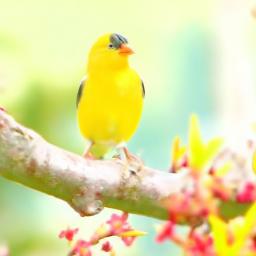} &
                \includegraphics[width=\linewidth]{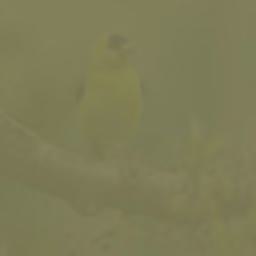} \\
                
            \end{tabular}} \\
	\end{center}
	\caption{Visualization of \emph{negative} generated images with diffusion models, given highest severity corrupted images during the test time.}
	\label{fig:corruption_negative}
\end{figure*}